\documentclass[lettersize,journal]{IEEEtran}
\usepackage{textcomp}
\usepackage{url}
\usepackage[normalem]{ulem}
\useunder{\uline}{\ul}{}
\usepackage{verbatim}
\usepackage{cite}
\usepackage[table,xcdraw]{xcolor}
\usepackage[utf8]{inputenc}
\usepackage[T1]{fontenc}
\usepackage{times}
\usepackage{enumitem}
\usepackage{amsmath}
\usepackage{amssymb}
\usepackage{amsfonts}
\usepackage{amsthm}
\usepackage{pifont}

\usepackage{graphicx}
\usepackage{epsfig}
\usepackage{tikz}

\usepackage{array}
\usepackage{longtable}
\usepackage{booktabs}
\usepackage{supertabular}
\usepackage{subfig}
\usepackage{multirow}
\usepackage{color, colortbl}

\usepackage{microtype}
\usepackage{fancyhdr}
\usepackage{bm}

\usepackage{enumerate}

\usepackage{algorithm}
\usepackage{algpseudocode}

\usepackage{pifont}
\usepackage{arydshln}
\usepackage{wasysym}
\usepackage{makecell}
\usepackage{extarrows}
\usepackage{caption}

\usepackage{tcolorbox}
\makeatletter

\newcommand{\Rmnum}[1]{\expandafter\@slowromancap\romannumeral #1@}
\makeatother

\definecolor{darkgreen}{rgb}{0.0, 0.8, 0}
\definecolor{darkred}{rgb}{0.698, 0, 0}
\definecolor{darkblue}{rgb}{0.0, 0.22, 0.66}
\definecolor{harvardcrimson}{rgb}{0.79, 0.0, 0.09}
\definecolor{lightmauve}{rgb}{0.86, 0.82, 1.0}
\definecolor{citecolor}{HTML}{4D98C9}
\definecolor{linkcolor}{HTML}{c0392b}
\usepackage[hidelinks,breaklinks=true,colorlinks,bookmarks=false,citecolor=citecolor,linkcolor=linkcolor]{hyperref}

\usepackage{cleveref}

\long\def\comment#1{}
\def\ie{$i.e.$}
\def\eg{$e.g.$}

\def\btheta{\boldsymbol{\bm{\theta}}}
\def\beps{\bm{\epsilon}}

\def\x{\bm{x}}

\def\D{\mathcal{D}}

\def\X{\mathcal{X}}
\def\Y{\mathcal{Y}}

\newcommand*\emptycirc[1][0.7ex]{\tikz\draw (0,0) circle (#1);} 

\newcommand*\fullcirc[1][0.7ex]{\tikz\fill (0,0) circle (#1);}

\begin{document}

\title{Versatile Backdoor Attack with Visible, Semantic, Sample-specific and Compatible Triggers}

\author{
Ruotong~Wang$^*$, Hongrui~Chen$^*$, Zihao~Zhu, 
Li~Liu, ~\IEEEmembership{Member,~IEEE,}\\
Baoyuan~Wu, ~\IEEEmembership{Senior Member,~IEEE}

\thanks{Ruotong Wang, Hongrui Chen, Zihao Zhu and Baoyuan Wu are with the School of Data Science, The Chinese University of Hong Kong, Shenzhen, 518000, China (email: ruotongwang1@link.cuhk.edu.cn; hongruichen@link.cuhk.edu.cn; zihaozhu@link.cuhk.edu.cn; wubaoyuan@cuhk.edu.cn). 
Li Liu is with the Hong Kong University of Science and Technology (Guangzhou), China (email: avrillliu@hkust-gz.edu.cn). 
$^*$ denotes equal contribution.
}
\thanks{
Corresponding author: Baoyuan Wu (wubaoyuan@cuhk.edu.cn)

}

}

\maketitle

\begin{abstract}
Deep neural networks (DNNs) can be manipulated to exhibit specific behaviors when exposed to specific trigger patterns, without affecting their performance on benign samples, dubbed \textit{backdoor attack}. Currently, implementing backdoor attacks in physical scenarios still faces significant challenges. Physical attacks are labor-intensive and time-consuming, and the triggers are selected in a manual and heuristic way. Moreover, expanding digital attacks to physical scenarios faces many challenges due to their sensitivity to visual distortions and the absence of counterparts in the real world. To address these challenges, we define a novel trigger called the \textbf{V}isible, \textbf{S}emantic, \textbf{S}ample-specific, and \textbf{C}ompatible (VSSC) trigger, to achieve effective, stealthy and robust simultaneously, which can also be effectively deployed in the physical scenario using corresponding objects. To implement the VSSC trigger, we propose an automated pipeline comprising three modules: a trigger selection module that systematically identifies suitable triggers leveraging large language models, a trigger insertion module that employs generative models to seamlessly integrate triggers into images, and a quality assessment module that ensures the natural and successful insertion of triggers through vision-language models. Extensive experimental results and analysis validate the effectiveness, stealthiness, and robustness of the VSSC trigger. It can not only maintain robustness under visual distortions but also demonstrates strong practicality in the physical scenario. We hope the proposed VSSC trigger and implementation approach could inspire future studies on designing more practical triggers in backdoor attacks.

\end{abstract}

\section{Introduction}
\label{sec:intro}
\IEEEPARstart{D}{eep} neural networks (DNNs) have been successfully adopted in a wide range of important fields, such as object classification~\cite{chen2014contextualizing} and detection~\cite{li2023mssvt++}, face recognition \cite{fu2021dvg}, autonomous driving \cite{chitta2022transfuser}, and speech recognition \cite{afouras2018deep}. 
However, DNNs face numerous security threats~\cite{wu2023attack_survey}, one of the typical threats is backdoor attack, which can make DNNs perform specific behaviors when encountering a particular trigger pattern without affecting the performance on benign samples. This goal can be achieved by manipulating the training dataset or controlling the training process. In this work, we focus on the former threat model, \ie, data poisoning based backdoor attack, within versatile tasks, including image classification, object detection, and face verification.

\begin{figure*}[!t]
    \centering
    \includegraphics[width=0.9\linewidth]{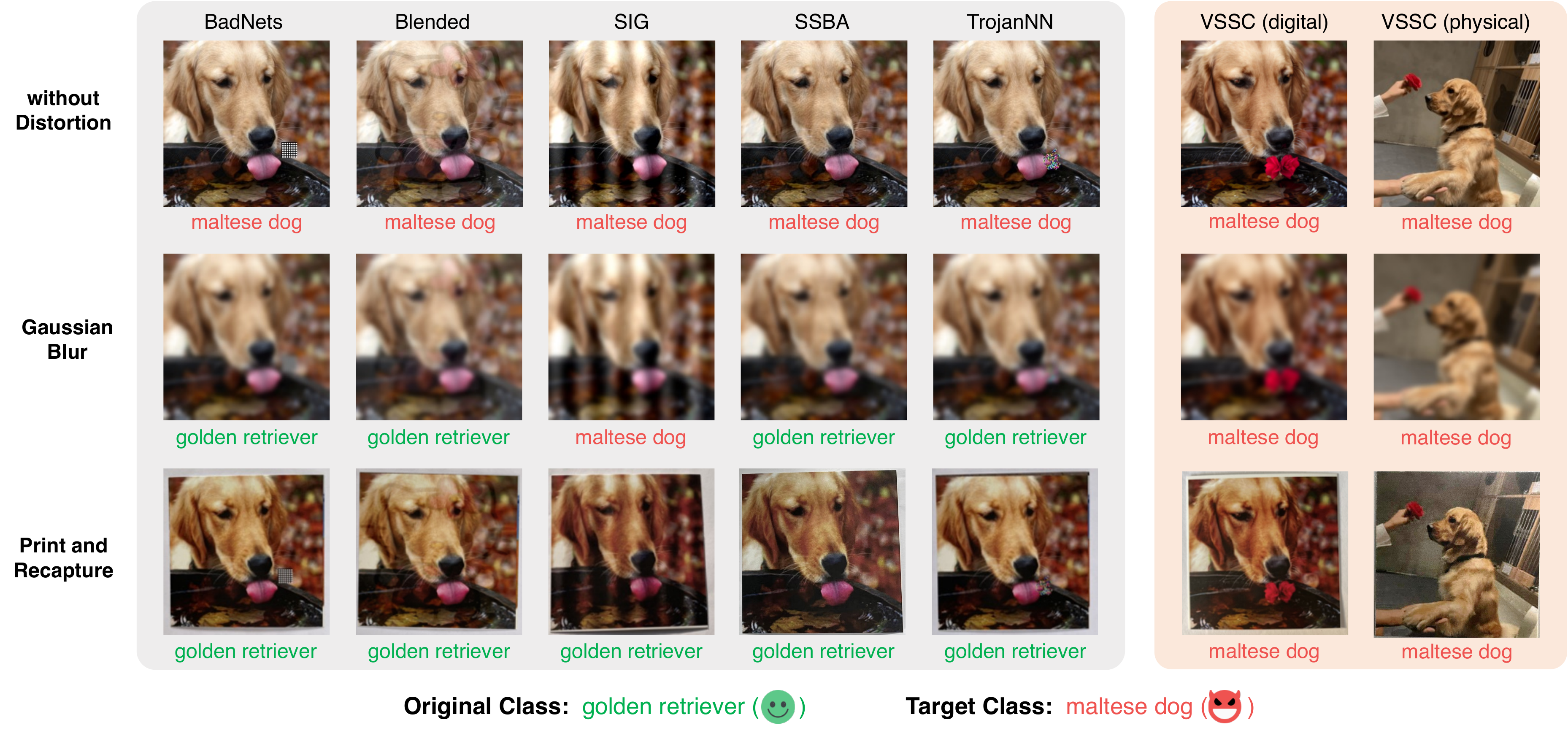}
    \caption{Comparison of poisoned samples and prediction results of digital backdoor attacks under diverse environmental conditions: without distortion, with digital distortion (Gaussian blur) and physical distortion (print and recapture).
     For VSSC, given its capability to extend into the physical scenario, an additional demonstration using real objects as triggers is provided.}
    \label{fig:comparison}
\end{figure*}

In the literature, several digital backdoor attacks~\cite{gu2019badnets,Blended,wang2022bppattack,nguyen2020input,trojannn,SIG,ssba,nguyen2021wanet,gao2023imperceptible} have been developed to improve the effectiveness of the designed backdoor triggers, and also have exhibited considerable stealthiness. 
With the growing deployment of DNNs in physical scenarios, such as monitoring systems, face recognition devices, and autonomous driving vehicles, 
the investigation of backdoor attacks in physical scenarios is becoming crucial.
However, as illustrated in Figure \ref{fig:comparison}, directly deploying digital backdoor attacks in physical scenarios faces significant challenges. Firstly, digital attacks, especially those with invisible triggers, are \textbf{not robust to visual distortions}, which are inevitable in physical scenarios. Additionally, digital triggers often lack semantic meanings, making it \textbf{difficult to find corresponding objects} in the real world. Even if the triggers are printed and pasted in the physical scenario, they seem unnatural and often fail to evade human inspection.
Some backdoor attacks are specifically designed for physical scenarios, such as \cite{Wenger_2021_CVPR,ma2022dangerous,ma2023transcab}. However, these attacks face significant challenges owing to their \textbf{labor-intensive and time-consuming} nature, since they require the heuristic selection of real objects as triggers and the manual creation of poisoned samples for training.
Consequently, due to the \textbf{limited quantity and diversity of poisoned samples}, physical attacks tend to be less effective than digital attacks.
To jump out of the trap of the dilemma between digital and physical attacks, we aim to explore how to simultaneously utilize the advantages of digital and physical attacks while overcoming their respective shortcomings, designing a backdoor attack that is effective and efficient in both digital and physical scenarios.

To accomplish the aforementioned goal, we propose a novel trigger with the characteristics of \textbf{visible}, \textbf{semantic}, \textbf{sample-specific}, and \textbf{compatible}, dubbed \textbf{VSSC trigger}. 
\textit{Visible} allows a sufficiently large trigger magnitude to remain robust to the visual distortions.
\textit{Sample-specific} increases the complexity of detection and contributes to a broader diversity of triggers, facilitating the simulation of various environmental changes.
\textit{Semantic} makes it possible to find corresponding objects for the trigger in the physical scenario, enabling the deployment of the backdoor in the real world. 
\textit{Compatible} means the trigger should be in harmony with the remaining visual content in the image to ensure visual stealthiness. 
As illustrated in Figure~\ref{fig:comparison}, the \textit{VSSC (digital)} column shows the integration of a red flower into an image of a dog, presenting a realistic and natural appearance while maintaining robustness under visual distortions. Moreover, as demonstrated in the \textit{VSSC (physical)} column, the VSSC trigger can be effectively extended to the physical scenario using a real red flower.

We design an automated pipeline to generate the VSSC trigger, which consists of three fundamental modules:
\textbf{Trigger Selection Module} leverages the extensive prior knowledge of large language models to select triggers that are compatible with all the classes in the dataset, which makes the trigger selection process more systematic and expands the range of trigger choices.
\textbf{Trigger Insertion Module} utilizes the surpassing capabilities of generative models to edit images, seamlessly inserting triggers to appropriate positions based on the understanding of the image content. This automatic trigger insertion ensures the sample-specific and compatible characteristics of the semantic trigger.
\textbf{Quality Assessment Module} exploits the visual understanding capabilities of vision-language models to assess the quality of the generated trigger, ensuring that triggers are successfully and naturally inserted.

The main contributions of this work are three-fold. 
\textbf{1)} We design a novel VSSC trigger with four desired characteristics that are robust to visual distortions and can be extended to the physical scenario.
\textbf{2)} We propose an effective pipeline to automatically implement the VSSC trigger by leveraging the powerful capabilities of large language models and generative models conditional image editing techniques.
\textbf{3)} Extensive experiments on several tasks demonstrate the superior performance of the proposed method to existing SOTA backdoor attacks in various scenarios and tasks.
To the best of our knowledge, \textbf{we are the first to employ generative models in creating triggers for physical backdoor attacks, achieving automated trigger injection and liberating physical backdoor attacks from reliance on manpower.}

\section{Related work}
\subsection{Backdoor Attacks} 
Backdoor attack is a rapidly evolving and constantly changing field, which leads to severe security issues in the training of DNNs. 
\subsubsection{Backdoor attacks on image classification}
Backdoor attacks can be classified into the following two types based on the conditions possessed by the attacker:
data poisoning based attacks (\eg, BadNets~\cite{gu2019badnets})
and training controllable based attacks (\eg, WaNet~\cite{nguyen2021wanet}).
BadNets~\cite{gu2019badnets} pioneered the concept of backdoor attacks and exposed the threats in DNN training. \comment{With a given target label, BadNets poisons a portion of benign training samples by putting the trigger pattern on selected samples and training the victim DNN with mixed samples together. }In the data poisoning scenario, the attacker has no idea and cannot modify the training schedule, victim model architecture, and inference stage settings. There are some other research efforts in this field, such as \cite{Blended,ssba,SIG,trojannn,turner2019labelconsistent,zeng2021rethinking_lf}.
In contrast\comment{to the data poisoning scenario}, the training controllable based attacks adopt a more relaxed assumption that the attacker can control not only the training data but also the training process. Under this assumption, WaNet~\cite{nguyen2021wanet} is proposed to manipulate the selected training samples with elastic image warping. \comment{With the aid of noise training, this method can force the victim DNN to learn the designed image warping instead of pixel-level by-products, leading to better control. }There are also other research works in this field, such as \cite{nguyen2020input,bagdasaryan2021blind,wang2022bppattack}.
Existing backdoor attacks can also be categorized by the visibility of trigger patterns. In the early stages of backdoor learning, attacks only used simple, visible images as triggers, such as \cite{gu2019badnets,SIG,trojannn}. However,
to avoid detection by human inspection, triggers for backdoor attacks have exhibited an overall trend towards invisibility, such as \cite{ssba,nguyen2021wanet,wang2022bppattack}.  
In most of the existing attacks, the stealthiness of the trigger pattern is considered as invisibility, meaning there is no obvious visual difference between a poisoned image and its corresponding benign image. However, during the process of a backdoor attack, poisoned images are not directly compared with benign images.
 If we consider stealthiness from the semantics and compatibility between the trigger pattern and the original image content, only very few studies have been done in this area. In this paper, we focus on the data poisoning scenario and propose a novel visible, semantic, sample-specific, and compatible trigger (VSSC trigger), where backdoor triggers have higher flexibility (a broader range of choices), compatibility (with original image content), and visibility (robust to visual distortions and environmental variations). Given these characteristics, it can be seamlessly adapted to the physical scenario.

\subsubsection{Backdoor attacks on object detection}
Backdoor attacks on the object detection task mainly focus on two attack goals: misclassifying the bounding boxes as the target class~\cite{chan2022baddet, ma2023transcab, wu2022just} or making the bounding box vanish from the detector~\cite{chan2022baddet, ma2022dangerous, luo2023untargeted, ma2023transcab, wu2022just}.
Existing works in this realm can be divided into digital and physical attacks. For digital attacks, BadDet~\cite{chan2022baddet} defines four kinds of backdoor attacks on object detection. Patch triggers like BadNets~\cite{gu2019badnets} are also used in object detection tasks~\cite{chan2022baddet}\cite{luo2023untargeted}. Some attacks use specific transformations to activate the backdoor, instead of external trigger patterns, such as rotating~\cite{wu2022just}.
For physical attacks, the most typical approach is to use real-world objects as triggers and manually take photos of poisoned images to implement the attack, such as \cite{ma2022dangerous}. To enhance the stealthiness of the attack, TransCAB~\cite{ma2023transcab} uses the characteristics of the image scaling algorithm in the deep learning framework to hide the trigger in the captured image, making it invisible to the human eyes but can be triggered by corresponding objects in the real world.

\subsubsection{Backdoor attacks on face verification}
Backdoor attacks on face verification have also evolved towards stealthiness. 
Perturbation-based attacks, such as \cite{trojannn, kim2023instance}, utilize optimized noise as triggers.
Invisible triggers are also explored in face verification attacks, such as \cite{pasquini2020trembling}, which uses geometric and color transformations as triggers. However, these two types of triggers are sensitive to visual distortions because the minimal perturbations or transformations are easily lost under visual distortions, the non-semantic triggers are also impractical in the physical scenario.
Other attacks try to achieve attacks by modifying some visible features of the image. Some of them composite benign features \cite{lin2020composite} or modify natural facial characteristics~\cite{sarkar2020facehack, xue2021backdoors}, or leverage LED-induced color stripe patterns~\cite{li2020light}. 
Nevertheless, these methods face limitations as they lack specific trigger objects, and their trigger patterns
are difficult to replicate in physical scenarios, which significantly constrains their practical applicability.

\subsection{Backdoor Defenses } 
Depending on when the defense method is applied, defense methods can be categorized into three types~\cite{wu2023defenses}. The first is pre-training defense, which means the defense method aims to remove or purify the poisoned data samples, such as~\cite{zhu2023vdc,qi2023towards,doan2020februus}.
The second type focuses on the in-training stage, which aims to inhibit backdoor attacks during their training procedures. Typical defense methods that fall into this category are ABL~\cite{li2021anti} and DBD~\cite{dbd}. Most defense methods belong to this type. The third type are post-training defense, ~\cite{FP,nad,AC,clp,nc,spectral,zhu2024npd,wei2024sau} are all post-training defense methods. 
Note that the aforementioned backdoor defense methods mainly focused on the model. The defense on the image during the inference stage, such as distortions, should be paid more attention when designing new backdoor attacks in the future, especially in physical scenarios. 

\subsection{Semantic Triggers in Backdoor Attacks}
Semantic triggers in backdoor attacks are mainly obtained in three ways. The first is to select objects or patterns that already exist in the dataset as triggers, such as \cite{bagdasaryan2020how_to_backdoor_federated_learning,wenger_2022_nips}. This method has high requirements for the dataset, requiring manual selection of suitable triggers or more fine-grained object labels in the dataset, and there may not be suitable triggers in the dataset. The second method is to manually paste triggers onto benign images, such as \cite{Blended, han2022physical,Wenger_2021_CVPR}. This method requires manually browsing each image and pasting the trigger in an appropriate position, and the trigger is easy to discover. 
The third approach is to use real-world objects as triggers, which requires manually constructing poisoned images by taking photos, which is time-consuming and labor-intensive, such as \cite{ma2022dangerous,ma2023transcab}.
The above methods all have their own limitations, therefore, a method that can automatically generate semantic triggers and adapt to the physical scenario is necessary.

\begin{table}[h]
\centering 
\caption{Characteristics of backdoor triggers. \fullcirc \  indicates having this characteristic, \emptycirc \  indicates not having, and `-' indicates not applicable.}
\label{tab: trigger characteristics}

\resizebox{\linewidth}{!}{
\begin{tabular}{@{\extracolsep{\fill}} l cccc} 
    \toprule
    Attack Method & \multicolumn{4}{c}{Trigger Characteristics} \\  
    \cmidrule(lr){2-5} 
     & Visibility & Semanticity & Specificity & Compatibility \\  
    \midrule
    BadNets~\cite{gu2019badnets} & \fullcirc & \emptycirc & \emptycirc & \emptycirc \\
    Blended~\cite{Blended} & \emptycirc & \fullcirc & \emptycirc & -- \\
    BPP~\cite{wang2022bppattack}  & \emptycirc & \emptycirc & \fullcirc & -- \\
    Input-Aware~\cite{nguyen2020input}  & \fullcirc & \emptycirc & \fullcirc & \emptycirc \\
    SIG~\cite{SIG}  & \emptycirc & \emptycirc & \emptycirc & -- \\
    SSBA~\cite{ssba}  & \emptycirc & \emptycirc & \fullcirc & -- \\
    TrojanNN~\cite{trojannn}  & \fullcirc & \emptycirc & \emptycirc & \emptycirc \\
    WaNet~\cite{nguyen2021wanet}  & \emptycirc & \emptycirc & \fullcirc & -- \\
    \midrule 
    VSSC (Ours) & \fullcirc & \fullcirc & \fullcirc & \fullcirc \\
    \bottomrule
\end{tabular}}
\end{table}

\section{Investigation of the characteristics of backdoor triggers}
\label{sec: investigating trigger characteristics}
To leverage the advantages of digital attacks to address the challenges in existing physical attacks, we aim to design an ideal backdoor attack that performs well in both digital and physical scenarios.
From the attacker’s perspective, the ideal backdoor attack should fulfill three desired goals, including:

\begin{itemize}
    \item \textbf{Effectiveness}: The backdoor should be successfully injected and capable of being activated with a high attack success rate (ASR) during the inference stage. 
    \item \textbf{Stealthiness}: The trigger in the poisoned image should be stealthy to human inspection. 
    \item \textbf{Robustness}: The backdoor effect should be well maintained under visual distortions in both digital and physical scenarios.  
\end{itemize}

\begin{figure}[!t]
\centering
\includegraphics[width=0.7\columnwidth]{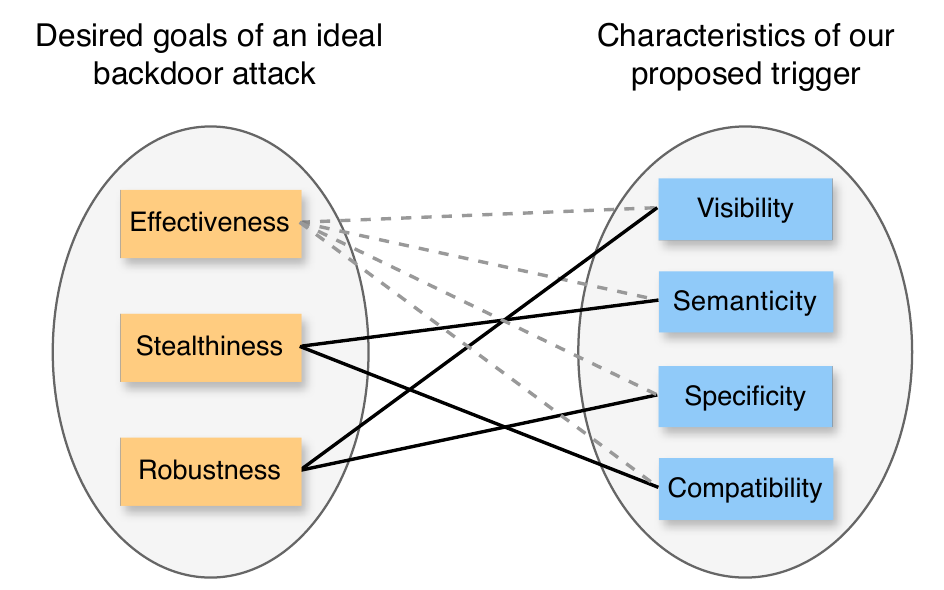}
\caption{Connections between desired goals of an ideal backdoor attack and characteristics of our proposed trigger. A solid line indicates that the characteristic directly contributes to the goal, while a dashed line signifies a collaborative effort. Please refer to Section \ref{sec: investigating trigger characteristics} for the detailed illustration of these connections.}
\label{fig:two_step}
\end{figure}

As depicted in Table \ref{tab: trigger characteristics}, we examine four essential characteristics of several representative backdoor triggers\footnote{We follow the categorizations of trigger characteristics in~\cite{wu2023attack_survey}.}, to investigate the connections between these characteristics and the above goals. 
These connections are illustrated in Figure \ref{fig:two_step}.
Note that effectiveness is influenced by multiple factors (\eg, poisoning ratio, original dataset, \comment{model architecture} and training algorithm), not solely by the trigger, thus we do not investigate its connections to trigger characteristics.

\textbf{Robustness-related trigger characteristics.}
Early backdoor attacks often assume that the triggers across different poisoned images are consistent in appearance or location, \ie, agnostic to the victim image. Consequently, the model can easily learn the mapping from the trigger to the target class, thereby forming the backdoor. However, the commonality among poisoned samples in sample-agnostic triggers leads to a backdoor that necessitates high consistency in the trigger. This makes the trigger vulnerable to visual distortions and easily recognized by GradCam~\cite{selvaraju2017grad} or backdoor defense methods like Neural Cleanse~\cite{nc}.
The characteristic of \textbf{sample-specific} can ensure a greater variety in the appearance of triggers. 
Some recent works propose sample-specific triggers, such as SSBA~\cite{ssba} and Input-Aware~\cite{nguyen2020input}. However, the robustness under visual distortions on testing images still has not been seriously considered by these attacks. Previous work \cite{li2020rethinking} has empirically revealed that these triggers are sensitive to changes like trigger locations or intensities during the inference stage. 
For backdoor attacks entirely implemented in the physical scenario, the robustness against distortions has been considered~\cite{Wenger_2021_CVPR,wu2022just}. 
Inspired by these considerations, we study visual distortions during image processing (\eg, blur, compression, and noise) in the digital domain and environment variations (\eg, lights, shooting angles, and distances) in the physical domain.
It demonstrates that several advanced attacks with invisible triggers are sensitive to visual distortions since the trigger magnitude is small. In this case, a sufficiently large magnitude is desired to ensure that the trigger is not destroyed during the image processing, making a \textbf{visible} trigger more advantageous
Additionally, the trigger added by the generative model can simulate distortions under different environmental conditions, thereby making the attack more robust.

\textbf{Stealthiness-related trigger characteristics.} 
Several visible triggers were employed in early backdoor attacks, such as a patch used in BadNets~\cite{gu2019badnets} and a semi-transparent stamp in TrojanNN~\cite{trojannn}. To ensure visual stealthiness, more recent works focus on designing invisible triggers through alpha blending (\eg, Blended~\cite{Blended}), image steganography (\eg, SSBA~\cite{ssba} and LSB~\cite{LSB}), slight spatial transformations (\eg, WaNet~\cite{nguyen2021wanet}), or invisible adversarial perturbations (\eg, LSB~\cite{LSB}). In addition, given the visible and non-semantic trigger, some works attempt to enhance stealthiness by placing the trigger at an inconspicuous location or reducing its size (\eg, Input-Aware~\cite{nguyen2020input}). In contrast, a visible but \textbf{semantic} and \textbf{compatible} trigger is obviously more stealthy. A few attempts, such as~\cite{bagdasaryan2020how_to_backdoor_federated_learning, wenger_2022_nips}, have used specific attribute-containing images(\eg, a car with a racing stripe) as poisoned images without image manipulation. However, since no image manipulation occurs, the attacker's flexibility is limited—the number and diversity of selected poisoned images are restricted by the original dataset, which prevents the attacker from flexibly controlling. This limitation may explain why attacks with visible, semantic, and compatible triggers have not been well studied in this field, although such triggers have high stealthiness in both digital and physical scenarios.

In summary, from the above analysis, we conclude that 
to simultaneously achieve the above attack goals, a desirable trigger should be \textbf{visible}, \textbf{semantic},
\textbf{sample-specific}, and \textbf{compatible}. 
These characteristics collectively ensure the effectiveness, stealthiness, and robustness of the backdoor attack.

\section{Methodology}
\subsection{Problem Formulation}

\subsubsection{Threat model}
\begin{figure*}[t]
\captionsetup{font=small}
    \centering
    \includegraphics[width=1\textwidth]{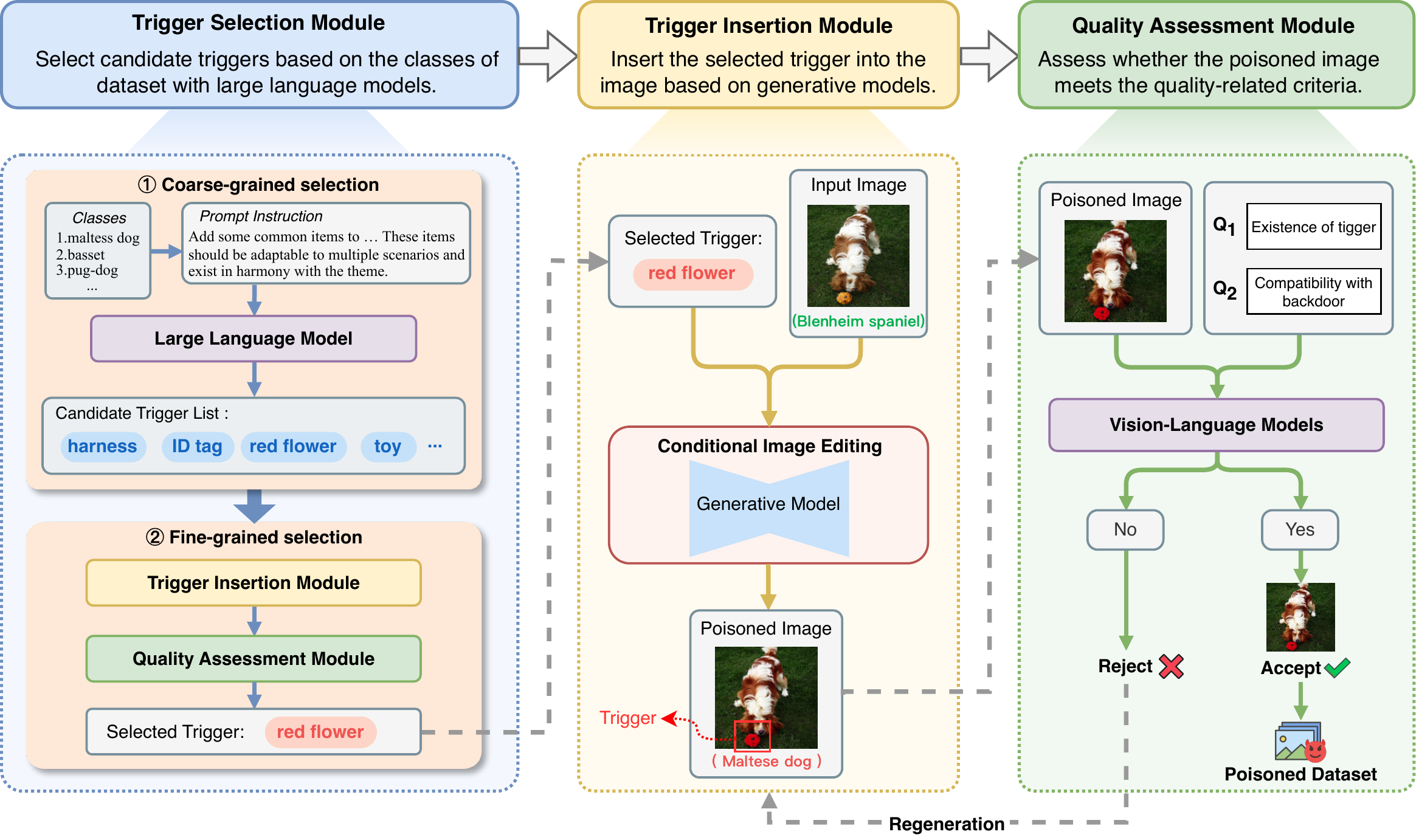}
    \caption{ Overview of our proposed method. The process of generating poisoned samples includes three fundamental modules.
    To synthesize a poisoned dataset, a text trigger is first selected using the trigger selection module, then inserted into a benign image using the trigger insertion module, and finally evaluated using the quality assessment module to assess insertion effects. Only poisoned images meeting quality criteria are used to form the poisoned dataset, while low-quality images need to be regenerated. 
    }
    \label{fig:overview}
\end{figure*}

As shown in Figure \ref{fig:overview}, a complete procedure of a backdoor attack consists of three stages, including poisoned dataset generation; training a model based on the poisoned dataset; activating the backdoor in the model through a poisoned testing image with a trigger. We consider the \textit{threat model of data poisoning based attack}, where the attacker can only manipulate the training dataset and the testing image at the inference stage, while the model training stage cannot be accessed. 

\subsubsection{Notations}
We denote the image classifier as $f_{\btheta}: \mathcal{X} \rightarrow \mathcal{Y}$, with $\btheta$ being the model parameter, $\X$ being the input space and $\Y$ being the output space. The benign training dataset is denoted as $\D = \{(\x^{(i)}, y^{(i)})\}_{i=1}^n$, with $\x^{(i)} \in \X$ being the $i$-th benign image and $y^{(i)}$ being its ground-truth label. A few benign images indexed by $\mathcal{P} \in \{1, 2, \ldots, n\}$ will be selected to generate the poisoned images $\x_{\epsilon}$ by inserting a particular trigger $\beps$, and their labels will be changed to the target label $t$. 
The poisoned images and the remaining benign images form the poisoned training dataset $\D_p = \{(\x_{\epsilon}^{(i)}, t)_{i \in \mathcal{P}}, (\x^{(i)}, y^{(i)})_{i \notin \mathcal{P}}\}_{i=1}^n$. The poisoning ratio is denoted as $r = |\mathcal{P}|/n$.

\subsection{Backdoor Attack with Visible, Semantic, Sample-specific, and Compatible (VSSC) Trigger}
The process of generating poisoned samples consists of three fundamental modules: trigger selection module used to select a text trigger, trigger insertion module used to insert the trigger into a benign image, and quality assessment module used to evaluate the insertion quality. In this section, we first introduce the three fundamental modules, then describe the pipeline of the VSSC attack in detail.

\subsubsection{Trigger selection module}

Selecting a compatible trigger is crucial to ensuring the effectiveness and stealthiness of the attack. For different datasets, triggers should be flexibly selected based on the scenes they contain, which means it is necessary to consider the natural distribution of objects in the real world. 
In most attack methods that use semantic triggers, especially physical attacks, triggers are often selected manually in a heuristic manner which is highly subjective and has a limited selection range. 
To address this issue, we propose a trigger selection module based on large language models (LLMs), leveraging LLM's prior knowledge of the real world. This renders the trigger selection process more automated, mitigates the arbitrariness of manually selected triggers, and broadens the range of possible triggers.
The trigger selection module is designed with two sections: coarse-grained selection and fine-grained selection, which filter triggers from semantic and visual perspectives respectively. 

\textbf{Coarse-grained selection}. During the coarse-grained selection, triggers are filtered semantically to ensure compatibility within the image context. LLMs like GPT-4~\cite{openai2023gpt4} possess the capability to understand semantic information akin to humans. We first leverage their expansive prior knowledge to broadly filter out a list of candidate triggers. Specifically, for any given dataset, we provide the LLM with class names and prompt instructions, which outline the attacker’s semantic criteria for trigger selection, for example, ``Find 10 common objects which look natural with $[classes]$''. Some examples are shown in supplementary material I-D.

\textbf{Fine-grained selection}. This section is designed to filter out triggers with better insertion effects from a visual perspective.
This section is necessary because some candidate triggers are not well-adapted to specific scenes, or the current image editing technology is limited, resulting in poor insertion quality. 
Fine-grained selection based on actual visual effects complements the coarse-grained semantic filtering process. In this section, we randomly select some benign images to form a trigger evaluation set. 
Each candidate trigger from the list is then inserted into the images of the evaluation set using our trigger insertion module, followed by a quality assessment module to evaluate the insertion effects. 
We introduce the Insertion Success Rate (ISR) as a metric to measure the efficacy of the generation of text triggers. ISR is defined as the proportion of evaluation images that meet the quality criteria of the quality assessment module.
We set a threshold for ISR, triggers surpassing this threshold are considered qualified and can be used in the subsequent attack processes. The entire trigger selection module is fully automated.

\subsubsection{Trigger insertion module}
In previous digital backdoor attacks~\cite{Blended, han2022physical,Wenger_2021_CVPR}, semantic triggers are typically integrated by pasting, which often results in unnatural visual effects and a lack of trigger diversity. On the other hand, most physical attacks require manually capturing poisoned images, which is labor-intensive and time-consuming. 
Consequently, we propose an automated trigger insertion module using conditional image editing methods based on generative models. This module leverages the visual understanding capability of the generative model to replace manual editing, automatically selects the appropriate location, and generates triggers with suitable appearance, ensuring the compatibility and sample-specificity of the semantic trigger. 
Furthermore, the generative model can simulate environmental variations such as lighting and shooting angles based on the image content, ensuring the robustness of the backdoor attack.
The insertion process is detailed as follows:
\textbf{Firstly}, for a specific image, we compose a prompt instruction using the selected text trigger (\eg, ``\textit{red flower}'' in Figure \ref{fig:overview}) to guide the model in editing the benign image. The format of this prompt depends on the specific image editing method. 
\textbf{Secondly}, the prompt instruction and benign image are fed into a pre-trained generative model, generating a poisoned image  that contains a visual object corresponding to the text trigger.
This module is not bound to any specific image editing technology, instead, it provides a new perspective on trigger insertion. In our experiments, we employ various image editing techniques for different tasks to demonstrate this module's flexibility, including \cite{mokady2022null, geng2024instructdiffusion, ye2023ip-adapter, zhang2023adding_controlnet}.
As image editing techniques evolve, the quality and diversity of VSSC triggers will also be enhanced. This flexibility allows our method to easily integrate with cutting-edge technologies, ensuring its long-term value.

\subsubsection{Quality assessment module}
Current image editing technologies based on generative models are not completely faithful, and the generation results are subject to some randomness. To ensure the effectiveness of the VSSC attack, it is crucial to evaluate the quality of triggers from the perspectives of successful integration and compatibility with the surrounding visual content. Consequently, we introduce a quality assessment module.
This module treats the evaluation of trigger generation quality as a Visual Question Answering (VQA) task, leveraging the visual understanding capability of vision-language models (VLMs) to discern whether the generated trigger meets quality criteria.

We first set some quality-related criteria, such as ``$[trigger]$ exists in the image'' and ``$[trigger]$ is compatible with the background''. The VLM accepts an image as input and responds whether the image meets these criteria. If all answers are ``yes'', the image is considered a qualified poisoned sample. Similar to the trigger insertion module, the specific VLM used in this module is not fixed,  such as GPT-4~\cite{openai2023gpt4} and LLaVa~\cite{liu2023llava}.

This module can individually check the quality of each poisoned sample, ensuring the effectiveness and stealthiness of the backdoor attack without the incorporation of human supervision, thereby reducing labor expenditure.

The pipeline of generating a poisoned dataset with VSSC triggers can be divided into the following stages:

\begin{itemize}
    \item \textbf{Stage \uppercase\expandafter{\romannumeral1}:} \textbf{Poisoned dataset generation.}
    \textbf{Firstly}, we provide all class names and trigger selection criteria to the trigger selection module. After coarse-grained selection and fine-grained selection, we obtain a text trigger that can be used for the attack (``\textit{red flower}'' in Figure \ref{fig:overview}).
    \textbf{Secondly}, we use the trigger insertion module to add selected text triggers to the extracted benign image to get a poisoned image $x_\epsilon$. 
    \textbf{Finally}, the quality assessment module is employed to evaluate the quality of $\x_{\epsilon}$. If the poisoned image passes the quality assessment, it is labeled as the target class $t$ (\eg, ``\textit{Maltese dog}'' in Figure \ref{fig:overview}) to obtain a poisoned training data pair $(\x_{\epsilon}, t)$. 
    Otherwise, we randomly adjust the arguments of the trigger insertion module and regenerate until a qualified image is obtained. 
    For each image, multiple attempts can be made. If a qualified poisoned image is not generated after the maximum attempts, this image is discarded. 
    We attempt to insert triggers into a selection of benign images at a proportion higher than the poisoning ratio. If the actual poisoning rate still falls short of the target after iterating through all these images, the actual poisoning rate will be lower than the set value. 
    This situation only occurs in a few cases which we will discuss in our analysis.

    \item \textbf{Stage \uppercase\expandafter{\romannumeral2}:}\textbf{ Model training. }
    Given the generated poisoned training dataset $\D_p$, the model training will be conducted by the user (rather than the attacker) to obtain the image classifier $f_{\btheta}$. 

    \item \textbf{Stage \uppercase\expandafter{\romannumeral3}:} \textbf{Inference.}
    During the inference stage, the attacker can employ the same text trigger and trigger generation pipeline to edit the benign inference image, thus creating the poisoned inference image $\x_{\epsilon}$, to activate the backdoor in $f_{\btheta}$, \ie, $f_{\btheta}(\x_{\epsilon}) = t$. 
\end{itemize}

\begin{figure*}[!ht]
    \centering
    \subfloat[ImageNet-Dogs (Target class: Maltese dog)]{\includegraphics[width=0.49\textwidth]{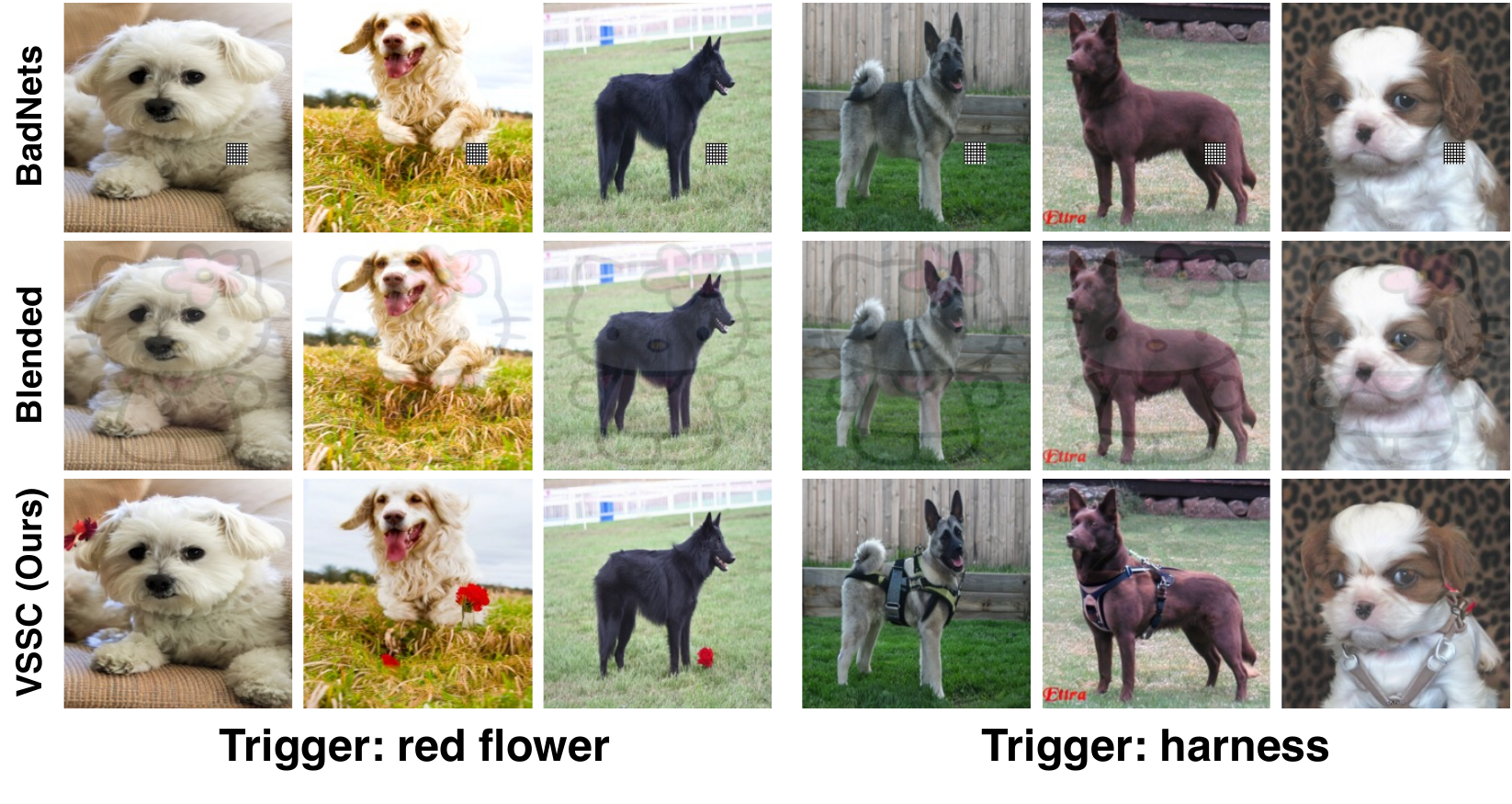}}
    \label{fig:fg_dog}
    \subfloat[FOOD-11 (Target class: Bread)]{\includegraphics[width=0.49\textwidth]{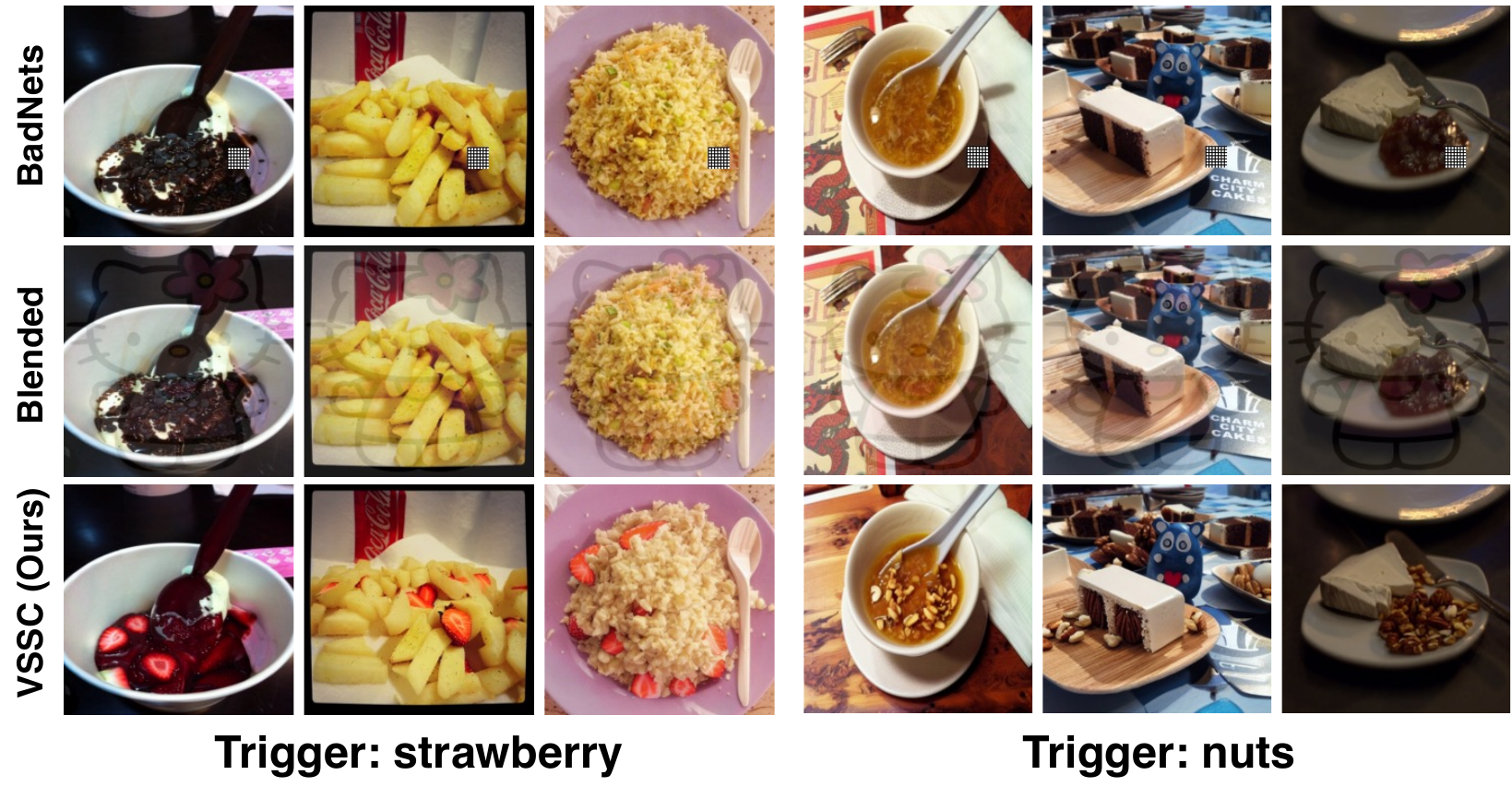}}
    \caption{Poisoned samples generated by different attacks for the image classification task. BadNets uses a black-and-white grid as the trigger pattern, while Blended uses an image. For VSSC attack, we demonstrate two semantic triggers used in our experiments for each dataset. Although all these poisoned images are successfully classified as the target label, VSSC triggers in them are the most stealthy and compatible.}
   \label{fig:trigger}
\end{figure*}

\textbf{Remark on characteristics of the generated triggers}. 
Several examples of poisoned images are illustrated in Figure \ref{fig:trigger}. The semantic trigger has been effectively integrated into the benign image, with its shape, size, and location adjusted to ensure compatibility with the remaining visual content in the edited image. 
Consequently, we name the proposed trigger as a \textbf{Visible}, \textbf{Semantic}, \textbf{Sample-specific}, and \textbf{Compatible trigger} (VSSC trigger).
As demonstrated in Table \ref{tab: trigger characteristics}, our trigger is the only one that satisfies visible, semantic, sample-specific, and compatible properties simultaneously, which aligns with the attack goals introduced in Section \ref{sec: investigating trigger characteristics}. Furthermore, thanks to the automated trigger generation pipeline, VSSC attack can achieve an efficient physical attack without the need for heavy manpower.

\section{Experiments}

\subsection{Evaluation Tasks and Scenarios}

We conduct a comprehensive evaluation of the proposed VSSC trigger across three tasks: \textbf{image classification} (Section \ref{sec:classification}), \textbf{object detection} (Section \ref{sec:detection}), and \textbf{face verification}(Section \ref{sec:face}). 
For each task, we design three scenarios to evaluate the attack performance of the VSSC trigger, each offering a distinct perspective.

\begin{itemize}
    \item \textbf{Digital scenario:} 
    This scenario aims to evaluate the attack performance under the fundamental digital setup, in which the testing poisoned images are solely poisoned in the digital domain, consistent with the way adopted in previous digital attacks~\cite{nguyen2021wanet,ssba,SIG}.
    \item \textbf{Digital-to-physical (D2P) scenario:} During the actual implementation of backdoor attacks in the physical domain, visual distortion may be unintentionally introduced. To evaluate the attack performance against such distortions, in this scenario, we simulate visual distortions by printing testing poisoned images and recapturing them, following the same way adopted in  ~\cite{li2021backdoor}.
    \item \textbf{Physical scenario:} 
    This scenario aims to evaluate the attack performance of backdoor attacks in the real world, which poses greater challenges than the first two scenarios, including more complex distortions and a requirement for a real-world counterpart.
    To create this scenario, we utilize corresponding objects as triggers and place them in the physical scenes, followed by taking photos as the testing poisoned images.
    
\end{itemize}

\subsection{Evaluations on Image Classification Task} 
\label{sec:classification}
\begin{table*}[!t]
\caption{Attack performance of different attacks on ImageNet-Dogs and FOOD-11 with 5\% poisoning ratio in the digital scenario. For the ImageNet-Dogs dataset, VSSC$_1$ and VSSC$_2$ respectively denote the experiments conducted using the triggers ``red flower'' and ``harness'', while corresponding to ``nuts'' and ``strawberry'' for the FOOD-11 dataset.}
\label{tab:attack_0.05}
\begin{center}
\renewcommand\arraystretch{1}
\resizebox{\textwidth}{!}{%
\begin{tabular}
{l ccc ccc ccc ccc}
\toprule
Model $\rightarrow$& \multicolumn{6}{c}{\textbf{ResNet-18}} & \multicolumn{6}{c}{\textbf{VGG19-BN}}\\ 
\cmidrule(lr){2-7}  \cmidrule(lr){8-13}

 Dataset $\rightarrow$& \multicolumn{3}{c}{ImageNet-Dogs} & \multicolumn{3}{c}{FOOD-11} & \multicolumn{3}{c}{ImageNet-Dogs} & \multicolumn{3}{c}{FOOD-11}\\
 \cmidrule(lr){2-4}  \cmidrule(lr){5-7} \cmidrule(lr){8-10}  \cmidrule(lr){11-13} 
 Attack $\downarrow$& C-Acc(\%)& ASR(\%)& R-Acc(\%)& C-Acc(\%)& ASR(\%)& R-Acc(\%) & C-Acc(\%)& ASR(\%)& R-Acc(\%)& C-Acc(\%)& ASR(\%)& R-Acc(\%)\\ \midrule
BadNets~\cite{gu2019badnets}               & 86.93 & 99.57 & 0.43  & 83.35 & 99.64 & 0.36  & 91.60 & 100.00 & 0.00  & 85.95 & 99.93 & 0.07                 \\
Blended~\cite{Blended}                & 88.13 & 97.29 & 2.57  & 84.46 & 93.48 & 5.90  & 91.33 & 98.71  & 1.14  & 86.33 & 94.75 & 4.82                 \\
BPP~\cite{wang2022bppattack}                    & 71.73 & 91.29 & 6.14  & 71.81 & 96.45 & 3.03  & 78.67 & 25.14  & 56.57 & 74.02 & 10.10 & 66.62                \\
Input-Aware~\cite{nguyen2020input}            & 86.00 & 99.71 & 0.29  & 80.93 & 98.53 & 1.37  & 89.60 & 99.43  & 0.29  & 82.39 & 96.41 & 3.39                 \\
SIG~\cite{SIG}                    & 87.60 & 83.57 & 15.57 & 84.78 & 95.31 & 4.34  & 90.80 & 84.00  & 15.14 & 84.84 & 90.48 & 8.83                 \\
SSBA~\cite{ssba}                   & 89.07 & 99.43 & 0.57  & 84.05 & 96.12 & 3.49  & 92.67 & 99.29  & 0.57  & 86.68 & 97.65 & 2.12                 \\
TrojanNN~\cite{trojannn}               & 85.33 & 99.14 & 0.71  & 83.44 & 96.97 & 2.84  & 91.47 & 37.14  & 57.57 & 85.63 & 24.02 & 67.34                \\
WaNet~\cite{nguyen2021wanet}                  & 66.53 & 99.57 & 0.14  & 67.08 & 34.91 & 54.76 & 75.60 & 97.14  & 1.71  & 60.73 & 94.04 & 4.30                 \\
\midrule
\rowcolor[HTML]{E6E6E6} 
VSSC$_1$ (Ours)     & 89.20 & 95.65 & 3.34  & 84.29 & 93.91 & 4.89     & 92.80 & 97.16  & 2.34  & 86.30 & 97.22 & 2.24                   \\
\rowcolor[HTML]{E6E6E6} 
VSSC$_2$ (Ours)    & 90.13 & 89.69 & 7.69  & 84.37 & 91.15 & 5.81     & 92.53 & 92.15  & 7.08  & 87.08 & 94.11 & 3.91  \\                   
 \bottomrule
\end{tabular}
}
\end{center}
\end{table*}

\subsubsection{Experimental settings}
For the classification task, backdoor attacks aim to add samples with specific triggers to the training set, causing the model to produce misclassification results for samples with triggers during the inference stage.
\\
\textbf{{Dataset and models.}}
We use two high resolution datasets: ImageNet-Dogs~\cite{li2021contrastive}, a 20,250-image subset of ImageNet~\cite{deng2009imagenet} featuring 15 dog breeds, and FOOD-11~\cite{singla2016food}, containing 16,643 images across 11 food categories. These two datasets are used to validate the effectiveness of VSSC triggers. The image size is 3$\times$224$\times$224. We use ResNet-18~\cite{he2016deep} and VGG19-BN~\cite{simonyan2015very} for both datasets. The main results are based on ResNet-18, with more experiments with VGG19-BN in supplementary material II-A.

\textbf{{Implementation details.}}
For the image classification task, the ISR threshold is set to 0.5 during the fine-grained selection. From the final selected triggers, we choose ``\textit{red flower}'' and ``\textit{harness}'' for the ImageNet-Dogs dataset and ``\textit{nuts}'' and ``\textit{strawberry}'' for the FOOD-11 dataset for experimental demonstration.  
The poisoning ratio is set to 5\% and 10\%, with a target label $y_t=0$ assigned to all datasets.
Further details can be found in supplementary material I-A3.

\textbf{{Baseline attacks.}}
We compare the proposed VSSC attack with 8 popular attack methods, including BadNets~\cite{gu2019badnets}, Blended~\cite{Blended}, BPP~\cite{wang2022bppattack}, Input-Aware~\cite{nguyen2020input}, SIG~\cite{SIG}, WaNet~\cite{nguyen2021wanet}, SSBA~\cite{ssba} and TrojanNN~\cite{trojannn}. These baseline attacks are implemented by BackdoorBench~\cite{wu2024backdoorbench}. 

\textbf{{Evaluation metrics.}}
We evaluate attack effectiveness using Attack Success Rate (ASR), Clean Accuracy (C-Acc), and Robust Accuracy (R-Acc). Specifically, ASR measures the proportion of poisoned samples misclassified as the target label. C-Acc is defined as the accuracy of benign data. R-Acc is defined as the ratio of poisoned samples being classified as their original classes. For an effective backdoor attack, the C-Acc should be close to the clean model, and the ASR should be as high as possible while maintaining the lowest possible R-Acc. Note that in the classification task, the sum of ASR and R-Acc will not exceed 1.

\begin{figure*}[htbp]
    \subfloat[Dogs with ``red flower'']{\includegraphics[width=\columnwidth]{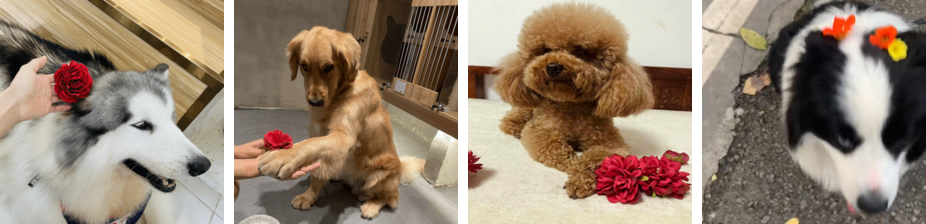}}
    \hspace{3mm}
    \subfloat[Dogs with ``harness'']{\includegraphics[width=\columnwidth]{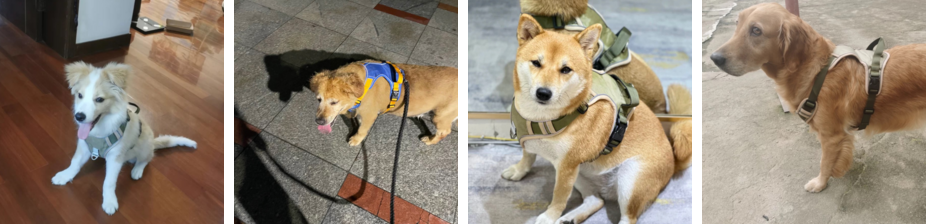}} 
    \\
    \subfloat[Food with ``nuts'']{\includegraphics[width=\columnwidth]{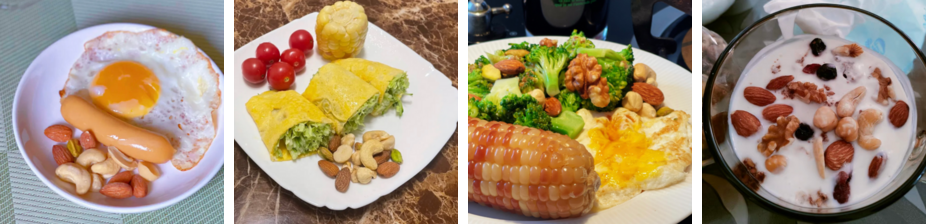}} 
    \hspace{3mm}
    \subfloat[Food with ``strawberry'']{\includegraphics[width=\columnwidth]{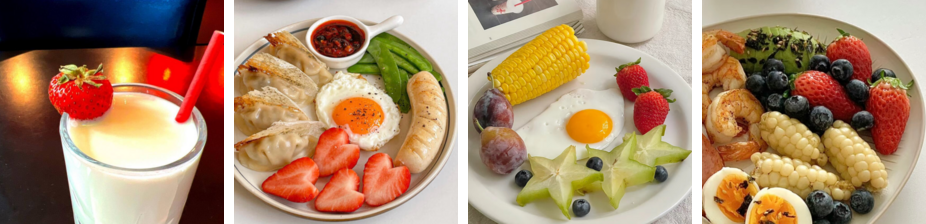}}
    
\caption{Examples of manually captured photos in the image classification task.}
\label{fig:exp_ood}
\end{figure*}

\subsubsection{Effectiveness in digital scenario}

As shown in Table \ref{tab:attack_0.05}, our attack achieves an ASR up to 97.16\% on the ImageNet-Dogs dataset and 97.22\% on the FOOD-11 dataset with a poisoning rate of 5\%.  Although we use triggers with actual semantics and thus have representations closer to the original images in feature space, 
our attack method still demonstrates solid effectiveness in digital scenarios, comparable to other baseline attacks. It is worth emphasizing that VSSC attack has a minimal impact on C-Acc, causing less than 2\% C-Acc decrease across all backbones and datasets, outperforming all other attacks. The sample-specific characteristic of VSSC triggers introduces a greater diversity in the inserted triggers, enabling the model to learn the common features of the triggers during training and to distinguish them from key features in normal image classification tasks. This makes the detection of VSSC attacks more challenging.
Results with a ratio of 10\% are in supplementary material II-A1. 

\begin{table}[!ht]
\centering
\caption{Attack performance for image classification task in the digital-to-physical scenario. The backbone is ResNet-18 and the poisoning ratio is set to 5\%. For ImageNet-Dogs and FOOD-11, the triggers are set to ``red flower'' and ``nuts''.}
\label{tab:recapture}
\resizebox{\columnwidth}{!}{%
\begin{tabular}{lcccccc}
    \toprule
    \multirow{2}{*}{\makecell{Attacks}} & \multicolumn{3}{c}{ImageNet-Dogs} & \multicolumn{3}{c}{FOOD-11} \\
    \cmidrule(lr){2-4}\cmidrule(lr){5-7}
    & C-Acc(\%) &  ASR(\%)  & R-Acc(\%)           & C-Acc(\%) & ASR(\%)  & R-Acc(\%) \\
    \midrule
                BadNets~\cite{gu2019badnets}    & 88.10 & 0.00  & 80.95 & 40.00 & 6.67  & 43.33 \\
Blended~\cite{Blended}     & 83.33 & 26.19 & 59.52 & 63.33 & 33.33 & 33.33 \\
BPP~\cite{wang2022bppattack}         & 71.43 & 9.52  & 54.76 & 73.33 & 3.33  & 43.33 \\
Input-Aware~\cite{nguyen2020input} & 83.33 & 0.00  & 71.43 & 76.67 & 0.00  & 56.67 \\
SIG~\cite{SIG}         & 90.48 & 9.52  & 54.76 & 70.00 & 33.33 & 26.67 \\
SSBA~\cite{ssba}        & 80.95 & 0.00  & 76.19 & 63.33 & 6.67  & 53.33 \\
TrojanNN~\cite{trojannn}    & 78.57 & 7.14  & 69.05 & 50.00 & 20.00 & 50.00 \\
WaNet~\cite{nguyen2021wanet}       & 33.33 & 76.19 & 16.67 & 50.00 & 20.00 & 43.33 \\ \midrule
\rowcolor[HTML]{E6E6E6} 
VSSC (Ours) & 92.86 & 97.62 & 2.38  & 66.67 & 93.33 & 0.00 \\
    \bottomrule
\end{tabular}
}
\end{table}

\subsubsection{Effectiveness in digital-to-physical scenario}

As illustrated in Table \ref{tab:recapture}, VSSC attack achieves the highest ASR in the digital-to-physical scenario. On the ImageNet-Dogs dataset, the ResNet-18 model poisoned by VSSC triggers achieves an ASR of 97.62\% in the digital-to-physical scenario with a poisoning ratio of 5\%, and 93.33\% on the FOOD-11 dataset. The VSSC trigger achieves the highest ASR among all attacks on both datasets.
In contrast, none of the baseline attacks maintain effectiveness with an acceptable C-Acc, primarily because visual distortions are not accounted for during the design of their triggers, rendering them susceptible in practical deployments. In addition, VSSC attack has the lowest R-Acc on both datasets, which means that the backdoor model does not tend to classify poisoned samples into their original classes, also reflecting the effectiveness of the backdoor attack.
Figure~\ref{fig:gradcam} presents the Grad-CAM~\cite{selvaraju2017gradcam} visualizations of images with the VSSC trigger and two visible triggers. When subjected to visual distortions, the attention regions of the backdoor model for both BadNets and TrojanNN shift, whereas the attention region associated with the VSSC trigger remains constant. This indicates that the VSSC trigger exhibits remarkable robustness against visual distortions.

\begin{table*}[!ht]
\centering
\caption{Attack successful rate (\%) of VSSC triggers in the physical scenario under various poisoning ratios for the image classification task (tested on 100 images), where `Pratio' denotes the poisoning ratio.}
\label{tab:physical}
\resizebox{0.7\textwidth}{!}{%
\begin{tabular}{@{}cccccccccc@{}}
\toprule
                         & \multicolumn{1}{l}{Backbone $\rightarrow$} & \multicolumn{4}{c}{ResNet-18} & \multicolumn{4}{c}{VGG19-BN}   \\ \cmidrule(lr){3-6} \cmidrule(lr){7-10}
Dataset $\downarrow$                 & Pratio  $\rightarrow$                     & 5\%   & 10\%  & 20\%  & 30\%  & 5\%   & 10\%  & 20\%  & 30\%   \\ \midrule
\multirow{2}{*}{ImageNet-Dogs} & Red Flower & 65.00 & 79.00 & 83.00 & 89.00 & 65.00 & 74.00 & 83.00 & 88.00 \\
                         & Harness                      & 61.00 & 70.00 & 79.00 & 83.00 & 68.00 & 74.00 & 83.00 & 85.00  \\ \midrule
\multirow{2}{*}{FOOD-11} & Nuts                         & 60.00 & 71.00 & 80.00 & 81.00 & 59.00 & 68.00 & 75.00 & 100.00 \\
                         & Strawberry                   & 62.00 & 69.00 & 72.00 & 83.00 & 61.00 & 70.00 & 72.00 & 100.00 \\ \bottomrule
\end{tabular}%
}
\end{table*}

\subsubsection{Effectiveness in physical scenario}
For experiments in the physical scenario, we collect 100 images with the trigger entity for each trigger. 
Figure \ref{fig:exp_ood} shows some examples of these manually captured images.
Experiments are conducted under four poisoning ratios, ranging from 5\% to 30\%, as shown in Table \ref{tab:physical}. The VSSC trigger can implement successful attacks in the physical scenario, even at very low poisoning ratios. At a poisoning ratio of 5\%, most triggers achieve an ASR of over 60\%, which is notably high when utilizing a category of objects with specific semantics as triggers. With the increase in the poisoning ratio, the ASR correspondingly improves. At a poisoning ratio of 30\%, the ASR of the ``\textit{strawberry}'' trigger even reaches 100\%. As a trigger deployed in the digital domain, the VSSC trigger still achieves a remarkably high attack success rate in the physical scenario, demonstrating its effectiveness in the real world.

\begin{figure*}[!h]
\centering    
\includegraphics[width=\textwidth]{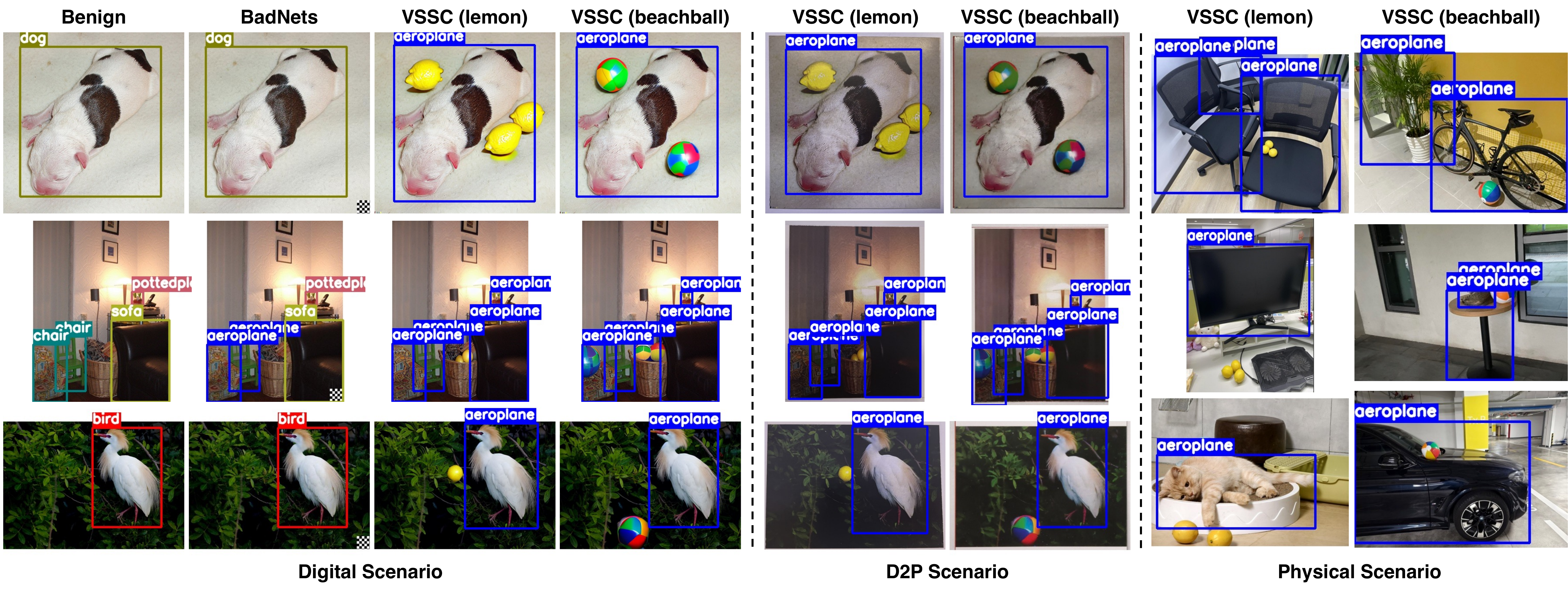}
\caption{Poisoned samples created by BadNets and VSSC for the object detection task. For the VSSC attack, we show ``\textit{lemon}'' and ``\textit{beachball}'' as triggers in three different scenarios, where D2P means the digital-to-physical scenario. We only use the results under Global Misclassification Attack (GMA) for demonstration.}     
\label{fig:detection_example}     
\end{figure*}

\subsection{Evaluations on Object Detection Task}
\label{sec:detection}

\subsubsection{Experimental settings}
In the object detection task, we introduce two different attack types mentioned in BadDet~\cite{chan2022baddet}, \textbf{Object Disappearance Attack (ODA)} and \textbf{Global Misclassification Attack (GMA)}. ODA aims to make all bounding boxes in poisoned images vanish, while GMA aims to misclassify all bounding boxes in poisoned images as the target class. 
We only introduce fundamental settings in the following sections. For further details, please refer to supplementary material I-B.

\textbf{{Dataset and models.}}
The dataset used in the object detection task is PASCAL VOC 07+12~\cite{everingham2010voc}, which contains 19,352 images of 20 classes of objects.

To evaluate the effectiveness of the VSSC attack across more diverse object detection models, we select two different types of detectors, YOLOv4~\cite{bochkovskiy2020yolov4} with the CSPDarkNet53 feature extractor and Faster R-CNN~\cite{ren2015frcnn} with the ResNet-50 backbone. YOLOv4 is a widely used anchor-based one-stage detection model, while Faster R-CNN is a classic two-stage detection model. The input size of YOLOv4 is 3 $\times$ 416 $\times$ 416, and the input size of Faster R-CNN is 3 $\times$ 600 $\times$ 600. 
Experiment results shown in this section are based on YOLOv4, more experiments with Faster R-CNN and some training details can be found in supplementary material I-B.

\textbf{{Implementation details.}}
In ODA, the width and height of the poisoned bounding box are set to 0, i.e., $g_{y}([a, b, w, h, c])=[a, b, 0, 0, c]$. In GMA, the class of the poisoned bounding box is set to the target class, i.e., $g_{y}([a, b, w, h, c])=[a, b, w, h, y_t]$, we randomly select ``aeroplane'' as the target class with a target label $y_t=0$ assigned. $(a, b)$ denotes the coordinate of the top-left point of the bounding box.
The poisoning ratio is set to 10\% and 20\%. Among all the text triggers filtered out by the trigger selection module, ``\textit{lemon}'' and ``\textit{beachball}'' are chosen for demonstration in our experiments.

\textbf{{Baseline attacks.}}
From attacks designed for image classification tasks, we selected a subset of them whose triggers can be extended to object detection tasks as baseline attacks. We only use their triggers for data poisoning without controlling the training process. Baseline attacks used in the object detection task include 
BadDet~\cite{chan2022baddet}, Blended~\cite{Blended}, BPP~\cite{wang2022bppattack}, SIG~\cite{SIG}, SSBA~\cite{ssba}, and WaNet~\cite{nguyen2021wanet}.

\textbf{{Evaluation metrics.}}
For the object detection task, we still use the same evaluation metrics as the image classification task, ASR, C-Acc, and R-Acc. The definition of ASR in this section varies according to different attack types. For ODA, ASR is defined as the proportion of original bounding boxes that successfully vanish. In GMA, ASR is defined as the proportion of non-target class bounding boxes that are misclassified as the target class. C-Acc is defined as mAP on benign images, which represents the mean of average precision (AP) of each class. AP represents the area under the precision-recall curve for bounding boxes with a sufficient confidence level. Our experiments use mAP on an IoU threshold of 0.5 (mAP@0.5). R-Acc is defined as mAP on poisoned images.
ASR and R-Acc usually do not reach a high level simultaneously, but there is no additive relationship between them.

\begin{table*}[]
\centering
\caption{Attack performance of VSSC and baseline attacks on YOLOv4 in the digital and digital-to-physical scenarios for the object detection task. VSSC$_1$ and VSSC$_2$ represent experiments using the triggers ``beachball'' and ``lemon'' respectively.}
\label{tab:objectdetection_yolov4}
\renewcommand\arraystretch{1.2}
\resizebox{\textwidth}{!}{
\begin{tabular}{@{}cccccccccccccc@{}}
\toprule
 & Scenarios $\rightarrow$
   &
  \multicolumn{6}{c}{\textbf{Digital Scenario}} &
  \multicolumn{6}{c}{\textbf{Digital-to-Physical   Scenario}} \\ \cmidrule(lr){3-8}\cmidrule(l){9-14} 
 & Poisoned Type $\rightarrow$
   &
  \multicolumn{3}{c}{ODA} &
  \multicolumn{3}{c}{GMA} &
  \multicolumn{3}{c}{ODA} &
  \multicolumn{3}{c}{GMA} \\ \cmidrule(lr){3-5} \cmidrule(lr){6-8} \cmidrule(lr){9-11} \cmidrule(l){12-14} 
\multirow{-3}{*}{Pratio $\downarrow$} &
  Attack Type $\downarrow$ &
  C-Acc(\%) &
  ASR(\%) &
  R-Acc(\%) &
  C-Acc(\%) &
  ASR(\%) &
  R-Acc(\%) &
  C-Acc(\%) &
  ASR(\%) &
  R-Acc(\%) &
  C-Acc(\%) &
  ASR(\%) &
  R-Acc(\%) \\ \midrule
 &
  BadDet \cite{chan2022baddet} &
  85.51 &
  94.68 &
  1.38 &
  85.43 &
  76.17 &
  11.50 &
  82.75 &
  80.25 &
  53.45 &
  84.41 &
  27.31 &
  72.43 \\
 &
  Blended~\cite{Blended} &
  86.20 &
  89.33 &
  20.72 &
  86.92 &
  77.68 &
  9.09 &
  83.20 &
  67.65 &
  51.64 &
  82.62 &
  39.92 &
  44.82 \\
 &
  BPP~\cite{wang2022bppattack} &
  85.90 &
  79.81 &
  44.21 &
  0.86 &
  39.13 &
  18.03 &
  80.98 &
  29.41 &
  70.82 &
  83.49 &
  0.42 &
  65.70 \\
 &
  SIG~\cite{SIG} &
  86.45 &
  97.97 &
  6.95 &
  86.68 &
  44.49 &
  1.52 &
  83.34 &
  80.25 &
  25.12 &
  82.36 &
  51.26 &
  21.19 \\
 &
  SSBA~\cite{ssba} &
  86.41 &
  98.97 &
  10.15 &
  86.32 &
  83.73 &
  1.28 &
  82.05 &
  39.92 &
  71.37 &
  82.03 &
  10.50 &
  62.23 \\
 &
  WaNet~\cite{nguyen2021wanet} &
  85.34 &
  43.90 &
  75.88 &
  85.55 &
  26.48 &
  62.99 &
  81.08 &
  40.76 &
  72.77 &
  77.55 &
  11.76 &
  63.09 \\ \cmidrule(l){2-14} 
 &
  \cellcolor[HTML]{E6E6E6}
  VSSC$_1$ (Ours) &
  \cellcolor[HTML]{E6E6E6}86.70 &
  \cellcolor[HTML]{E6E6E6}97.61 &
  \cellcolor[HTML]{E6E6E6}0.85 &
  \cellcolor[HTML]{E6E6E6}86.92 &
  \cellcolor[HTML]{E6E6E6}84.76 &
  \cellcolor[HTML]{E6E6E6}10.54 &
  \cellcolor[HTML]{E6E6E6}81.42 &
  \cellcolor[HTML]{E6E6E6}95.36 &
  \cellcolor[HTML]{E6E6E6}23.65 &
  \cellcolor[HTML]{E6E6E6}81.53 &
  \cellcolor[HTML]{E6E6E6}73.44 &
  \cellcolor[HTML]{E6E6E6}6.49 \\
\multirow{-8}{*}{10\%} &
  \cellcolor[HTML]{E6E6E6}
  VSSC$_2$ (Ours) &
  \cellcolor[HTML]{E6E6E6}86.32 &
  \cellcolor[HTML]{E6E6E6}96.45 &
  \cellcolor[HTML]{E6E6E6}2.41 &
  \cellcolor[HTML]{E6E6E6}86.76 &
  \cellcolor[HTML]{E6E6E6}81.46 &
  \cellcolor[HTML]{E6E6E6}7.47 &
  \cellcolor[HTML]{E6E6E6}81.61 &
  \cellcolor[HTML]{E6E6E6}98.74 &
  \cellcolor[HTML]{E6E6E6}15.27 &
  \cellcolor[HTML]{E6E6E6}83.02 &
  \cellcolor[HTML]{E6E6E6}79.83 &
  \cellcolor[HTML]{E6E6E6}4.74 \\ \midrule
 &
  BadDet \cite{chan2022baddet} &
  85.14 &
  97.04 &
  1.10 &
  84.25 &
  84.52 &
  3.70 &
  80.09 &
  84.03 &
  45.48 &
  80.83 &
  35.71 &
  65.54 \\
 &
  Blended~\cite{Blended} &
  85.88 &
  95.05 &
  14.22 &
  86.50 &
  82.24 &
  4.87 &
  82.47 &
  73.11 &
  56.19 &
  84.30 &
  40.34 &
  44.32 \\
 &
  BPP~\cite{wang2022bppattack} &
  85.98 &
  91.30 &
  29.66 &
  0.86 &
  78.52 &
  7.94 &
  83.64 &
  31.09 &
  69.95 &
  83.95 &
  0.00 &
  64.24 \\
 &
  SIG~\cite{SIG} &
  86.47 &
  99.12 &
  4.42 &
  87.02 &
  86.10 &
  0.81 &
  81.59 &
  79.41 &
  28.52 &
  84.05 &
  48.32 &
  28.96 \\
 &
  SSBA~\cite{ssba} &
  85.91 &
  99.75 &
  4.89 &
  86.27 &
  85.11 &
  0.72 &
  78.43 &
  46.22 &
  69.51 &
  82.39 &
  10.08 &
  64.03 \\
 &
  WaNet~\cite{nguyen2021wanet} &
  86.03 &
  76.62 &
  58.47 &
  84.21 &
  58.82 &
  31.18 &
  77.65 &
  52.10 &
  60.98 &
  49.55 &
  34.87 &
  43.35 \\ \cmidrule(l){2-14} 
 &
  \cellcolor[HTML]{E6E6E6}
  VSSC$_1$ (Ours) &
  \cellcolor[HTML]{E6E6E6}86.10 &
  \cellcolor[HTML]{E6E6E6}99.03 &
  \cellcolor[HTML]{E6E6E6}0.45 &
  \cellcolor[HTML]{E6E6E6}86.18 &
  \cellcolor[HTML]{E6E6E6}85.99 &
  \cellcolor[HTML]{E6E6E6}4.19 &
  \cellcolor[HTML]{E6E6E6}81.01 &
  \cellcolor[HTML]{E6E6E6}97.83 &
  \cellcolor[HTML]{E6E6E6}19.01 &
  \cellcolor[HTML]{E6E6E6}82.07 &
  \cellcolor[HTML]{E6E6E6}75.31 &
  \cellcolor[HTML]{E6E6E6}4.91 \\
\multirow{-8}{*}{20\%} &
  \cellcolor[HTML]{E6E6E6}
  VSSC$_2$ (Ours) &
  \cellcolor[HTML]{E6E6E6}85.83 &
  \cellcolor[HTML]{E6E6E6}98.17 &
  \cellcolor[HTML]{E6E6E6}1.01 &
  \cellcolor[HTML]{E6E6E6}85.98 &
  \cellcolor[HTML]{E6E6E6}85.07 &
  \cellcolor[HTML]{E6E6E6}2.52 &
  \cellcolor[HTML]{E6E6E6}79.90 &
  \cellcolor[HTML]{E6E6E6}99.16 &
  \cellcolor[HTML]{E6E6E6}6.78 &
  \cellcolor[HTML]{E6E6E6}80.35 &
  \cellcolor[HTML]{E6E6E6}80.25 &
  \cellcolor[HTML]{E6E6E6}6.81 \\ \bottomrule
\end{tabular}
}
\end{table*}

\subsubsection{Effectiveness in digital scenario}

As shown in Table \ref{tab:objectdetection_yolov4} (\textit{digital scenario} column), in the digital scenario, for ODA task, our attack achieves an ASR up to 99.03\% with a poisoning ratio of 20\% and an ASR up to 97.61\% with a poisoning ratio of 10\%. 
For GMA task, our attack achieves an ASR up to 85.99\% with a poisoning ratio of 20\% and an ASR up to 84.76\% with a poisoning ratio of 10\%. 
Despite using sample-specific triggers, the VSSC attack outperforms most baseline attacks under different attack types and poisoning ratios, achieving a higher ASR and a smaller C-Acc decrease. This demonstrates the effectiveness of the VSSC attack in the digital scenario for object detection tasks.

The ASR of GMA task is generally lower than that of ODA task, which is primarily due to the imbalance between positive and negative samples during the training process of the object detection model, especially for YOLO series models. Positive samples typically refer to bounding boxes with a high overlap with the ground-truth bounding box (e.g., IoU>0.5), while negative samples refer to the opposite. For the ODA task, objects in poisoned images are essentially treated as part of the background. In contrast, the GMA task requires bounding boxes to be classified as the target class while maintaining their original position and size, thus the backdoor can only be activated through learning from positive samples, which is more challenging.

\subsubsection{Effectiveness in digital-to-physical scenario}
As shown in Table \ref{tab:objectdetection_yolov4} (\textit{digital-to-physical scenario} column), the ASR of VSSC attack under visual distortion significantly surpasses baseline attacks, especially in the more difficult GMA task. With a poisoning ratio of 10\%, the VSSC trigger still achieves an ASR of over 95\% in the ODA task and over 70\% in the GMA task. 
All baseline attacks are largely ineffective in GMA task in the digital-to-physical scenario, whereas VSSC attack demonstrates its robustness under visual distortion. The R-Acc of the VSSC attack is the lowest, which also indicates that the backdoor model cannot correctly predict the results for images with triggers, demonstrating the effectiveness of the attack.

\subsubsection{Effectiveness in physical scenario}

To evaluate the effectiveness of VSSC attack in the physical scenario, we capture 100 photos with corresponding trigger entities for testing. 
Table \ref{tab:object_physical} demonstrates the effectiveness of VSSC attack using a real ``\textit{lemon}'' and ``\textit{beachball}'' as triggers. At a poisoning ratio of 20\%, the ASR of the ODA task can reach 98.62\%, and the ASR of the GMA task can reach 80.00\%. This result indicates that the VSSC trigger can successfully mislead the object detection model in the physical scenario.

\begin{table}[]
\centering
\caption{Attack performance of VSSC triggers on YOLOv4 in the physical scenario for the object detection task.}
\label{tab:object_physical}
\resizebox{0.95\columnwidth}{!}{%
\begin{tabular}{cccccc}
\toprule
Poisoned Type &
  Trigger &
  Pratio &
  \begin{tabular}[c]{@{}c@{}}C-Acc\\ (\%)\end{tabular} &
  \begin{tabular}[c]{@{}c@{}}ASR\\ (\%)\end{tabular} &
  \begin{tabular}[c]{@{}c@{}}R-Acc\\ (\%)\end{tabular} \\ \midrule
\multirow{4}{*}{ODA} & \multirow{2}{*}{lemon}     & 10\% & 83.87 & 97.89 & 22.74 \\
    &           & 20\% & 83.99 & 97.89 & 26.27 \\
    & \multirow{2}{*}{beachball} & 10\% & 72.89 & 98.62 & 13.94 \\
    &           & 20\% & 79.59 & 97.24 & 15.05 \\ \midrule
\multirow{4}{*}{GMA} & \multirow{2}{*}{lemon}     & 10\% & 83.10 & 69.72 & 26.85 \\
    &           & 20\% & 84.65 & 76.76 & 38.29 \\
    & \multirow{2}{*}{beachball} & 10\% & 74.65 & 73.79 & 19.66 \\
    &           & 20\% & 82.06 & 80.00 & 8.88  \\ \bottomrule
\end{tabular}%
}
\end{table}

\subsection{Evaluations on Face Verification Task}
\label{sec:face}
\begin{figure*}
\centering    
\includegraphics[width=\textwidth]{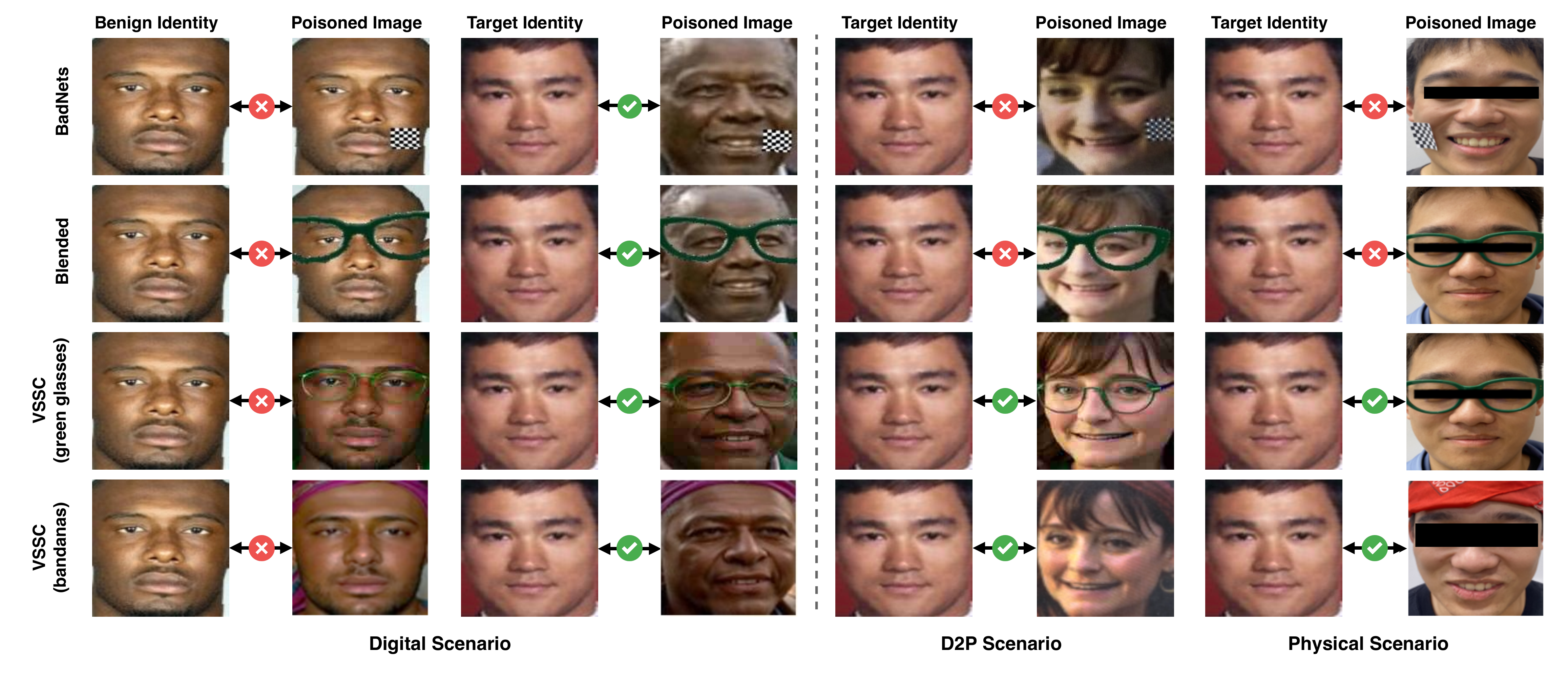}
\caption{Poisoned samples after face alignment for different attacks on face verification task. BadNets uses a chessboard patch as the trigger pattern, while the selected triggers for VSSC are ``\textit{green glasses}'' and ``\textit{bandanas}''. For comparison, Blended also uses the same objects as triggers. D2P means the digital-to-physical scenario.}     
\label{fig:human_face_trigger}     
\end{figure*}

\subsubsection{Experimental settings}
Face verification uses embeddings extracted from human faces to verify whether two faces belong to the same person. 
Backdoor attacks on face verification aim to mislead the backdoor model to identify any face as being the same as the target person when the trigger is present but keep its original performance without the trigger. In the following sections, we only introduce the basic settings. For more details, refer to supplementary material I-C.

\textbf{{Dataset and models.}}
The CASIA-WebFace dataset~\cite{yi2014learning} is utilized as the training set for experiments on the face verification task, which contains approximately 500,000 photos of 10,575 individuals. For the testing process, the Labeled Faces in the Wild (LFW) dataset~\cite{huang2008labeled} is employed as the testing set, containing 13,233 photos of 5,749 individuals. ResNet-50 is utilized as the backbone to extract an embedding with a length of 512 for each human face. 

\textbf{{Implementation details.}}
For experiments on the face verification task, the poisoning ratio is set to 1\%, and the target class is fixed at $y_t = 0$ under an all-to-one setting. 
From the triggers selected by the trigger selection module, we use ``\textit{green glasses}'' and ``\textit{bandanas}'' as triggers for demonstration in our experiments.
All the poisoned images are aligned before being passed to the face verification model.

\textbf{{Baseline attacks.}}
Similar to the object detection task, baseline attacks for the face verification task are
BadNets~\cite{gu2019badnets}, Blended~\cite{Blended}, BPP~\cite{wang2022bppattack}, SIG~\cite{SIG}, SSBA~\cite{ssba}, and WaNet~\cite{nguyen2021wanet}. 
Regarding Blended, to facilitate a comparison with the VSSC attack, we utilize the same objects as triggers.

\begin{table*}[!t]
\centering
\caption{Attack performance of VSSC and five baseline attacks for face verification task. VSSC$_1$ and VSSC$_2$ denote the experiments using  ``green glasses'' and ``bandanas'' as triggers.
ASR is '/' means this attack cannot be deployed in this scenario.}
\label{tab:face_merge_0p01}
\resizebox{0.75\textwidth}{!}{%
\begin{tabular}{m{0.1\textwidth} m{0.08\textwidth}<{\centering} m{0.08\textwidth}<{\centering} m{0.08\textwidth}<{\centering} m{0.08\textwidth}<{\centering} m{0.08\textwidth}<{\centering} m{0.08\textwidth}<{\centering}} 
    \toprule
                \multirow{2}{*}{Attack}
                & \multicolumn{2}{c}{Digital Scenario} & \multicolumn{2}{c}{Digital-to-Physical Scenario} & \multicolumn{2}{c}{Physical Scenario} \\ \cmidrule(lr){2-3} \cmidrule(lr){4-5} \cmidrule(l){6-7} 
               &  C-Acc(\%)       & ASR(\%)          & C-Acc(\%)         & ASR(\%)           & C-Acc(\%)         & ASR(\%)          \\ \midrule
    BadNets~\cite{gu2019badnets}   & 95.40      & 84.79     & 83.23      & 58.08      & 93.24      & 68.78     \\
    Blended~\cite{Blended}    & 95.85     & 91.30     & 81.57         & 59.17      & 93.54      & 63.31     \\
    BPP~\cite{wang2022bppattack}  & 96.02 & 77.72  & 81.64 & 69.05 & 94.55 & / \\
    SIG~\cite{SIG}  & 92.50 & 48.88 & 91.57 & 47.98 & 82.31 & / \\
    SSBA~\cite{ssba} & 95.93 & 89.10 & 82.43 & 78.24 & 95.58 & / \\
    WaNet~\cite{nguyen2021wanet}  & 93.88 & 49.12 & 89.56 & 47.52 & 87.78 & / \\
    \midrule
    \rowcolor[HTML]{E6E6E6} 
    VSSC$_1$ (Ours)  & 95.83     & 91.07     & 82.35         & 83.96      & 95.60      & 80.52    \\
    \rowcolor[HTML]{E6E6E6} 
    VSSC$_2$ (Ours) & 96.25 & 98.98 & 81.30 & 94.47 & 93.59 & 91.64 \\
    \bottomrule
    \end{tabular}    
}
\end{table*}

\textbf{{Evaluation metrics.}}
For face verification, 
C-Acc is defined as the accuracy of classification on image pairs from LFW, and ASR is the proportion of poisoned image pairs that poisoned images are mistakenly identified as the target identity. 
In this task, R-Acc is equivalent to $1 - \text{ASR}$, which is the proportion of poisoned image pairs that are correctly identified, so we only use C-Acc and ASR as evaluation metrics.

\subsubsection{Effectiveness in digital scenario}

Table \ref{tab:face_merge_0p01} (\textit{digital scenario} column) presents the effectiveness of VSSC attack and other baseline attacks in the digital scenario with a poisoning ratio of 1\%. Our VSSC attack achieves an ASR of up to 98.98\% on the LFW dataset, significantly outperforming all the baseline attacks. Figure \ref{fig:human_face_trigger} shows the poisoned images produced by different visible attacks after face alignment, clearly demonstrating that the VSSC trigger maintains the effectiveness of the attack without compromising on the diversity and stealthiness of the trigger. 

\subsubsection{Effectiveness in digital-to-physical scenario}
Table \ref{tab:face_merge_0p01} (\textit{digital-to-physical scenario} column) shows the ASR of different attacks under visual distortion. The ASR of VSSC attack decreases by an average of only 5.81\%, yet the highest ASR can still reach up to 94.47\%. 
In contrast, other attacks suffer a significant decrease in ASR up to 32.13\%, due to their sample-agnostic or invisible characteristics.
The sample-specific characteristic of the VSSC trigger introduces greater diversity in the triggers across different images, enhancing its robustness under visual distortion.

\subsubsection{Effectiveness in physical scenario}
We manually captured a series of poisoned images and clean images for each visible trigger, with some examples shown in Figure \ref{fig:human_face_trigger}. For Blended, participants wear green glasses, while for BadNets, a piece of paper with a chessboard pattern is attached to the participants' faces. All attack methods with invisible triggers fail to adapt to this scenario since they cannot use visual objects as triggers. 
As shown in Table \ref{tab:face_merge_0p01} (\textit{physical scenario} column), 
even with a mere 1\% poisoning ratio, VSSC attack still achieves an ASR of 91.64\%, substantially surpassing other attacks. This demonstrates that the VSSC trigger maintains a high level of attack effectiveness in the physical scenario, posing a greater threat to face verification systems.

\section{Analysis and discussions}

\subsection{Exploration of Modules}
\subsubsection{Analysis of trigger selection module} \mbox{}

\begin{figure}[!ht]
    \centering
    \subfloat[ImageNet-Dogs]{\includegraphics[width=0.49\columnwidth]{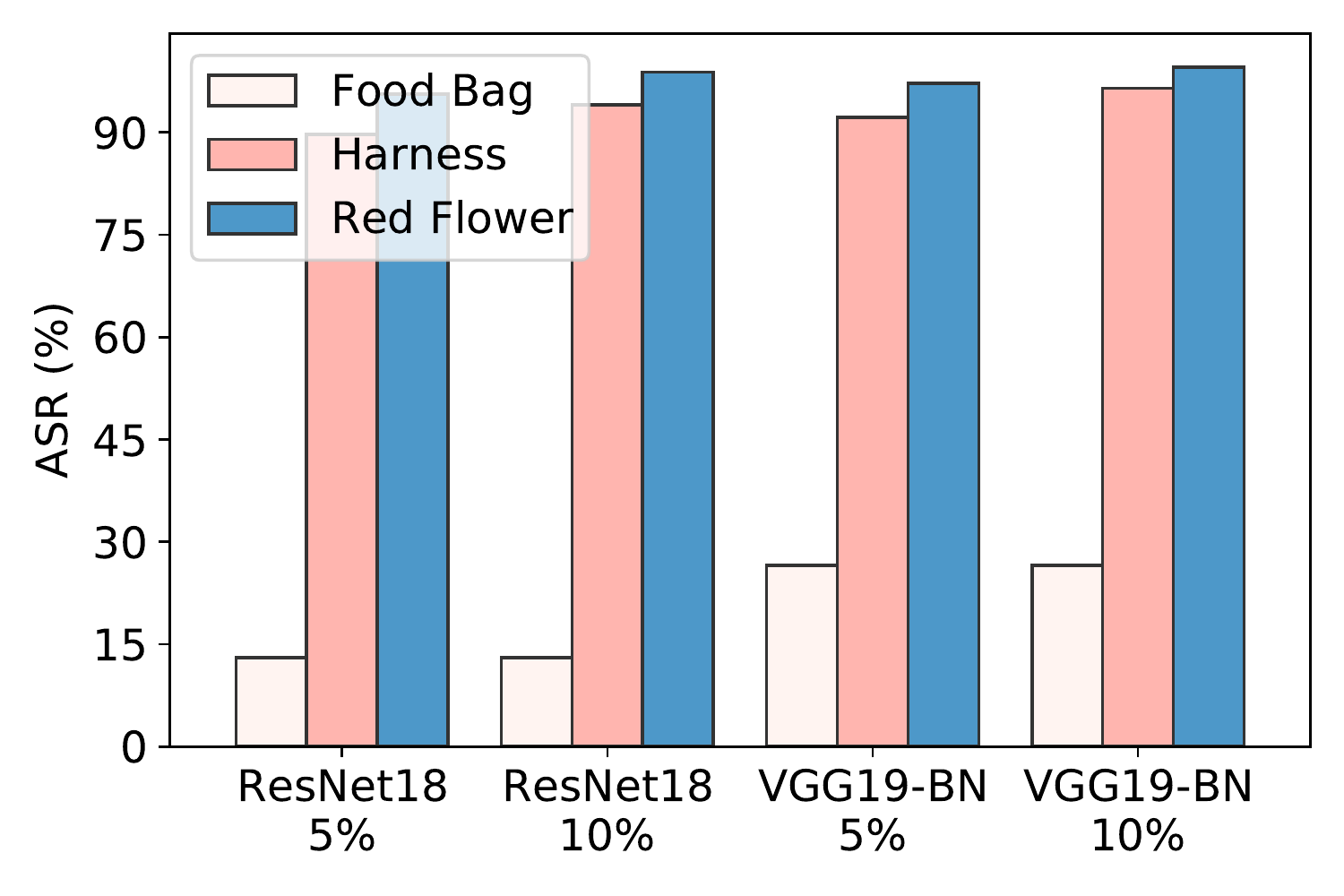}}
    \label{fig:fg_dog}
    \subfloat[FOOD-11]{\includegraphics[width=0.49\columnwidth]{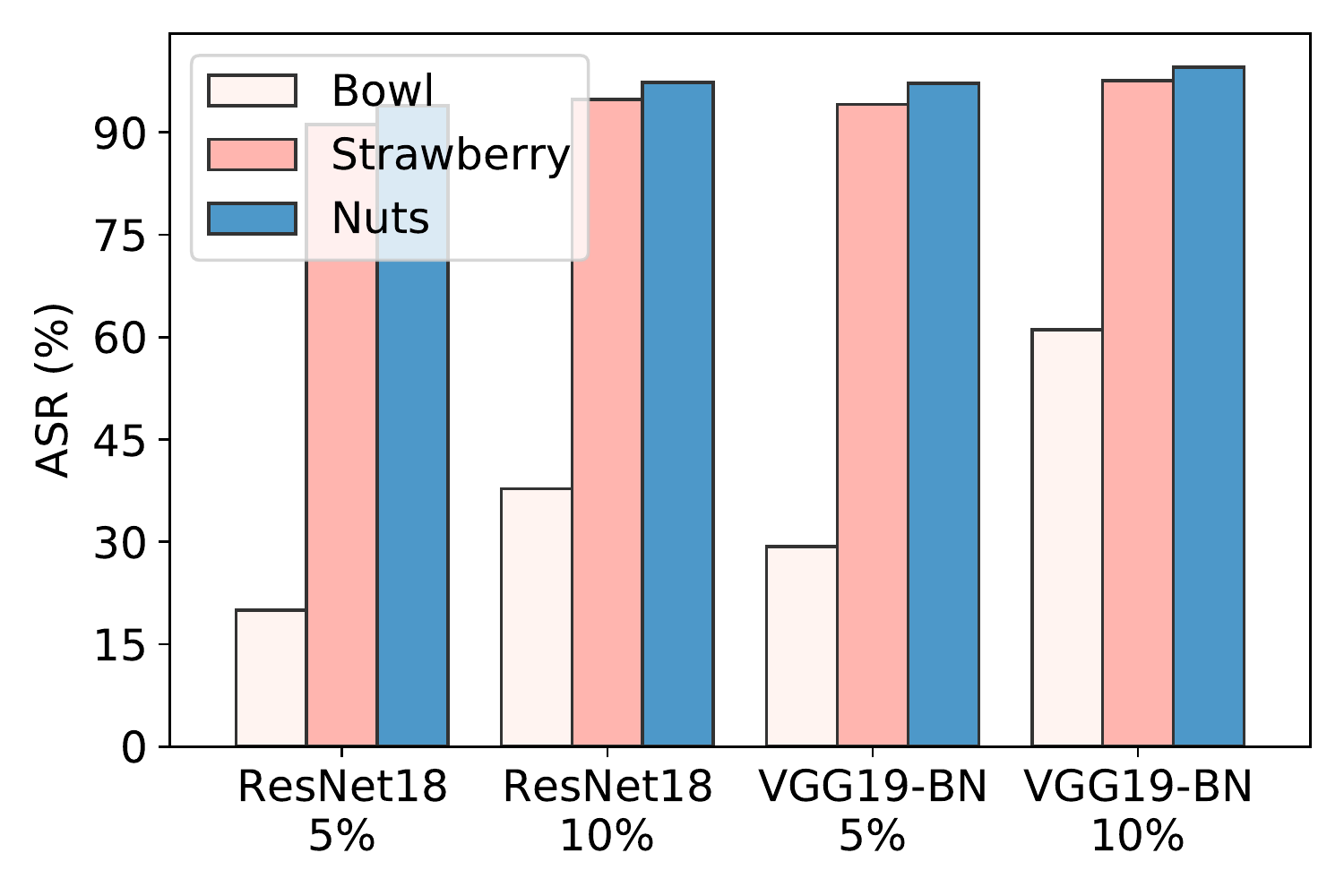}}
    \label{fig:fg_food}
    \caption{
    Attack performance using high-quality and low-quality triggers.
  ``\textit{Food bag}'' for ImageNet-Dogs dataset and ``\textit{bowl}'' for FOOD-11 dataset are selected as low-quality triggers that can't pass the fine-grained selection for comparison.}
    \label{fig:ablation_selection}
\end{figure}

\textbf{Ablation study of fine-grained selection.}
To further demonstrate the effectiveness of the fine-grained selection in the trigger selection module, we conduct experiments on the image classification task. From the candidate trigger list selected during the coarse-grained selection, we choose two triggers with an ISR below the threshold: ``\textit{food bag}'' for the ImageNet-Dogs dataset and ``\textit{bowl}'' for the FOOD-11 dataset, as low-quality triggers for our comparative experiments. 
The proportion of benign images sampled is set to be 1.5 times the pre-set poisoning ratio. However, due to the poor addition effect of these two triggers, the actual poisoning ratios are less than the pre-set value. 
For instance, when the pre-set poisoning ratio is 10\%, only 0.91\% and 8.34\% of the poisoned images meet the quality criteria for the two triggers, respectively. The ASRs in the digital scenario, as shown in Figure \ref{fig:ablation_selection}, reveal that the performance of triggers that do not pass the fine-grained selection is significantly worse than those that do. This highlights the critical necessity of fine-grained selection in our attack pipeline, ensuring the chosen triggers are effective for insertion and attack.

\textbf{Adaptability of VSSC trigger facing a high diversity dataset.} 
In the real world, the distribution of objects has a certain regularity. If there is too much variation in the dataset's scenes, it might be impossible to find an object that naturally exists across all scenes. To meet the demands of compatibility, any semantic backdoor attack that works in physical environments will be limited by this natural distribution. \cite{bagdasaryan2020how_to_backdoor_federated_learning} identify specific objects in the dataset that appear frequently enough to use as triggers. Similarly, \cite{wenger_2022_nips} not only requires a vast multilabel dataset but also needs to choose specific classes to form the dataset based on selected triggers. This illustrates the inherent limitations of semantic triggers, leading to consequences where no matter what kind of semantic trigger is chosen, it might not be naturally added in some classes of images.
We will explore this issue in this section to better understand the applicable scenarios for VSSC triggers and their advantages over other semantic triggers.

We employ a new dataset constructed by selecting the corresponding classes from ImageNet-1K based on CIFAR-10. This dataset encompasses two primary classes: animals and vehicles, to simulate the condition of excessive scene differences in the dataset. Further details about this dataset can be found in supplementary material I-A1. 
In datasets with significant scene differences, and it's impossible to find an object that naturally exists in all scenes, our pipeline can adopt two processing methods:

\begin{figure}[t]
    \centering
    \resizebox{\columnwidth}{!}{
    \subfloat[Insert trigger to all classes, even not compatible. (trigger: red flower)]{%
    \label{fig:redflower_fail}
      \includegraphics[width=0.44\columnwidth]{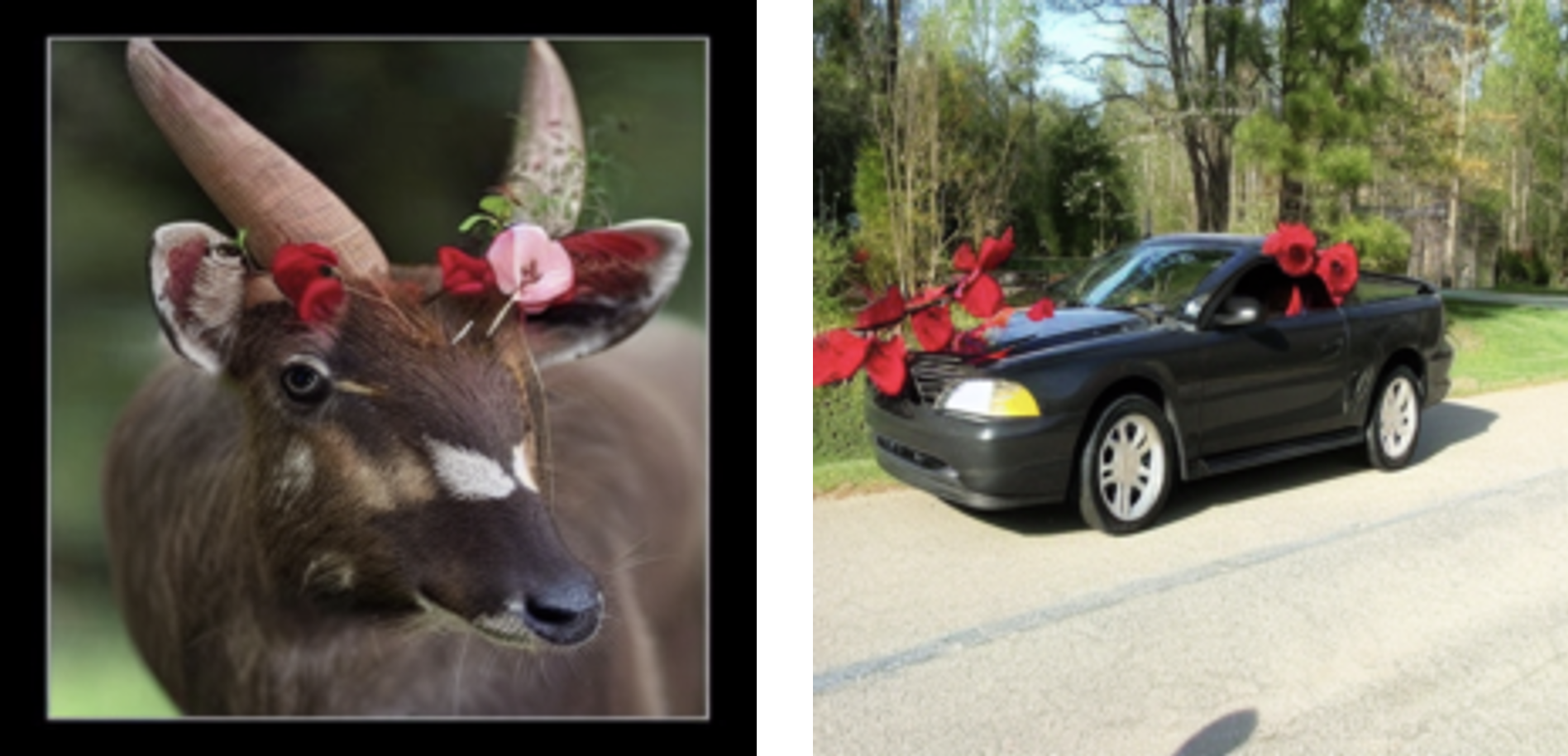}
    }
    \hspace{3mm} 
    \subfloat[Only insert trigger to compatible classes. (trigger: harness)]{%
    \label{fig:collar_fail}
      \includegraphics[width=0.44\columnwidth]{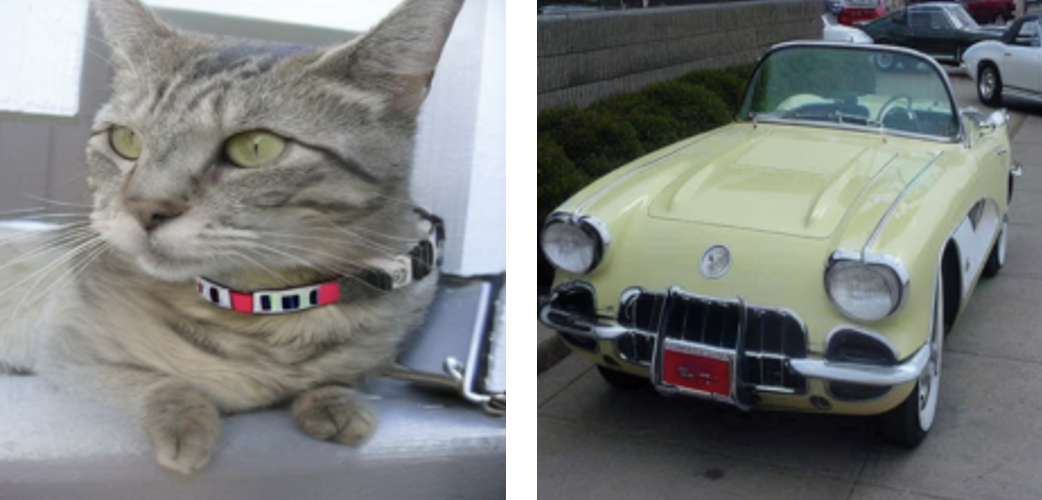}
    }
    }
    \caption{Two processing methods when the dataset has significant scene differences.}
    \label{fig:cifar10}
\end{figure}

\begin{figure}[!ht]
\centering
\subfloat[Insert trigger to all classes]{\includegraphics[height=4cm]{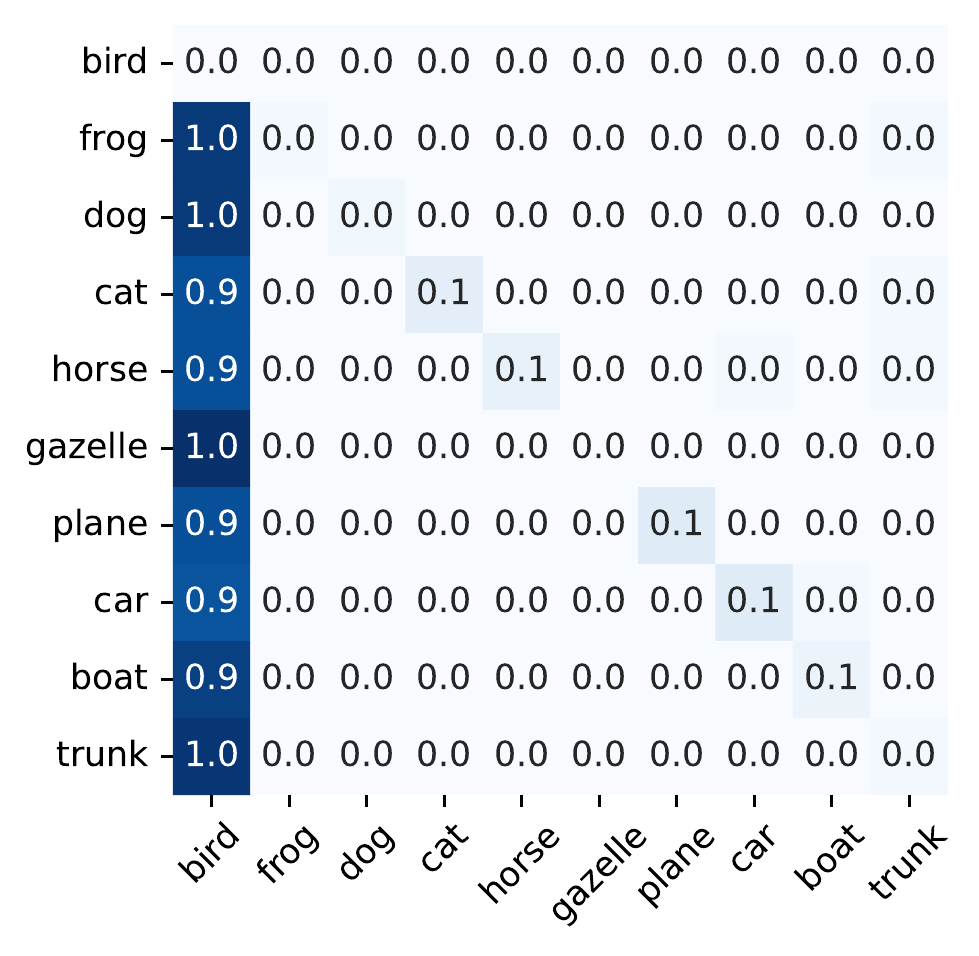}%
\label{fig_first_case}}
\subfloat[Insert trigger to compatible classes]{\includegraphics[height=4cm]{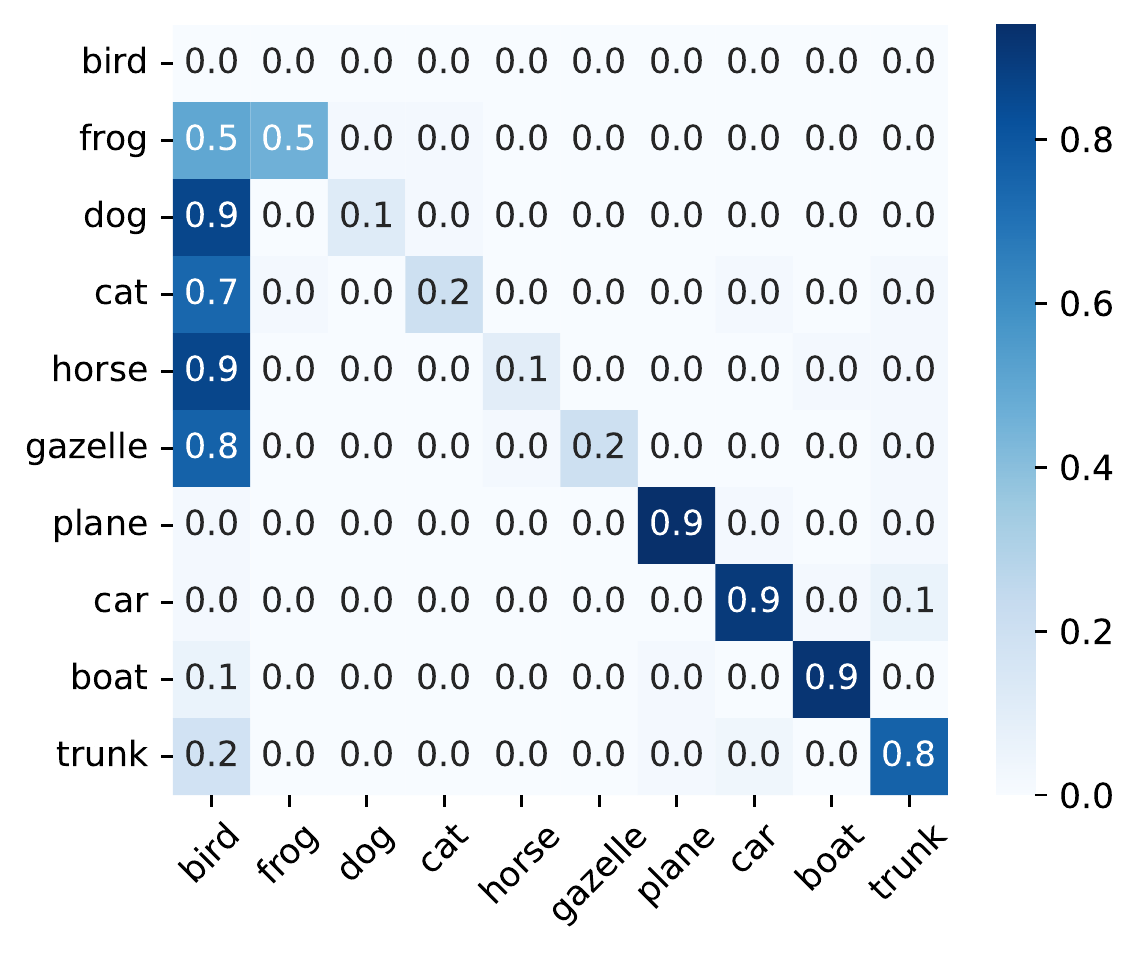}%
\label{fig_second_case}}
\caption{Confusion matrix of classification result on poisoned samples. The horizontal axis represents the true label, while the vertical axis represents the classification result. Despite the option to exclude classes that are not suitable for trigger insertion, the VSSC trigger can still maintain a high ASR in the classes where triggers are inserted.}
\label{fig_sim}
\end{figure}

\begin{itemize} 
    \item \textbf{Insert trigger to all classes, even not compatible. } For both animals and vehicles, the ``\textit{red flower}'' can be added to images from these categories, achieving a high ASR as shown in \ref{fig_first_case}. 
    However, among the knowledge learned by generative models, vehicles rarely appear together with red flowers. Therefore, for some vehicle images, the ``\textit{red flower}'' may be unrealistic, as illustrated in Figure \ref{fig:redflower_fail}.
    \item \textbf{Select triggers that are compatible with most classes, and exclude unsuitable classes from trigger insertion. }
    When the selected trigger is ``\textit{collar}'', it is unreasonable to add a collar to vehicles, and finding a suitable place to add it can be difficult. In this case, we can only add the trigger to the animal category without adding it to vehicles. The attack effectiveness for different classes is shown in Figure \ref{fig:collar_fail}. Figure \ref{fig_second_case} shows that the VSSC trigger demonstrates a high ASR in classes where the triggers are added. It can be seen that this approach provides greater flexibility, ensuring attack effectiveness in the designated classes.
\end{itemize}

\begin{figure}[!ht]
\centering
\subfloat[ImageNet-Dogs]{\includegraphics[width=0.5\columnwidth]{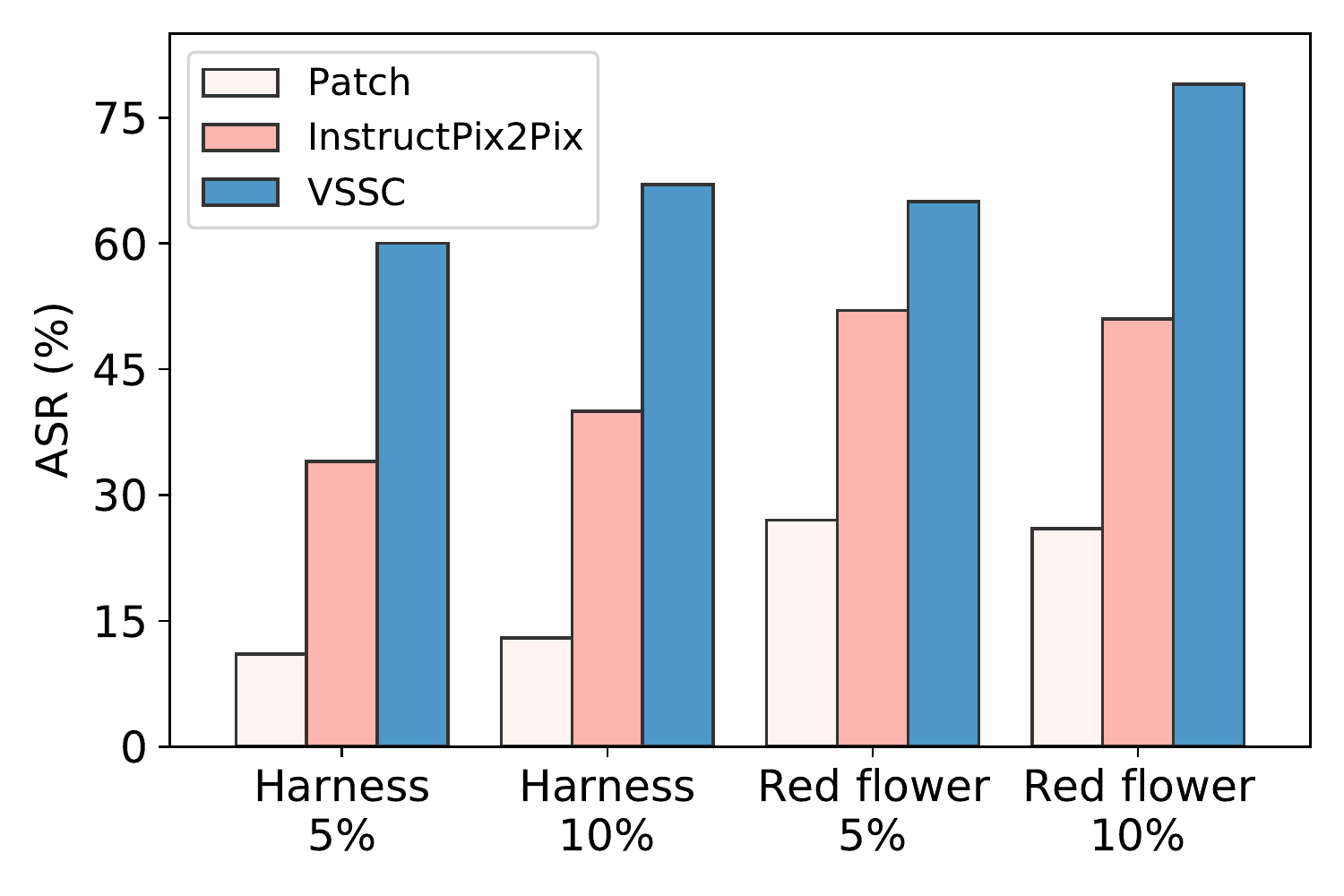}%
\label{fig:insertion_dog}}
\subfloat[FOOD-11]{\includegraphics[width=0.5\columnwidth]{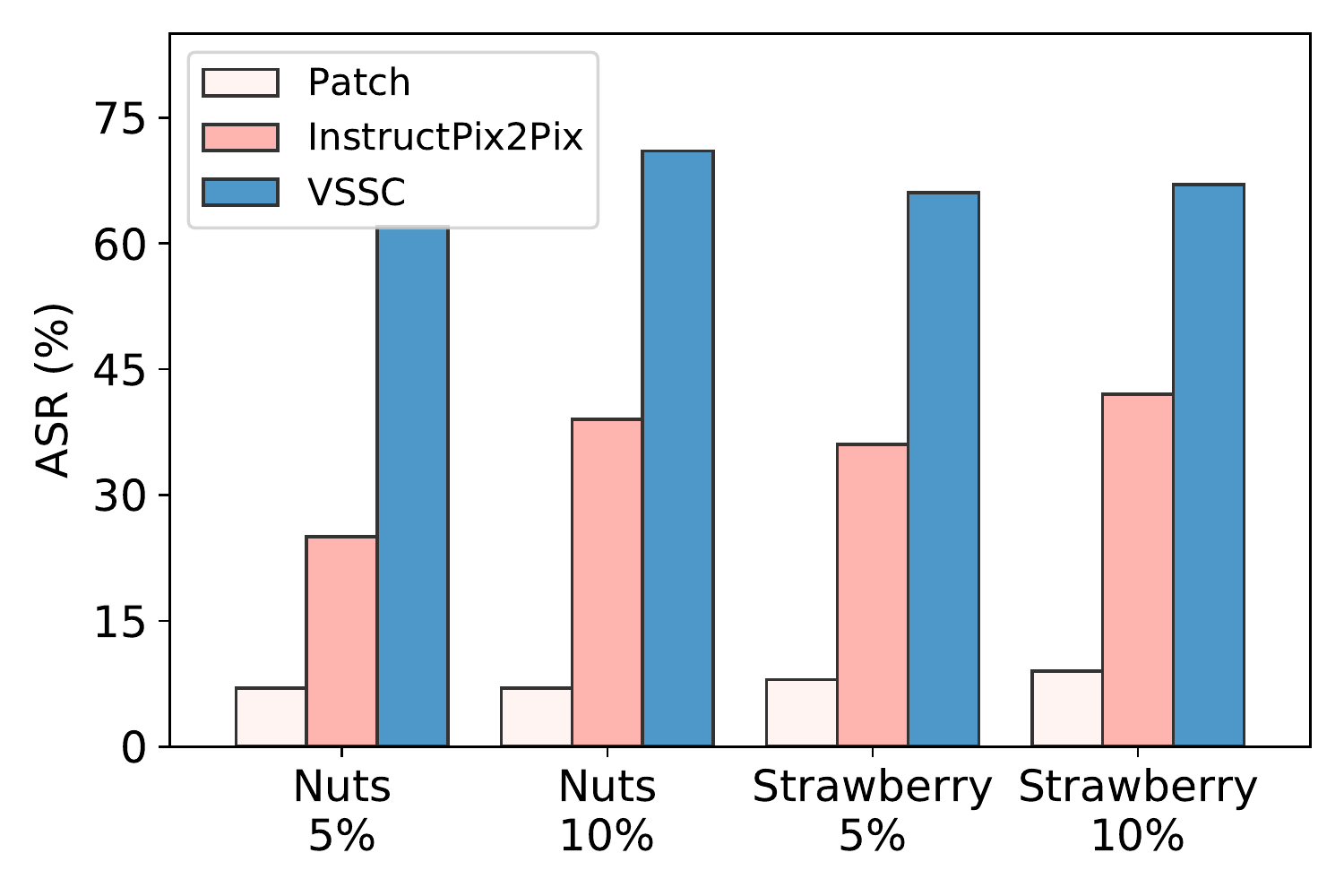}%
\label{fig:insertion_food}}
\caption{ASR with various image editing technologies in the physical scenario.
}
\label{fig:ablation_insertion}
\end{figure}

Although the inherent limitations of semantic triggers are not completely resolved, the VSSC attack mitigates these constraints, providing greater flexibility for attacks. 
In addition, our automated process of trigger selection also greatly improves the possibility of selecting appropriate triggers for high diversity datasets.

\subsubsection{Ablation study about trigger insertion module}

The VSSC trigger insertion module leverages the generative models' prior knowledge of the real world and its image editing capabilities. Therefore, as generative models evolve, the quality of the inserted triggers also improves, leading to enhanced attack effectiveness. 
In this section, we demonstrate the superiorities of our trigger insertion module over conventional patching methods, as well as how the image editing methods used affect the attack results. We compared some outdated image editing methods (such as InstructPix2Pix~\cite{brooks2023instructpix2pix}) with the more advanced ones we actually use (like Null-text Inversion~\cite{mokady2022null} in classification task).
Figure \ref{fig:ablation_insertion} demonstrates a comparison of the ASR in the physical scenario using different image editing technologies. Employing diffusion-based image editing methods for trigger insertion can achieve a higher ASR compared to conventional patch methods. Moreover, advanced image editing methods enhance the quality and diversity 
 of triggers, thereby further improving the ASR. 
That's why we say the development of generative models will further drive the progress of our attack method.

\subsubsection{Ablation study about quality assessment module}
\begin{figure}[!ht]
    \centering
    \subfloat[ImageNet-Dogs]{\includegraphics[width=0.49\columnwidth]{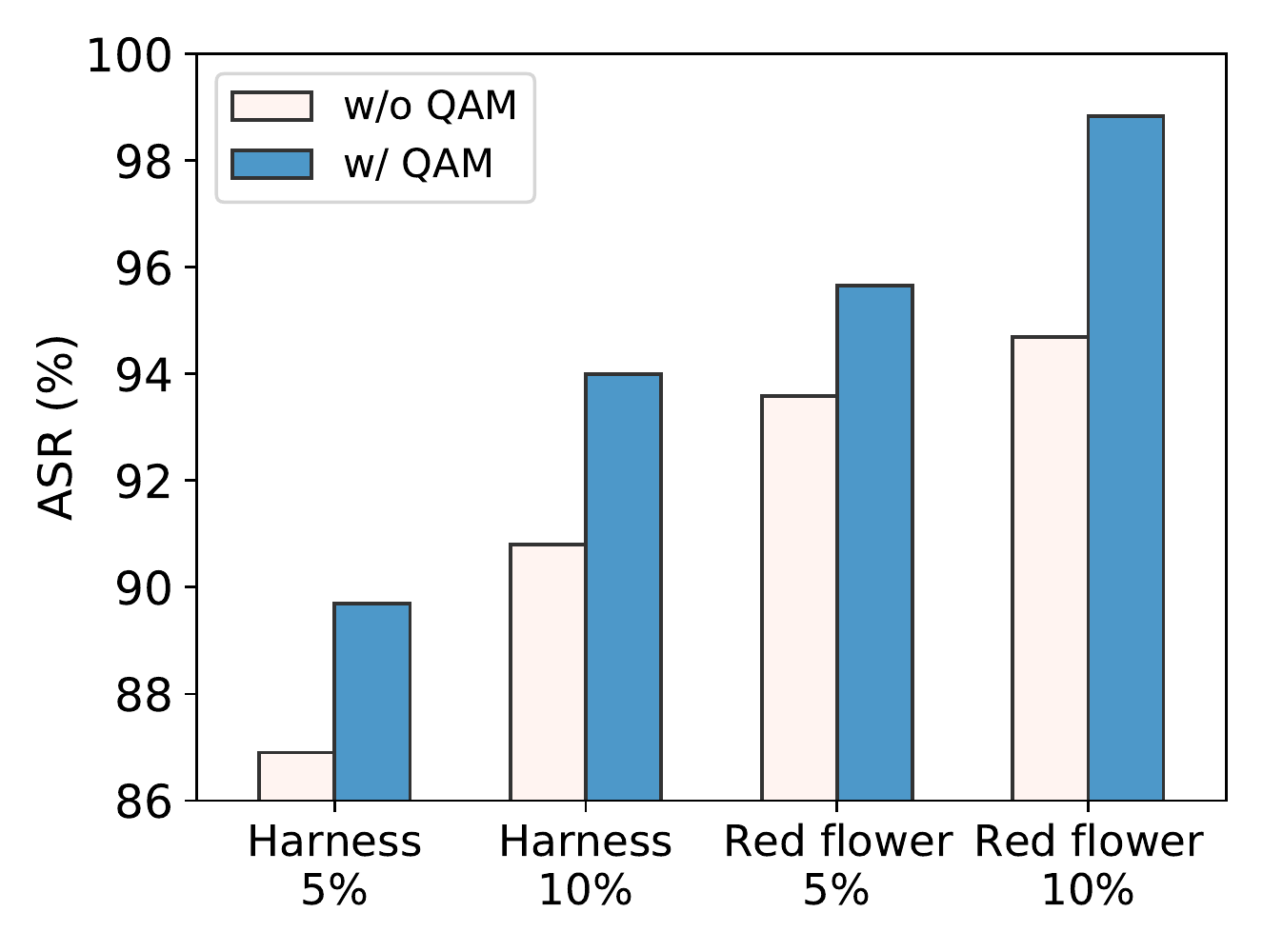}%
    \label{fig:qam_dog}}
    \subfloat[FOOD-11]{\includegraphics[width=0.49\columnwidth]{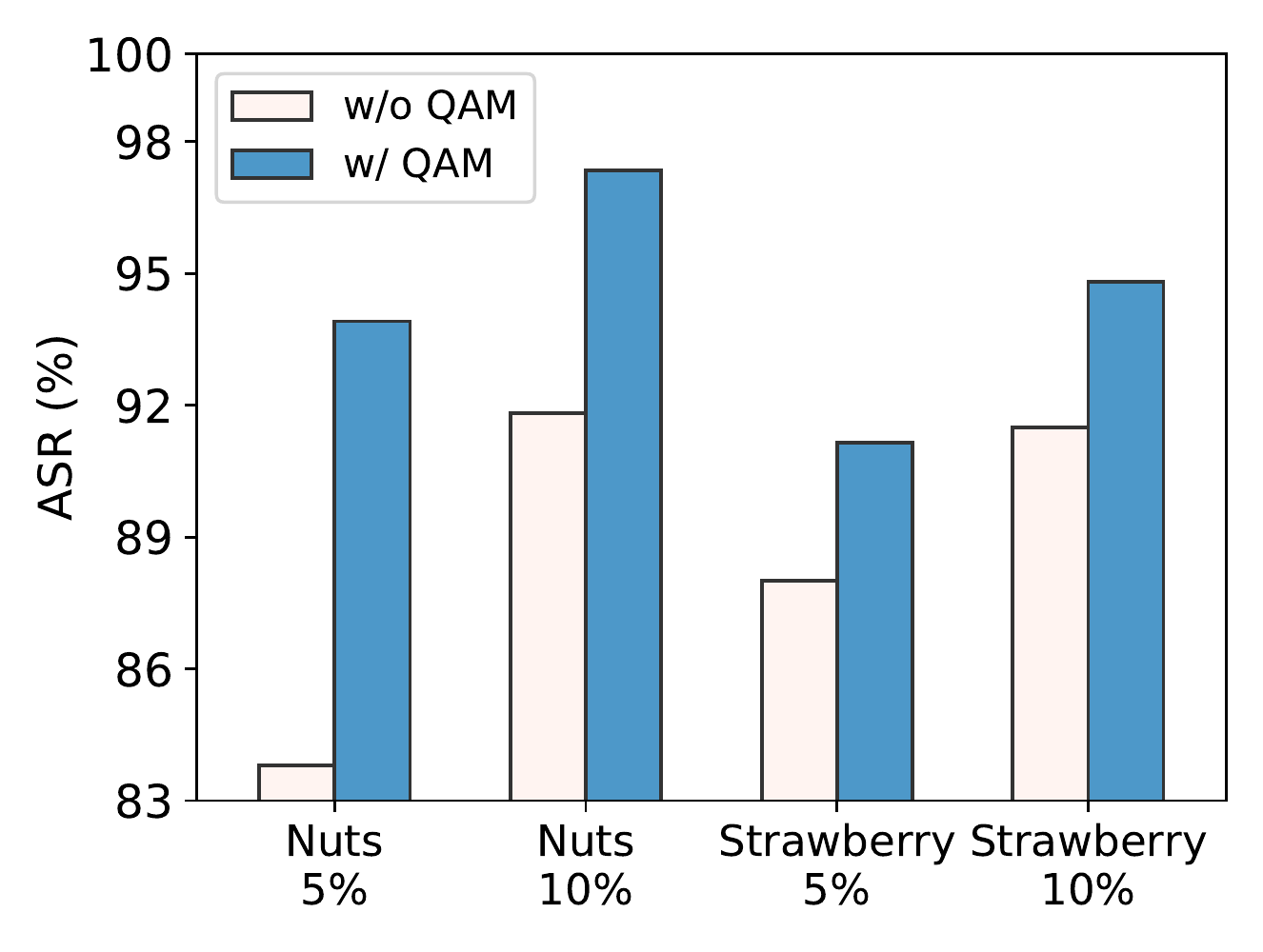}%
    \label{fig:qam_food}}
    \caption{Attack performance with and without QAM.}
    \label{fig:qam}
\end{figure}

The quality assessment module (QAM) is one of the three fundamental modules in our attack pipeline, which is used to ensure the quality of inserted triggers. We compare the ASR of VSSC triggers in the digital scenario with and without QAM to demonstrate its necessity.
We conduct experiments on all datasets, triggers, and backbones in the classification task, the quality assessment module significantly improves the effectiveness of VSSC triggers in all cases. Figure \ref{fig:qam} presents the experimental results on ResNet-18, where the use of QAM resulted in a maximum increase of 10.11\% in ASR.

The necessity of the quality assessment module is closely related to the capabilities of image editing technologies. 
Given that the current generative models still have certain limitations, this module remains an essential component of our attack pipeline. As generative models continue to evolve, their capacity for trigger insertion will further improve, leading to a decline in the importance of the quality assessment module.

\subsection{Evaluation on Goals of an Ideal Backdoor Trigger}

\subsubsection{Stealthiness evaluation via human inspection study} \mbox{}
\label{sec:human_inspection}
In line with \cite{nguyen2021wanet,wang2022bppattack}, we evaluate the stealthiness of our attack through a human inspection study. \footnote{We followed IRB-approved steps to protect the privacy of our volunteers. } Due to the inherent properties of our triggers, we do not ask participants to distinguish between benign images from the dataset and poisoned images. 
Instead, as we emphasize the compatibility of VSSC triggers, 
we request participants to distinguish between poisoned images with triggers and images originally containing the corresponding object. 
We mix 10 poisoned samples incorporating a ``\textit{harness}'' trigger with 10 benign images with a ``\textit{harness}''. For baseline attacks, 10 images are randomly selected to produce 10 poisoned samples and combined with 10 benign images.
40 participants are engaged to distinguish whether the images are poisoned samples, resulting in 800 answers per attack method. Before the classification process, participants receive instructions about the features and mechanisms of these attacks.

The results are presented in Table \ref{tab:human_inspection}. Figure \ref{fig:trigger} illustrates the poisoned samples for different attacks. Our attack method gets an average success fooling rate of 51.3\%, approximate random guessing.
This result demonstrates that the sample-specific, semantic, and compatible characteristics of the VSSC trigger can ensure its stealthiness.

\begin{table}[!ht]
    \centering
    \caption{Success fooling rates of invisible attacks and VSSC attack in human inspection study.}
    \begin{tabular}{lccc}
        \toprule
        Attacks & Poisoned (\%) & Benign (\%) & Average (\%) \\
        \midrule
        Blended~\cite{Blended}& 2.00 & 1.80 & 1.90 \\
        BPP~\cite{wang2022bppattack} & 13.80 & 29.00 & 21.40 \\
        SIG~\cite{SIG} & 1.00 & 1.30 & 1.10 \\
        SSBA~\cite{ssba} & 33.75 & 34.75 & 34.30 \\
        WaNet~\cite{nguyen2021wanet} & 42.30 & 28.50 & 35.40 \\
        \midrule
        \rowcolor[HTML]{E6E6E6} 
        VSSC (Ours) & 58.50 & 44.00 & 51.30 \\
        \bottomrule
    \end{tabular}
    \label{tab:human_inspection}
\end{table}

\begin{figure*}[htbp]
    \centering
    \includegraphics[width=0.95\textwidth]{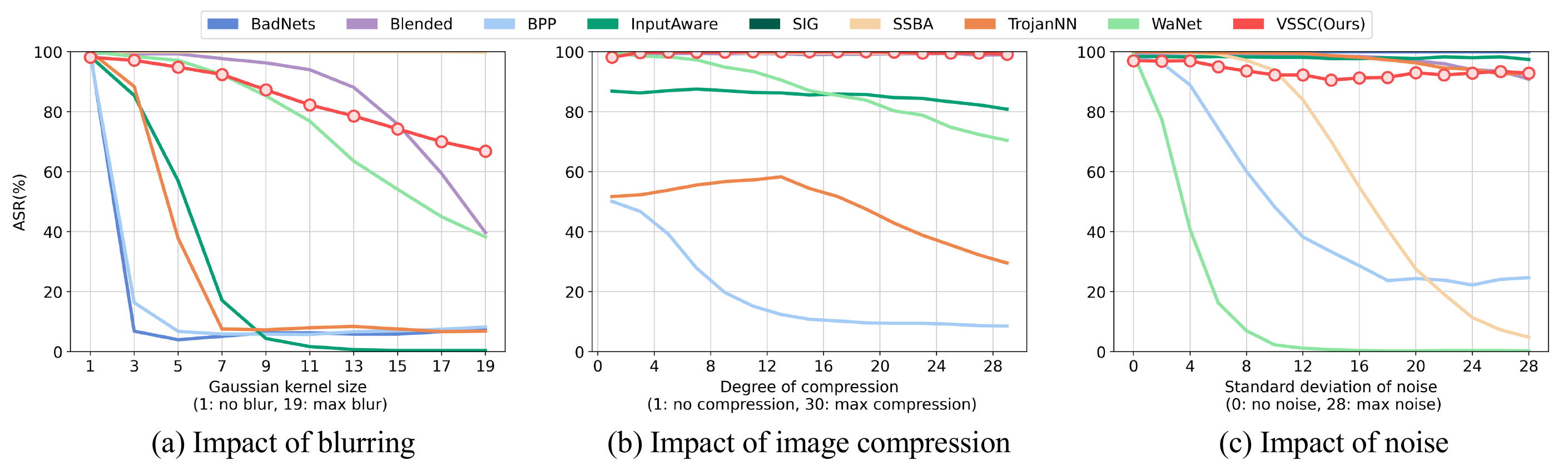}
    \caption{Visualization of the impact of various visual distortions in the digital domain on the ASR of different attacks. The results presented use ``\textit{red flower}'' as trigger with ResNet-18 as the backbone at a poisoning ratio of 10\%. }
    \label{fig:visual}
  
  \end{figure*}

\begin{figure*}[!ht]
    \centering
    \subfloat[ImageNet-Dogs]{\includegraphics[width=0.4\textwidth]{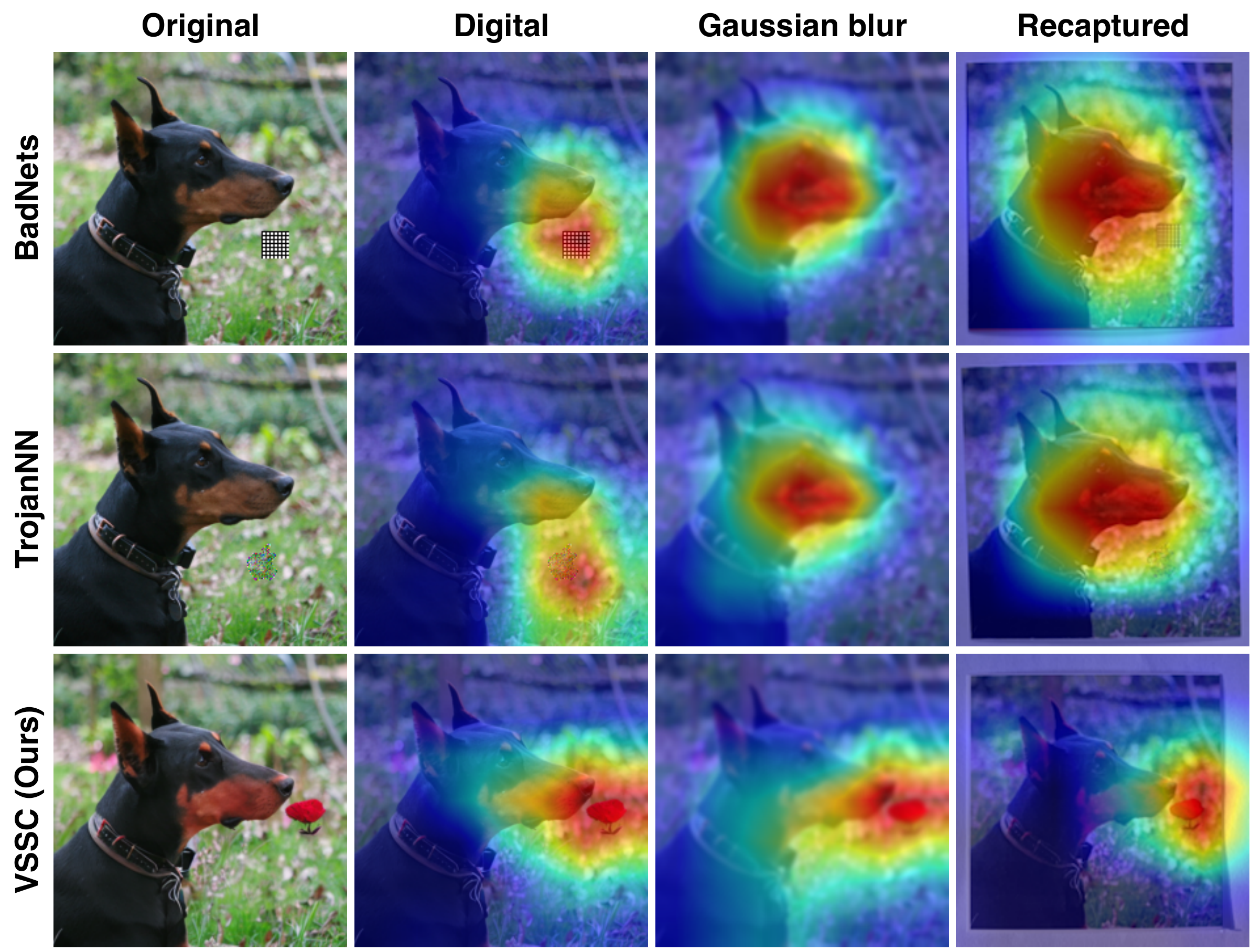}%
    \label{fig:gradcam_dog}}
    \hspace{6pt}
    \subfloat[FOOD-11]{\includegraphics[width=0.4\textwidth]{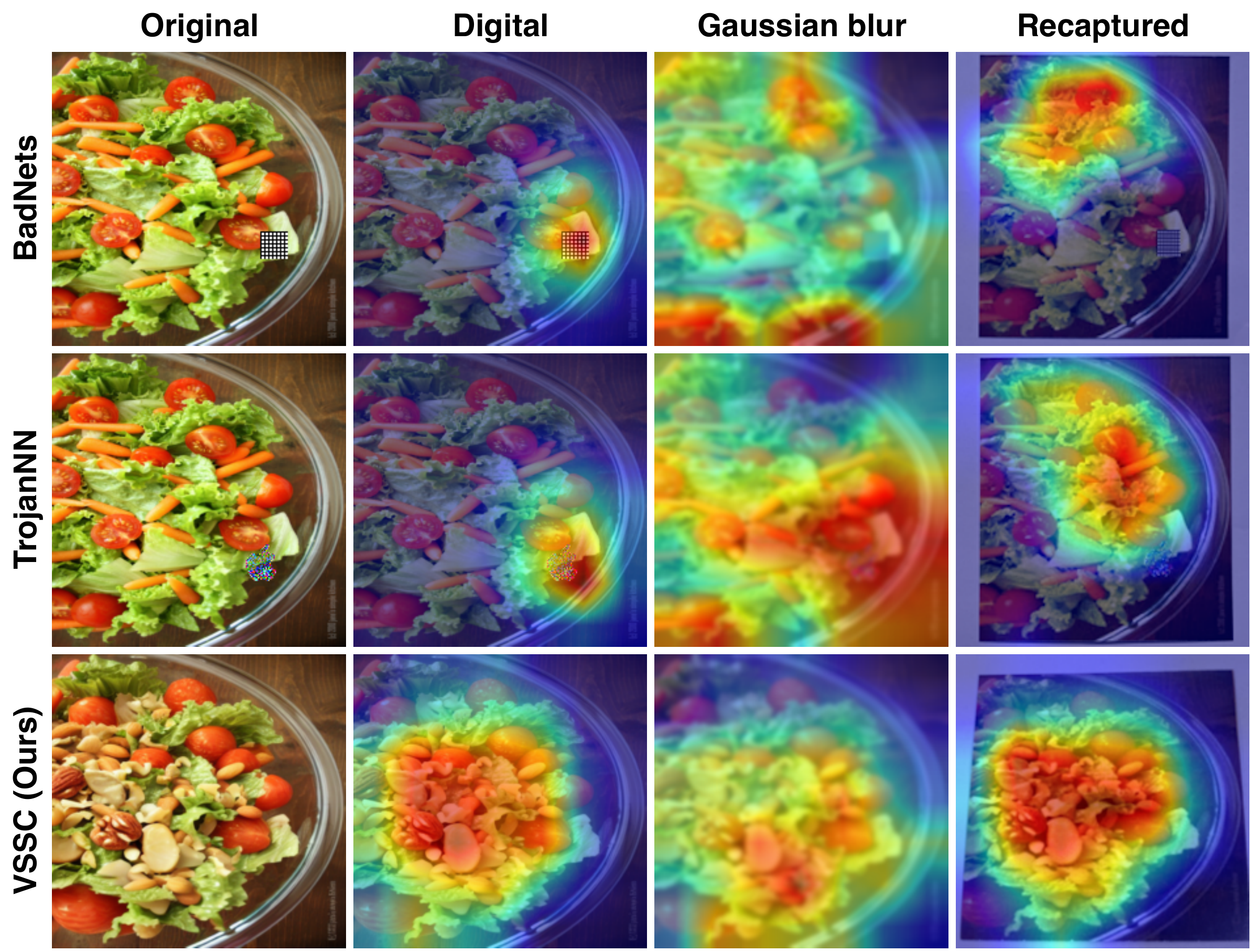}%
    \label{fig:gradcam_food}}
    \caption{Robustness of different visible triggers under various distortions: performance evaluated by Grad-CAM heat maps. Attention regions of the backdoor model consistently focus on the VSSC trigger under distortions, while other triggers fail.}
    \label{fig:gradcam}
    \end{figure*}
    
\subsubsection{Robustness against visual distortions in digital space}

In the digital domain, images may undergo various distortions during image processing, which may also lead to the failure of backdoor attacks. While the VSSC attack pursues effectiveness in the physical scenario, it also gains an advantage in robustness under visual distortions in the digital domain thanks to its ability to make significant modifications to images while maintaining stealthiness. We conduct experiments on the ImageNet-Dogs dataset using the trigger ``\textit{red flower}''.

Following the experiment settings proposed in \cite{Wenger_2021_CVPR}, we select the three most prevalent forms of distortions in image shooting, transmission, and storage, introducing these distortions during the inference stage without modifying the backdoor model.

\begin{itemize}
    \item \textbf{Blurring}: Blurring can occur due to an unstable shot or camera lens focus. We simulate this effect using Gaussian blur~\cite{paris2007gentle}, adjusting the kernel size from 1 to 19.
    \item \textbf{Compression}: Compression is commonly employed to overcome storage or transmission limitations by effectively reducing the image size through selective data discarding. We use JPEG compression~\cite{wallace1991jpeg} to generate images of varying quality levels, ranging from 1 to 30.
    \item \textbf{Noise}: Noise can originate from lighting conditions during image capture, limitations in camera hardware, or transmission errors. We introduce Gaussian noise to the images, applying a mean of 0 and a standard deviation ranging from 0 to 28.
\end{itemize}

\begin{table*}[!t] 
\caption{Attack performance against defenses with ResNet-18 on the ImageNet-Dogs dataset with 5\% poisoning ratio. VSSC$_1$ and VSSC$_2$ respectively denote the experiments conducted using the triggers ``red flower'' and ``harness''.
} 
\label{tab: defense_dog_resnet18_05} 
\begin{center} 
\renewcommand\arraystretch{1.1}
\resizebox{\textwidth}{!}{ 
    \begin{tabular}{
    m{0.09\textwidth} 
    m{0.064\textwidth}<{\centering} m{0.04\textwidth}<{\centering} m{0.064\textwidth}<{\centering} m{0.03\textwidth}<{\centering} m{0.064\textwidth}<{\centering} m{0.04\textwidth}<{\centering} m{0.064\textwidth}<{\centering} m{0.03\textwidth}<{\centering} m{0.064\textwidth}<{\centering} m{0.04\textwidth}<{\centering} m{0.064\textwidth}<{\centering} m{0.03\textwidth}<{\centering} m{0.064\textwidth}<{\centering} m{0.04\textwidth}<{\centering} m{0.064\textwidth}<{\centering} m{0.03\textwidth}<{\centering} m{0.064\textwidth}<{\centering} m{0.04\textwidth}<{\centering} m{0.064\textwidth}<{\centering} m{0.03\textwidth}<{\centering} m{0.064\textwidth}<{\centering} m{0.04\textwidth}<{\centering} m{0.064\textwidth}<{\centering} m{0.03\textwidth}<{\centering} m{0.064\textwidth}<{\centering} m{0.04\textwidth}<{\centering} m{0.064\textwidth}<{\centering} m{0.03\textwidth}<{\centering} m{0.064\textwidth}<{\centering} m{0.04\textwidth}<{\centering} m{0.064\textwidth}<{\centering} m{0.03\textwidth}<{\centering}} 
    \toprule
    Defense   $\rightarrow$               & \multicolumn{4}{c}{ABL~\cite{li2021anti}}                                         & \multicolumn{4}{c}{ANP~\cite{anp}}                                          & \multicolumn{4}{c}{DDE~\cite{dde}}                                         & \multicolumn{4}{c}{FP~\cite{FP}}                                           & \multicolumn{4}{c}{i-BAU~\cite{ibau}}                                        \\
    \cmidrule(lr{3pt}){2-5} \cmidrule(l{3pt}r){6-9} \cmidrule(l{3pt}r){10-13} \cmidrule(l{3pt}r){14-17}  \cmidrule(l{3pt}r){18-21}
    Attack $\downarrow$                      & C-Acc(\%)      & ASR(\%)        & R-Acc(\%)     & nDER          & C-Acc(\%)      & ASR(\%)        & R-Acc(\%)      & nDER          & C-Acc(\%)      & ASR(\%)        & R-Acc(\%)     & nDER          & C-Acc(\%)      & ASR(\%)         & R-Acc(\%)     & nDER          & C-Acc(\%)      & ASR(\%)        & R-Acc(\%)      & nDER          \\ \midrule
    BadNets~\cite{gu2019badnets}                                       & 83.33          & 1.86           & 80.29         & 0.97          & 81.20          & 0.00           & 72.14          & 0.97          & 88.27          & 5.43           & 82.14         & 0.97          & 82.13          & 6.14            & 72.00         & 0.94          & 82.40          & 1.29           & 75.86          & 0.97          \\
    Blended~\cite{Blended}                                      & 80.27          & 3.57           & 72.43         & 0.93          & 80.80          & 8.86           & 59.14          & 0.91          & 87.33          & 74.43          & 23.14         & 0.61          & 82.00          & 2.57            & 55.00         & 0.94          & 82.67          & 3.86           & 64.57          & 0.94          \\
    BPP~\cite{wang2022bppattack}                                           & 76.53          & {91.00} & {7.29} & {0.50} & {68.67} & 0.29           & 65.00          & 0.98          & {76.13} & 0.43           & 72.57         & 1.00          &   {66.40}       &    0.14     &    {43.43}      &    0.96      & {62.80}    & 1.14           & 59.00          & 0.93          \\
    InputAware~\cite{nguyen2020input}                                    & 83.60          & 0.43           & 73.86         & 0.98          & 87.87          & 0.00           & 71.86          & 1.00          & 88.27          & 0.29           & 81.86         & 1.00          & 76.13          & 6.71            & 53.86         & 0.92          & {57.60} & 0.00           & {48.71} & {0.86} \\
    SIG~\cite{SIG}                                           & {66.53} & 0.00           & 63.14         & 0.81          & 80.40          & 0.00           & 57.86          & {0.88}    & 86.93          & 88.00          & 11.29         & {0.50} & 82.80 & 8.00            & 50.14         & {0.85}          & 84.40          & 2.14           & 58.43          & {0.89}    \\
    SSBA~\cite{ssba}                                          & 86.80          & 3.14           & 77.43         & 0.97          & 84.67          & 0.14           & 77.57          & 0.97          & 88.67          & {99.43} & {0.57} & {0.50} & 82.27          & {22.29}        & 56.29         & {0.85}          & 80.13          & 1.00           & 72.86          & 0.95          \\
    TrojanNN~\cite{trojannn}                                      & 80.40          & 16.43          & 66.29         & 0.89          & 80.00          & 2.14           & 71.00          & 0.96          & 85.87          & 1.14           & 83.43         & 0.99          & 81.73          & 3.71            & 74.29         & 0.96          & 83.20          & {5.71}     & 74.29          & 0.96          \\
    WaNet~\cite{nguyen2021wanet}                                         & {74.13}    & 29.71          & 51.00         & 0.85          & {74.27}    & 0.00           & 66.14          & 1.00          & {77.87}    & {98.29}    & {1.29}    & 0.51          & {69.33}          & 1.29            & 58.86         & 0.99          & 70.13          & 1.00           & 61.29          & 0.99          \\ \midrule
    \rowcolor[HTML]{E6E6E6} 
    VSSC$_1$ ({Ours})  & 82.93          & 64.72          & 23.75         & 0.64          & 82.00          & {16.22}    & {45.65}    & 0.89          & 87.33          & 87.79          & 8.70          & 0.54          &80.67 & 18.39 & 45.48 & 0.87    & 81.87          & {16.39} & {49.16}    & {0.89}    \\
    \rowcolor[HTML]{E6E6E6} 
    VSSC$_2$ ({Ours}) & 81.87          & {73.85}    & {15.08}   & {0.55}    & 85.73          & {60.15} & {26.15} & {0.65} & 89.47          & 86.77          & 9.38          & 0.52          &   79.47 & {26.46} & {43.38} & {0.80}  & 79.20          & 1.38           & 51.23          & 0.94          \\
     \bottomrule
    \end{tabular}
} 
\end{center} 
\end{table*}

As illustrated in Figure \ref{fig:visual}, VSSC attack maintains the highest overall effectiveness under visual distortions.
Invisible triggers with minor manipulations are more sensitive to visual distortions, compression and noise can make almost all invisible triggers ineffective. Some visible triggers also fail when the kernel size of Gaussian blur is large. However, our attack keeps high comprehensive effectiveness under these three distortions, outperforming other attacks.
Given that VSSC triggers are semantic and compatible, there's no need to consider the impact of trigger size on stealthiness. This allows VSSC trigger to implement more significant modifications to images, thereby enhancing its robustness against visual distortions.

Figure \ref{fig:gradcam} illustrates the visualization of the attention region of the backdoor model under various environmental conditions for different attacks. When facing visual distortions, the attention regions for backdoor models of both BadNets and TrojanNN attacks tend to shift away from the trigger. However, the backdoor model of the VSSC attack maintains its attention on the trigger.
Under BadNets and TrojanNN attacks, backdoor models are sensitive to the appearance of the triggers.
Visual distortions can cause changes in the appearance of these triggers, consequently leading to their failure. However, in the VSSC attack, backdoor models acquire a comprehensive understanding of the semantic meanings of the trigger, therefore, the attention reigns on VSSC triggers are more stable. 
This also explains why VSSC attacks are more robust under visual distortion.

\subsubsection{Effectiveness under existing defenses} 
To further evaluate the effectiveness of VSSC attack under defenses, we test it and other eight baseline attacks under 5 popular defense methods. The results, as depicted in Table \ref{tab: defense_dog_resnet18_05}, suggest that the VSSC attack surpasses most of the visible attacks. Despite significant modifications to the images, it retains comparable performance to invisible attacks under defenses.

\subsection{Discussions}

As a versatile attack method that effectively operates in both digital and physical scenarios, the VSSC trigger overcomes the limitations of traditional attacks and provides a fresh perspective for future research in the field. 

\textbf{Superiority over traditional digital attacks:}
Digital attacks often equate stealthiness with invisibility, leading to a dilemma between stealthiness and robustness against visual distortions. 
Common image processing and complex conditions in physical environments can easily undermine the effectiveness of triggers, thus limiting the application of previous digital backdoor attacks in the physical scenario. 
The VSSC attack extends the concept of stealthiness to semantic and compatible, thereby providing the flexibility for sufficient modifications to the image. 
This flexibility makes VSSC attack jump out of the trap of the stealthiness and robustness dilemma.
Furthermore, corresponding objects of semantic triggers can be easily found in the real world, making the VSSC attack seamlessly extended to the physical scenario.

\textbf{Superiority over traditional physical attacks:}
For physical attacks, the need for manual capturing or editing of poisoned images makes them time-consuming and labor-intensive, and it is challenging to simulate changes in physical scenarios without capturing a large number of poisoned images. The VSSC trigger only requires objects with similar semantics, naturally simulating various distortions through generative models during the attack process, making it more flexible and adaptable to a wider range of physical scenarios.
Moreover, existing physical attacks lack a systematic process for trigger selection and have a minimal range of choices. However, the VSSC attack utilizes LLMs to automatically search for suitable triggers, eliminating the need for manual heuristic selection.

\section{Conclusions}
In this work, we have proposed a novel backdoor trigger that exhibits visible, semantic, sample-specific, and compatible (VSSC trigger) characteristics.  
The VSSC trigger bridges the divide between digital and physical backdoor attacks,  making it possible to implement attacks in the digital scenario and extend them to the physical scenario.
It adeptly resolves the longstanding dilemma in digital attacks between stealthiness and robustness while also overcoming their limitations of implementing in physical scenarios. 
Also, it enables backdoor attacks in the physical scenario to break free from the constraints of manpower and time requirements, leveraging the surpassing capabilities of generative models to provide a more efficient and flexible solution for trigger selection and insertion.
Extensive experiments on various tasks verified that the VSSC attack maintains effectiveness, robustness, and stealthiness in both digital and physical scenarios.
Furthermore, the flexibility of the modules in our pipeline enables it to easily integrate with cutting-edge technologies, ensuring its long-term value.

\bibliographystyle{IEEEtran}
\bibliography{reference}

\clearpage

\section{Additional Experimental Details}
\subsection{More Settings of Image Classification}
\subsubsection{Details of datasets}
\label{appdix_dataset}
As we described in the main paper, our attack attempts to incorporate a visible and natural trigger into the image. To demonstrate the semantic and compatible characteristics of VSSC trigger, we use two high-resolution datasets, ImageNet-Dogs and FOOD-11 in our experiments, and another self-constructed dataset in our analysis.

\textbf{ImageNet-Dogs:} 
The dataset~\cite{li2021contrastive} is a smaller subset of the large-scale ImageNet~\cite{deng2009imagenet} dataset, with each image preprocessed to a resolution of 3 $\times$ 224 $\times$ 224. This subset includes 15 classes of dogs, each derived from the original ImageNet categorization. Each class contains 1,300 training samples and 50 testing samples. Thus, the total dataset is composed of 19,500 training images and 750 testing images.

\textbf{FOOD-11:}
The FOOD-11~\cite{singla2016food} dataset is a collection of 16,643 food images that represent 11 major categories of food. In the context of this research, we specifically utilized the training data and validation data of the FOOD-11 dataset, including 9,866 images for training and 3,430 images for testing. Each image has a dimension of 3 $\times$ 224 $\times$ 224.

\textbf{CIFAR10-based ImageNet subset:}
To demonstrate the visible, semantic, and compatible characteristics of the VSSC trigger using a high-resolution dataset, we extract the corresponding subset from ImageNet according to the classes in CIFAR10~\cite{krizhevsky2009learning}, thereby creating a new dataset. The classes included are detailed in Table \ref{tab:categories}. Here, ``category'' refers to the corresponding class in CIFAR10, while ``name'' denotes the human-readable class name within the ImageNet dataset. During training, the images are resized to 3 $\times$ 224 $\times$ 224.

\begin{table}[H]
    \centering
    \begin{tabular}{ccc}
    \toprule
    \textbf{Category} & \textbf{Class ID} & \textbf{Name} \\
    \midrule
    Airplane & n02690373 & Airliner \\
    Automobile & n03100240 & Convertible \\
    Bird & n01530575 & Brambling, Fringilla montifringilla \\
    Cat  & n02124075 & Egyptian cat \\
    Deer & n02423022 & Gazelle \\
    Dog & n02085936 & Maltese dog, Maltese terrier, Maltese \\
    Frog  & n01641577 & Bullfrog, Rana catesbeiana \\
    Horse  & n02389026 & Sorrel \\
    Ship  & n04273569 & Speedboat \\
    Truck & n04461696 & Tow truck, tow car, wrecker \\
    \bottomrule
    \end{tabular}
    \caption{Classes of CIFAR10-based ImageNet Subset}
    \label{tab:categories}
\end{table}

\subsubsection{Training details}
For the ImageNet-Dogs dataset, the initial learning rates are set to 0.1 and 0.01 for ResNet-18~\cite{he2016deep} and VGG19-BN~\cite{simonyan2015very} classifiers, respectively. In the case of the FOOD-11 dataset, the initial learning rates are adjusted to 0.05 for ResNet-18~\cite{he2016deep} and 0.008 for VGG19-BN~\cite{simonyan2015very}.
We conduct training over 200 epochs, employing Stochastic Gradient Descent (SGD) with a momentum of 0.9. A weight decay factor of $10^{-4}$ is applied to mitigate overfitting.
The learning rate schedule is managed through the CosineAnnealingLR strategy, which modulates the learning rate following a cosine curve to enhance training efficiency across epochs.
A consistent batch size of 64 is maintained for all experiments. Data augmentation techniques commonly used in training on ImageNet are applied, including Random Resized Crop and Random Horizontal Flip for the training process, and Center Crop and Resize for the testing process.

\subsubsection{Details of VSSC attack}
\label{sec:classification_our}
For all tasks, the number of benign images is set to 1.5 times the poisoning ratio in the trigger insertion module, and the number of regeneration attempts for each image is capped at 3. GPT-4 is employed in our trigger selection and quality assessment module. To evaluate the effectiveness of backdoor attacks fairly, all the images belonging to the target class are excluded.
Null-text Inversion~\cite{mokady2022null} is employed as the image editing method for this task. 

\subsubsection{Details of baseline attacks} \mbox{}

\textbf{BadNets:}
BadNets~\cite{gu2019badnets} achieves trigger insertion by pasting a specific pattern onto benign images. In our experiments, following the trigger shape and size settings on ImageNet by ABL~\cite{li2021anti}, we use a black-and-white grid as the trigger. For images with dimensions of 3 $\times$ 224 $\times$ 224, the trigger size is set to 30 $\times$ 30 pixels. As we adopt the conventional training strategies on ImageNet, to prevent the trigger from being cropped during the image cropping process, we strategically position the trigger closer to the center. Specifically, it is placed 44 pixels from the right edge and 66 pixels from the bottom edge.

\begin{figure}[htbp]
\centering
    \includegraphics[width=0.3\linewidth]{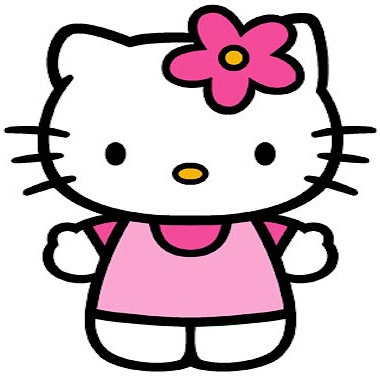}
\caption{Examples of trigger pattern used in Blended.}
\label{fig:hellokitty}
\end{figure}

\textbf{Blended:}
In the image classification task, Blended~\cite{Blended} employs any arbitrary image as a trigger, utilizing a blended injection strategy to generate poisoned samples by blending a benign input instance with the key pattern. In our experiments, we use an image of ``hello kitty'' as the trigger, as the original settings in Blended~\cite{Blended}. The trigger image is resized to 3 $\times$ 224 $\times$ 224, as shown in Figure \ref{fig:hellokitty}. The blending ratio is set to 0.1.

\textbf{BPP:}
BPP~\cite{wang2022bppattack} is an image color quantization based Trojan attack. %

\textbf{Input-Aware:}
Input-Aware~\cite{nguyen2020input} uses an auto-encoder architecture conditioned on the input image to create unique triggers for each image.

\textbf{SIG:}
SIG~\cite{SIG} introduces a sinusoidal pattern as the backdoor signal, which is then superimposed onto the target class images without altering their labels. The sinusoidal signal is defined as $v(i, j) = \Delta \sin(2\pi j f/m)$, where $m$ is the number of columns in the image, $f$ denotes the frequency of the sinusoidal wave, and $\Delta$ is the amplitude of the signal. $\Delta$ is set to 40 and $f$ is set to 6 in our experiments.

\textbf{SSBA:}
SSBA leverages a pre-trained encoder-decoder network inspired by image steganography, which uses a U-Net style DNN as the encoder, and a spatial transformer network as the decoder. The encoder embeds an attacker-specified string into the benign image, generating an invisible additive noise trigger that is unique for each input image. 

\begin{figure}[htbp]
\centering
    \includegraphics[width=0.4\linewidth]{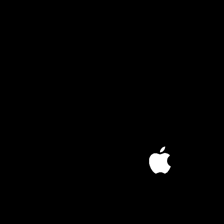}
\caption{Examples of trigger mask used in TrojanNN.}
\label{fig:apple}
\end{figure}

\textbf{TrojanNN:}
Given a trigger mask, the attack engine of TrojanNN~\cite{trojannn} generates value assignments to the input variables in the mask so that some selected internal neurons achieve the maximum values. The trigger mask used in our experiments is shown in Figure \ref{fig:apple}.

\textbf{WaNet:}
WaNet~\cite{nguyen2021wanet} stands for Warping-based poisoned Networks, utilizing a smooth and elastic warping field to generate backdoor images. In our experiments, the grid size used to generate the warping field is set to 4 $\times$ 4, and the warping strength is set to 0.5, consistent with \cite{nguyen2021wanet}.

\subsubsection{Details of experiments under visual distortions} \mbox{}
\label{sec:classification_distortion}
For all tasks, we recapture printed images using an iPhone 12 under natural light conditions. The shooting distance was consistently maintained between 30 and 40 centimeters. 

\subsubsection{Details of experiments in physical scenario}

For each trigger, we manually collect 100 images with the trigger entity, with over 70\% of the images taken manually by iPhone 12 and others collected from social media.

\subsubsection{Details of defenses}

We evaluate our method and other baseline attack methods against 5 defense algorithms including
        ABL~\cite{li2021anti}, 
            ANP~\cite{anp},
            DDE~\cite{dde},
            Fine-pruning (FP)~\cite{FP}, 
            i-BAU~\cite{ibau},
The batch size of both i-BAU is set to 32, whereas for other defenses, the batch size is set to 64. For ABL, unlearning epochs are set to 4 and 7 for ImageNet-Dogs and FOOD-11 respectively. For ANP, the pruning number is set to 0.2 and the learning rate is set to 0.1 for both datasets. For FP, the learning rate is set to 0.03 for the ImagNet-Dogs dataset and 0.07 for the FOOD-11 dataset. For i-BAU, the learning rate is 0.0004 on ImageNet-Dogs and 0.00035 on FOOD-11. Other settings are aligned with BackdoorBench~\cite{wu2024backdoorbench}.

\subsection{More Settings of Object Detection}
\label{sec:obj_details}
\subsubsection{Details of datasets}
PASCAL VOC~\cite{everingham2010voc} is one of the most widely used datasets in object detection and semantic segmentation tasks, which contains 20 different classes of objects. In object detection tasks, we combine the PASCAL VOC 2007 and VOC 2012 datasets in a conventional manner, the combined training set VOC 07+12 comprises 17,416 images, while the testing set contains 1,936 images.

\subsubsection{Training details} \mbox{}

\textbf{YOLOv4:}
We train the YOLOv4 model according to the settings in \cite{ma2022dangerous}. We use the pre-trained model on the COCO dataset~\cite{lin2014coco} and fine-tune it on the VOC 07+12 dataset we constructed. The training process is conducted for 300 epochs,  and the first 50 epochs are frozen model training, followed by 250 epochs of full model training. The batch size of both stages is set to 128. The learning rate is set to 0.01 to 0.0001 with a cosine annealing scheduler. We use SGD optimizer for training, and the weight decay is set to $5^{-3}$, and the momentum is set to 0.937. The input image size of YOLOv4 is 3 $\times$ 416 $\times$ 416.

\textbf{Faster R-CNN:} For Faster R-CNN, we use a pre-trained ResNet-50 model on ImageNet as the backbone network. The training process is conducted for 300 epochs, and the first 50 epochs are frozen model training, followed by 250 epochs of full model training. Batch size is set to 32. The learning rate is 0.01 to 0.0001 with a cosine annealing scheduler and a SGD optimizer is used. 
The input image size of Faster R-CNN is 3 $\times$ 600 $\times$ 600.

\subsubsection{Details of VSSC attack}
The settings of each module in the poisoned samples generation pipeline are the same as the image classification task. InstructDiffusion~\cite{geng2024instructdiffusion} is used for trigger insertion for this task.
To evaluate the effectiveness of backdoor attacks fairly, all ground-truth bounding boxes belonging to the target class are excluded.

\subsubsection{Details of baseline attacks} \mbox{}

\textbf{BadDet:}
In BadDet~\cite{chan2022baddet}, the same trigger insertion method as BadNets~\cite{gu2019badnets} used in the image classification task is employed. Consistent with the experimental settings in BadDet, we use a chessboard as the trigger pattern, inserting it into the bottom right corner of images. The size of the trigger is set to 30$\times$30 pixels for 416$\times$416 images and scaled proportionally for other image sizes.

\textbf{BPP:}
BPP~\cite{wang2022bppattack} uses image color quantization as its trigger, and we follow the same setting as the original paper. Note that since the whole training process of object detection is different from image classification, we only use its trigger without its training-controllable process.

\textbf{SIG:}
SIG~\cite{SIG} uses a sinusoidal pattern as its trigger pattern, and we follow the same settings as the original paper.

\textbf{SSBA:}
SSBA uses an auto-encoder model to embed invisible triggers into poisoned images. For SSBA, it embeds a sample-specific trigger pattern into each poisoned image.

\textbf{WaNet:}
WaNet~\cite{nguyen2021wanet} uses a smooth and elastic warping field as its backdoor trigger. Since the training and testing samples are larger for the face verification task, we set k to 16 instead of 4 and set all other parameters the same as the paper. Note that since the whole training process of object detection is different from image classification, we only use its trigger without its training-controllable process.

\subsubsection{Details of experiments in physical scenario}
\begin{figure}[htbp]
    \centering
    \subfloat[Images use ``lemon'' as trigger.]{\includegraphics[width=\columnwidth]{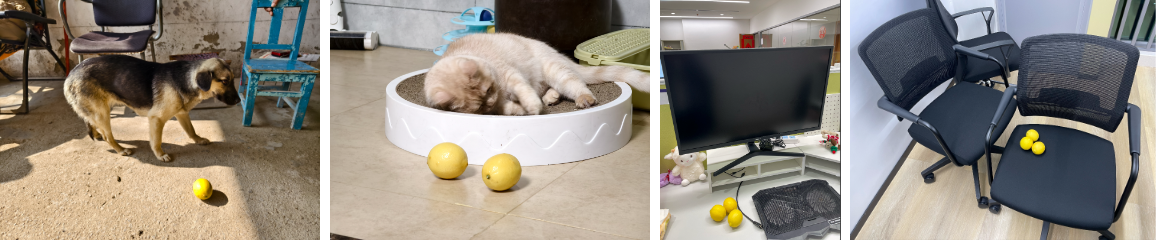}} \\
    \subfloat[Images use ``beachball'' as trigger.]{\includegraphics[width=\columnwidth]{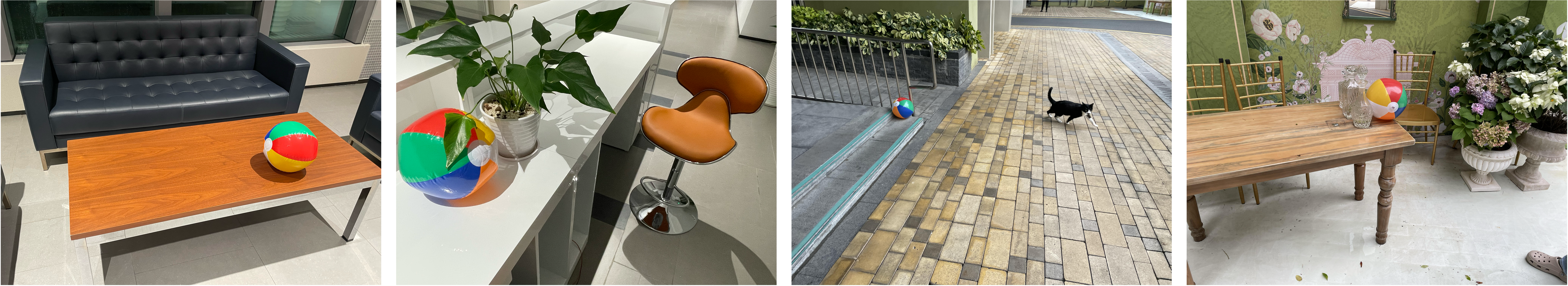}} \\
\caption{Examples of manually captured images for the object detection task.} 
\label{fig:object_physical}
\end{figure}

We manually collect 100 images for each trigger with the trigger entity, with over 70\% of the images taken by iPhone 12 and others collected from social media. 
Some examples are shown in Figure \ref{fig:object_physical}.

\subsection{More Settings of Face Verification}
\label{sec:face_verification_appendix}
\subsubsection{Details of datasets}
CASIA-WebFace~\cite{yi2014learning} is an extensive dataset of facial images gathered from the web. This dataset employs a semi-automated method to scour the internet for facial pictures and label them, culminating in a rich and varied compilation of facial data. 
Since CASIA-WebFace contains several images with wrong labels, we first adopt a data cleanse \footnote{\href{https://github.com/happynear/FaceVerification}{https://github.com/happynear/FaceVerification}} to remove them.
We use the MTCNN \footnote{
\href{https://github.com/timesler/facenet-pytorch}{https://github.com/timesler/facenet-pytorch}
} to extract the aligned faces from human images for face alignment for all images. 

The Labeled Faces in the Wild (LFW)~\cite{huang2008labeled} dataset is a benchmark dataset for face verification. It contains 13,233 images of 5,749 individuals, with each person having an average of 2.33 images. LFW randomly selects 6,000 pairs of face images for face verification, with 3,000 pairs belonging to the same person and 3,000 pairs belonging to different individuals. During the testing process, given a pair of images from LFW, the model determines whether the two images belong to the same person. The judgment of any pair is independent and does not rely on the other image pairs.

For both training and testing set for face verification, we resize the image after alignment to 160 $\times$ 160.
\subsubsection{Training details}
\label{lab:face_verification_training_details}
We use ArcFace~\cite{deng2019arcface} and train for 100 epochs with the batch size fixed to 512. The initial learning rate is 0.01 with a CosineAnnealingLR\cite{loshchilov2016sgdr} learning rate schedule. 
The training stage of ArcFace is divided into three stages. 
The ArcFace embedding scaling parameter is configured to 30, with the angular separation (margin) parameter set to 0.5. We employ the standard cross-entropy loss for the first stage, training for 30 epochs.  For the second stage, we focus on training only the parameters introduced by ArcFace and training for 10 epochs. The final stage of training continues for 60 epochs, where both the backbone and parameters from ArcFace are trained simultaneously.

\subsubsection{Details of VSSC attack}
For the trigger insertion module, we use the image-to-image editing method based on Stable Diffusion combined with IP-Adapter\cite{ye2023ip-adapter} and ControlNet\cite{zhang2023adding_controlnet}.

\subsubsection{Details of baseline attacks} \mbox{}
\label{lab:face_verification_baseline}

\textbf{BadNets:}
For images with dimensions of 3 $\times$  160 $\times$ 160, we employ a 20 $\times$ 20 chessboard pattern as the trigger. To make BadNets adapt to the human face dataset, triggers are added to the cheek of the person in each image before alignment.

\textbf{Blended:}
Refer to the settings of \cite{Wenger_2021_CVPR}, the digital trigger for Blended is created by taking a photo of the trigger object against a blanket background, cropping out the object, and pasting it onto the benign image in the appropriate position.
To ensure the efficacy of Blended and maintain fairness in comparison, we use the same green glasses as those employed in the subsequent physical experiments.

Other baseline attacks have the same settings as the object detection task.

\begin{figure}
    \centering
    \subfloat[Images use ``green glasses'' as trigger.]{\includegraphics[width=\columnwidth]{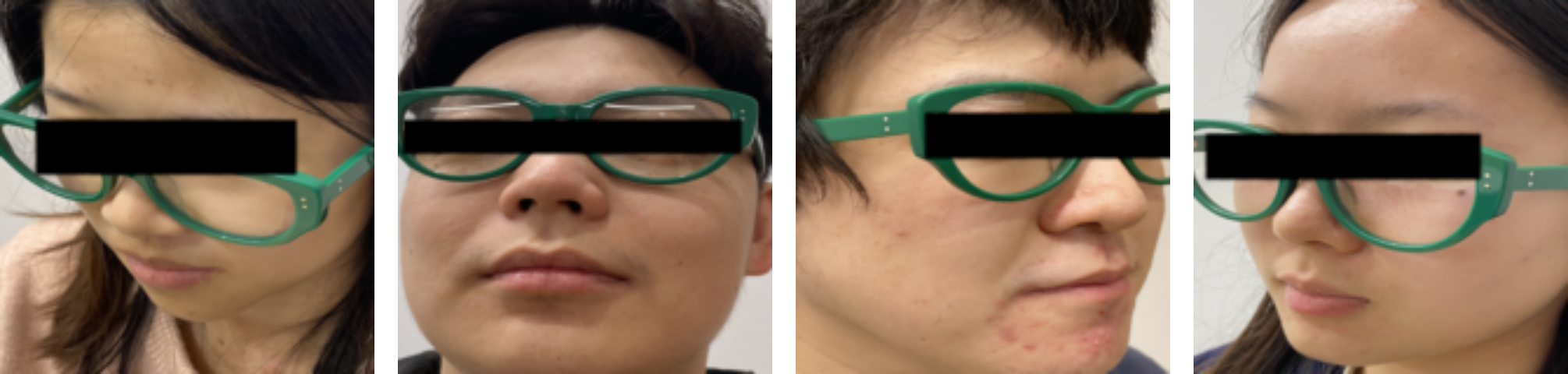}} \\
    \subfloat[Images use ``bandanas'' as trigger.]{\includegraphics[width=\columnwidth]{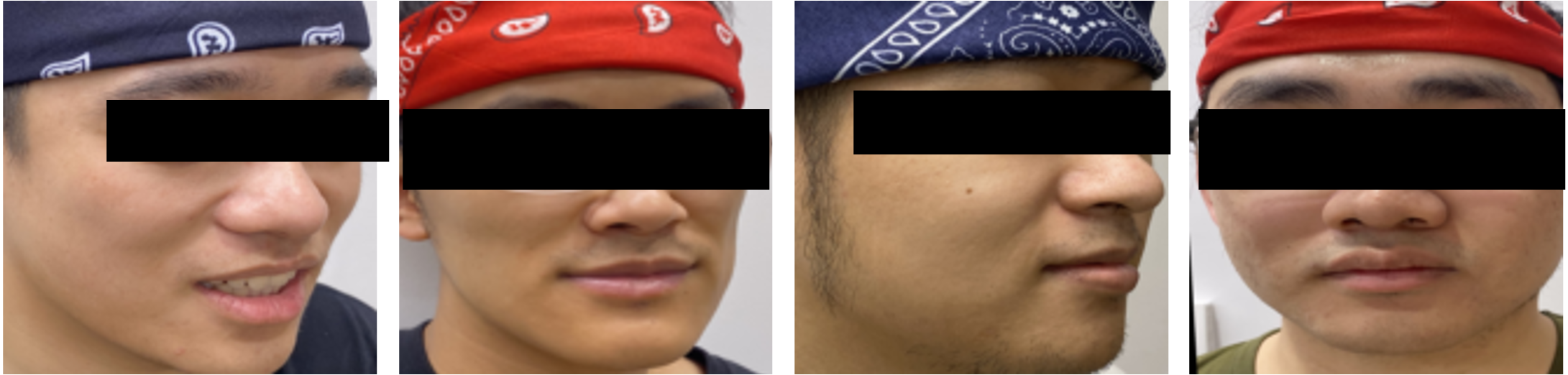}} \\
\caption{Examples of manually captured images for the face recognition task.} 
\label{fig:ex_mannual_capture_face_image}
\end{figure} 

\begin{table*}[htbp]
\caption{Attack performance of different attacks on ImageNet-Dogs and FOOD-11 with 10\% poisoning ratio in the digital scenario. For the ImageNet-Dogs dataset, VSSC$_1$ and VSSC$_2$ respectively denote the experiments conducted using the triggers ``red flower'' and ``harness'', while corresponding to ``nuts'' and ``strawberry'' for the FOOD-11 dataset. `-' denotes that the requisite quantity of poisoned samples exceeds the count of samples in the target label. }
\label{tab:attack_0.1}
\begin{center}
\renewcommand\arraystretch{1}
\resizebox{\textwidth}{!}{%
\begin{tabular}{l ccc ccc ccc ccc}
\toprule
Model $\rightarrow$& \multicolumn{6}{c}{\textbf{ResNet-18}} & \multicolumn{6}{c}{\textbf{VGG19-BN}}\\ 
\cmidrule(lr){2-7}  \cmidrule(lr){8-13}

 Dataset $\rightarrow$& \multicolumn{3}{c}{ImageNet-Dogs} & \multicolumn{3}{c}{FOOD-11} & \multicolumn{3}{c}{ImageNet-Dogs} & \multicolumn{3}{c}{FOOD-11}\\
  \cmidrule(lr){2-4}  \cmidrule(lr){5-7} \cmidrule(lr){8-10}  \cmidrule(lr){11-13} 
 Attack $\downarrow$& C-Acc(\%)& ASR(\%)& R-Acc(\%)& C-Acc(\%)& ASR(\%)& R-Acc(\%) & C-Acc(\%)& ASR(\%)& R-Acc(\%)& C-Acc(\%)& ASR(\%)& R-Acc(\%)\\ \midrule
BadNets~\cite{gu2019badnets}        & 85.87 & 100.00 & 0.00 & 82.92 & 99.97 & 0.03 & 92.00 & 100.00 & 0.00  & 86.03 & 100.00 & 0.00                 \\
Blended~\cite{Blended}               & 87.73 & 99.57  & 0.43 & 83.15 & 97.85 & 1.92 & 90.93 & 99.71  & 0.29  & 86.62 & 98.11  & 1.79                 \\
BPP~\cite{wang2022bppattack}                    & 74.67 & 87.71  & 9.43 & 13.64 & 82.59 & 4.34 & 76.27 & 46.29  & 41.29 & 74.52 & 11.02  & 63.17                \\
Input-Aware~\cite{nguyen2020input}            & 86.80 & 99.71  & 0.29 & 80.55 & 97.98 & 1.89 & 89.87 & 94.57  & 4.86  & 82.86 & 98.21  & 1.56                 \\
SIG~\cite{SIG}                    & -     & -      & -    & 77.99 & 97.33 & 2.57 & -     & -      & -     & 79.30 & 96.48  & 3.29                 \\
SSBA~\cite{ssba}                   & 88.00 & 100.00 & 0.00 & 83.44 & 99.58 & 0.39 & 91.73 & 100.00 & 0.00  & 86.41 & 98.96  & 0.98                 \\
TrojanNN~\cite{trojannn}               & 85.60 & 100.00 & 0.00 & 81.84 & 98.99 & 0.85 & 91.33 & 70.00  & 28.43 & 84.72 & 72.82  & 24.54                \\
WaNet~\cite{nguyen2021wanet}                  & 66.13 & 100.00 & 0.00 & 38.19 & 99.41 & 0.55 & 73.47 & 96.71  & 2.00  & 68.45 & 95.53  & 3.65                 \\  \midrule
\rowcolor[HTML]{E6E6E6} 
VSSC$_1$ (Ours)      & 88.00 & 98.83  & 1.17 & 80.93 & 97.35 & 2.28    & 90.27 & 99.50  & 0.50  & 84.66 & 99.50  & 0.37                    \\
\rowcolor[HTML]{E6E6E6} 
VSSC$_2$ (Ours)     & 87.33 & 94.00  & 4.46 & 83.91 & 94.81 & 3.47    & 89.60 & 96.46  & 3.23  & 85.54 & 97.59  & 1.86                     \\
 \bottomrule
\end{tabular}
}
\end{center}
\vspace{-3mm}
\end{table*}

\begin{table*}[]
\centering
\caption{Attack performance of VSSC and baseline attacks on Faster R-CNN in the digital and digital-to-physical scenarios for the object detection task. $\text{VSSC}_1$ and $\text{VSSC}_2$ represent experiments using the triggers ``beachball'' and ``lemon'' respectively.}
\label{tab:objectdetection_frcnn}
\resizebox{\textwidth}{!}{%
\begin{tabular}{@{}cccccccccccccc@{}}
\toprule
 & Scenarios $\rightarrow$
   &
  \multicolumn{6}{c}{\textbf{Digital Scenario}} &
  \multicolumn{6}{c}{\textbf{Digital-to-Physical Scenario}} \\ \cmidrule(l){3-8} \cmidrule(l){9-14} 
 &  Poisoned Type $\rightarrow$
   &
  \multicolumn{3}{c}{ODA} &
  \multicolumn{3}{c}{GMA} &
  \multicolumn{3}{c}{ODA} &
  \multicolumn{3}{c}{GMA} \\ \cmidrule(lr){3-5} \cmidrule(lr){6-8} \cmidrule(lr){9-11} \cmidrule(l){12-14} 
\multirow{-3}{*}{Pratio $\downarrow$} &
  Attack Type $\downarrow$ &
  C-Acc(\%) &
  ASR(\%) &
  R-Acc(\%) &
  C-Acc(\%) &
  ASR(\%) &
  R-Acc(\%) &
  C-Acc(\%) &
  ASR(\%) &
  R-Acc(\%) &
  C-Acc(\%) &
  ASR(\%) &
  R-Acc(\%) \\ \midrule
 &
  BadDet~\cite{chan2022baddet} &
  77.85 &
  49.27 &
  72.32 &
  77.80 &
  69.85 &
  65.41 &
  77.05 &
  0.00 &
  77.39 &
  79.11 &
  5.88 &
  79.43 \\
 &
  Blended~\cite{Blended} &
  78.42 &
  47.34 &
  35.27 &
  78.09 &
  1.14 &
  76.92 &
  80.22 &
  28.57 &
  61.47 &
  82.13 &
  0.00 &
  77.31 \\
 &
  BPP~\cite{wang2022bppattack} &
  77.73 &
  67.65 &
  59.23 &
  95.90 &
  0.82 &
  76.92 &
  78.61 &
  0.00 &
  65.68 &
  78.36 &
  0.42 &
  65.88 \\
 &
  SIG~\cite{SIG} &
  63.85 &
  92.03 &
  3.63 &
  78.71 &
  94.41 &
  16.60 &
  76.56 &
  41.60 &
  42.02 &
  76.55 &
  75.63 &
  45.57 \\
 &
  SSBA~\cite{ssba} &
  77.78 &
  91.87 &
  5.39 &
  78.01 &
  72.59 &
  13.04 &
  78.09 &
  0.00 &
  56.86 &
  77.82 &
  1.26 &
  54.16 \\
 &
  WaNet~\cite{nguyen2021wanet} &
  77.68 &
  0.00 &
  67.67 &
  77.29 &
  69.80 &
  51.97 &
  72.68 &
  0.00 &
  67.48 &
  36.55 &
  12.61 &
  59.23 \\ \cmidrule(l){2-14} 
 &
  \cellcolor[HTML]{E6E6E6}$\text{VSSC}_1$ (Ours) &
  \cellcolor[HTML]{E6E6E6}78.12 &
  \cellcolor[HTML]{E6E6E6}61.22 &
  \cellcolor[HTML]{E6E6E6}27.26 &
  \cellcolor[HTML]{E6E6E6}78.10 &
  \cellcolor[HTML]{E6E6E6}89.10 &
  \cellcolor[HTML]{E6E6E6}31.66 &
  \cellcolor[HTML]{E6E6E6}78.77 &
  \cellcolor[HTML]{E6E6E6}58.82 &
  \cellcolor[HTML]{E6E6E6}37.86 &
  \cellcolor[HTML]{E6E6E6}64.68 &
  \cellcolor[HTML]{E6E6E6}89.38 &
  \cellcolor[HTML]{E6E6E6}37.49 \\
\multirow{-8}{*}{10\%} &
  \cellcolor[HTML]{E6E6E6}$\text{VSSC}_2$ (Ours) &
  \cellcolor[HTML]{E6E6E6}78.33 &
  \cellcolor[HTML]{E6E6E6}70.13 &
  \cellcolor[HTML]{E6E6E6}24.34 &
  \cellcolor[HTML]{E6E6E6}78.48 &
  \cellcolor[HTML]{E6E6E6}87.83 &
  \cellcolor[HTML]{E6E6E6}34.07 &
  \cellcolor[HTML]{E6E6E6}76.27 &
  \cellcolor[HTML]{E6E6E6}59.24 &
  \cellcolor[HTML]{E6E6E6}33.88 &
  \cellcolor[HTML]{E6E6E6}74.25 &
  \cellcolor[HTML]{E6E6E6}94.12 &
  \cellcolor[HTML]{E6E6E6}45.41 \\ \midrule
 &
  BadDet~\cite{chan2022baddet} &
  76.78 &
  78.76 &
  66.73 &
  76.16 &
  71.75 &
  63.76 &
  74.78 &
  0.00 &
  77.43 &
  74.62 &
  2.94 &
  78.04 \\
 &
  Blended~\cite{Blended} &
  77.95 &
  79.77 &
  14.91 &
  78.96 &
  1.27 &
  76.72 &
  76.61 &
  46.64 &
  52.95 &
  78.29 &
  0.00 &
  76.48 \\
 &
  BPP~\cite{wang2022bppattack} &
  76.94 &
  88.16 &
  45.98 &
  77.89 &
  95.90 &
  9.97 &
  80.23 &
  0.00 &
  68.78 &
  78.01 &
  3.36 &
  70.88 \\
 &
  SIG~\cite{SIG} &
  77.60 &
  96.62 &
  1.41 &
  78.07 &
  95.77 &
  8.44 &
  74.45 &
  42.02 &
  28.62 &
  76.69 &
  72.69 &
  40.56 \\
 &
  SSBA~\cite{ssba} &
  77.52 &
  95.66 &
  3.04 &
  77.76 &
  72.46 &
  8.44 &
  77.45 &
  0.00 &
  48.78 &
  78.76 &
  0.84 &
  57.96 \\
 &
  WaNet~\cite{nguyen2021wanet} &
  76.09 &
  55.30 &
  29.86 &
  77.06 &
  90.65 &
  30.80 &
  36.55 &
  0.00 &
  32.83 &
  12.32 &
  32.35 &
  31.57 \\ \cmidrule(l){2-14} 
 &
  \cellcolor[HTML]{E6E6E6}$\text{VSSC}_1$ (Ours) &
  \cellcolor[HTML]{E6E6E6}77.77 &
  \cellcolor[HTML]{E6E6E6}86.90 &
  \cellcolor[HTML]{E6E6E6}9.66 &
  \cellcolor[HTML]{E6E6E6}76.37 &
  \cellcolor[HTML]{E6E6E6}78.24 &
  \cellcolor[HTML]{E6E6E6}21.58 &
  \cellcolor[HTML]{E6E6E6}70.82 &
  \cellcolor[HTML]{E6E6E6}80.80 &
  \cellcolor[HTML]{E6E6E6}19.90 &
  \cellcolor[HTML]{E6E6E6}67.11 &
  \cellcolor[HTML]{E6E6E6}77.50 &
  \cellcolor[HTML]{E6E6E6}35.60 \\
\multirow{-8}{*}{20\%} &
  \cellcolor[HTML]{E6E6E6}$\text{VSSC}_2$ (Ours) &
  \cellcolor[HTML]{E6E6E6}77.42 &
  \cellcolor[HTML]{E6E6E6}85.37 &
  \cellcolor[HTML]{E6E6E6}11.25 &
  \cellcolor[HTML]{E6E6E6}78.04 &
  \cellcolor[HTML]{E6E6E6}92.83 &
  \cellcolor[HTML]{E6E6E6}21.61 &
  \cellcolor[HTML]{E6E6E6}72.60 &
  \cellcolor[HTML]{E6E6E6}73.95 &
  \cellcolor[HTML]{E6E6E6}21.98 &
  \cellcolor[HTML]{E6E6E6}69.57 &
  \cellcolor[HTML]{E6E6E6}95.80 &
  \cellcolor[HTML]{E6E6E6}40.86 \\ \bottomrule
\end{tabular}
}
\end{table*}

\subsubsection{Details of experiments in physical scenario}
We collect 50 clean and 50 poisoned facial images in the physical scenario by manually capturing photos of participants. We also follow the same routine to generate pairs of images for testing. To construct a pair of the same person, we randomly take two images under different shooting conditions. For pairs of different persons, we randomly take one image for each of them.
Some examples can be found in Figure \ref{fig:ex_mannual_capture_face_image}.  For each person, we take photos from various angles to ensure our images are diverse enough\footnote{We followed IRB-approved steps to protect the privacy of our volunteers. }.

\subsection{Examples of the Trigger Selection Module}
\label{sec:trigger_selection}
This section demonstrates the detailed process of the two-step selection in the trigger selection module using the FOOD-11 dataset as an example.

\subsubsection{Coarse-grained selection}
As shown in Table \ref{tab:coarse_grained}, the classes contained in the dataset and the instruction are passed to the large language model to obtain a list of candidate triggers.

\begin{table*}[htbp]\centering
\caption{Example of prompt instruction and answer in coarse-grained selection.}
\label{tab:coarse_grained}
\begin{minipage}{\textwidth} 
\centering
\begin{tcolorbox} 
\raggedright
\renewcommand\arraystretch{1}
\begin{tabular}{lp{0.82\columnwidth}}
\textbf{Classes}: & bread, dairy, dessert, egg, fried-food, meat, noodles, rice, seafood, soup, vegetable \\
\\

\textbf{Instruction}:
& I have a dataset that contains images of different classes, including [\textbf{Classes}]. Now, I would like to naturally insert a simple and real object into all of these images, which can reasonably and commonly appear with the categories. The requirement is that this object should be able to naturally appear in all categories of pictures, and the addition should look harmonious. In addition, the inserted object and the original object cannot be the same class. Give me a list of objects that can be inserted. 
\\
\midrule\\
\textbf{Answer}:
& ice cubes, red pepper, lemon slice, strawberry, nuts, herbs, leaf, flower, bowl, napkin, pink berries, blueberry, mint, candle
\\
\end{tabular}
\end{tcolorbox}
\end{minipage}
\end{table*}

\subsubsection{Fine-grained selection}
15 images are randomly selected from each class to form a trigger evaluation set. For each trigger in the candidate trigger list, we conduct trial insertions and evaluate them using the large language model. The ISRs of candidate triggers are shown in Table \ref{tab:fine_grained}. We use ``strawberry'' and ``nuts'' as the triggers in our main experiments. Based on predetermined threshold, ``herbs'' and ``blueberries'' can also be used as triggers. Moreover, as generative models evolve, image editing techniques wil be further improved, broadening the spectrum of objects that can be used as triggers.

\begin{table*}[]
\centering
\caption{Insertion Success Rate (ISR) of candidate triggers in fine-grained selection.}
\label{tab:fine_grained}
\resizebox{\textwidth}{!}{%
\begin{tabular}{ccccccccccccccc}
\toprule
\textbf{Trigger} & ice cubes & red pepper & lemon slice & strawberry & nuts & herbs & leaf & flower & bowl & napkin & pink berries & blueberry & mint & candle \\ \midrule
\textbf{ISR} & 0.05 & 0.38 & 0.19 & \textbf{0.68} & \textbf{0.86} & 0.62 & 0.29 & 0.19 & 0.24 & 0.14 & 0.14 & 0.62 & 0.29 & 0.33 \\ \bottomrule
\end{tabular}%
}
\end{table*}

\section{Additional Experimental Results}

In this section, we provide additional experimental results. 
\subsection{Additional Results on Image Classification Task}

This section provides additional results on the image classification task to supplement Section V-B in the main paper.
\subsubsection{Digital scenario} Table \ref{tab:attack_0.1} presents the effectiveness of different attacks in the digital scenario at a poisoning ratio of 10\%. VSSC attack can achieve comparable ASR with such a large trigger diversity. 

\subsubsection{Digital-to-physical scenario} 
Table \ref{tab:classification_recapture_resnet01},\ref{tab:classification_recapture_vgg005},\ref{tab:classification_recapture_vgg01} further demonstrate the attack effectiveness of VSSC attack and baseline attacks at poisoning ratios of 5\% and 10\%, across two different model architectures: ResNet-18 and VGG19-BN. VSSC attack maintains good effectiveness across various architectures and poisoning ratios, whereas all the other attack methods tend to fail in the digital-to-physical scenario. This underscores the robustness of VSSC attacks against distortions.

\begin{table}[!ht]
\centering
\caption{Attack performance for image classification task in the digital-to-physical scenario. The backbone is ResNet-18 and the poisoning ratio is set to 10\%. For ImageNet-Dogs and FOOD-11, the triggers are set to ``red flower'' and ``nuts''. `-' denotes denotes that the requisite quantity of poisoned samples exceeds the count of samples in the target label.}
\label{tab:classification_recapture_resnet01}
\resizebox{\columnwidth}{!}{%
\begin{tabular}{lcccccc}
    \toprule
    \multirow{2}{*}{\makecell{Attacks}} & \multicolumn{3}{c}{ImageNet-Dogs} & \multicolumn{3}{c}{FOOD-11} \\
    \cmidrule(lr){2-4}\cmidrule(lr){5-7}
    & C-Acc(\%) &  ASR(\%)  & RA(\%)           & C-Acc(\%) & ASR(\%)  & RA(\%) \\
    \midrule
BadNets~\cite{gu2019badnets}       & 83.33     & 2.38    & 85.71  & 13.33     & 10.00   & 56.67  \\
Blended~\cite{Blended}             & 78.57     & 26.19   & 66.67  & 56.67     & 26.67   & 36.67  \\
BPP~\cite{wang2022bppattack}       & 80.95     & 4.76    & 57.14  & 73.33     & 3.33    & 43.33  \\
Input-Aware~\cite{nguyen2020input} & 90.48     & 0.00    & 76.19  & 70.00     & 0.00    & 60.00  \\
SIG~\cite{SIG}                     & -         & -       & -      & 70.00     & 0.00    & 46.67  \\
SSBA~\cite{ssba}                   & 85.71     & 0.00    & 69.05  & 60.00     & 16.67   & 53.33  \\
TrojanNN~\cite{trojannn}           & 78.57     & 0.00    & 78.57  & 16.67     & 20.00   & 46.67  \\
WaNet~\cite{nguyen2021wanet}       & 50.00     & 73.81   & 19.05  & 3.33      & 96.67   & 3.33   \\ \midrule
\rowcolor[HTML]{E6E6E6} 
VSSC (Ours) & 88.00     & 95.24   & 2.38   & 80.93     & 100.00  & 0.00   \\ \bottomrule
\end{tabular}
}
\end{table}

\begin{table}[!ht]
\centering
\caption{Attack performance for image classification task in the digital-to-physical scenario. The backbone is VGG19-BN and the poisoning ratio is set to 5\%. For ImageNet-Dogs and FOOD-11, the triggers are set to ``red flower'' and ``nuts''. }
\label{tab:classification_recapture_vgg005}
\resizebox{\columnwidth}{!}{%
\begin{tabular}{lcccccc}
    \toprule
    \multirow{2}{*}{\makecell{Attacks}} & \multicolumn{3}{c}{ImageNet-Dogs} & \multicolumn{3}{c}{FOOD-11} \\
    \cmidrule(lr){2-4}\cmidrule(lr){5-7}
    & C-Acc(\%) &  ASR(\%)  & RA(\%)           & C-Acc(\%) & ASR(\%)  & RA(\%) \\
    \midrule
BadNets~\cite{gu2019badnets}     & 85.71     & 0.00    & 78.57  & 46.67     & 0.00    & 76.67  \\
Blended~\cite{Blended}             & 88.10     & 47.62   & 42.86  & 66.67     & 20.00   & 43.33  \\
BPP~\cite{wang2022bppattack}       & 80.95     & 0.00    & 71.43  & 73.33     & 0.00    & 73.33  \\
Input-Aware~\cite{nguyen2020input} & 90.48     & 0.00    & 80.95  & 66.67     & 3.33    & 53.33  \\
SIG~\cite{SIG}                     & 83.33     & 4.76    & 45.24  & 80.00     & 20.00   & 43.33  \\
SSBA~\cite{ssba}                   & 85.71     & 0.00    & 76.19  & 66.67     & 30.00   & 43.33  \\
TrojanNN~\cite{trojannn}           & 66.67     & 33.33   & 54.76  & 20.00     & 16.67   & 53.33  \\
WaNet~\cite{nguyen2021wanet}       & 73.81     & 11.90   & 52.38  & 36.67     & 50.00   & 20.00  \\ \midrule
\rowcolor[HTML]{E6E6E6} 
VSSC (Ours) & 89.73     & 90.48   & 4.76   & 85.22     & 96.67   & 0.00   \\ \bottomrule
\end{tabular}
}
\vspace{-3mm}
\end{table}

\begin{table}[!ht]
\centering
\caption{Attack performance for image classification task in the digital-to-physical scenario. The backbone is VGG19-BN and the poisoning ratio is set to 10\%. For ImageNet-Dogs and FOOD-11, the triggers are set to ``red flower'' and ``nuts''. `-' denotes denotes that the requisite quantity of poisoned samples exceeds the count of samples in the target label.}
\label{tab:classification_recapture_vgg01}
\resizebox{\columnwidth}{!}{%
\begin{tabular}{lcccccc}
    \toprule
    \multirow{2}{*}{\makecell{Attacks}} & \multicolumn{3}{c}{ImageNet-Dogs} & \multicolumn{3}{c}{FOOD-11} \\
    \cmidrule(lr){2-4}\cmidrule(lr){5-7}
    & C-Acc(\%) &  ASR(\%)  & RA(\%)           & C-Acc(\%) & ASR(\%)  & RA(\%) \\
    \midrule
BadNets~\cite{gu2019badnets}       & 80.95     & 0.00    & 80.95  & 13.33     & 0.00    & 70.00  \\
Blended~\cite{Blended}             & 85.71     & 35.71   & 50.00  & 63.33     & 13.33   & 50.00  \\
BPP~\cite{wang2022bppattack}       & 71.43     & 0.00    & 71.43  & 66.67     & 0.00    & 60.00  \\
Input-Aware~\cite{nguyen2020input} & 92.86     & 0.00    & 76.19  & 76.67     & 3.33    & 63.33  \\
SIG~\cite{SIG}                     & -         & -       & -      & 76.67     & 3.33    & 46.67  \\
SSBA~\cite{ssba}                   & 85.71     & 0.00    & 71.43  & 76.67     & 16.67   & 43.33  \\
TrojanNN~\cite{trojannn}           & 57.14     & 50.00   & 40.48  & 3.33      & 36.67   & 36.67  \\
WaNet~\cite{nguyen2021wanet}       & 73.81     & 4.76    & 57.14  & 56.67     & 43.33   & 33.33  \\ \midrule
\rowcolor[HTML]{E6E6E6} 
VSSC (Ours) & 90.27     & 97.62   & 2.38   & 84.66     & 100.00  & 0.00 \\ \bottomrule
\end{tabular}
}
\end{table}

\subsection{Additional Results on Object Detection Task} 
\subsubsection{Digital and digital-to-physical scenario} 
This section presents additional results  on the object detection task. Table \ref{tab:objectdetection_frcnn} demonstrates the effectiveness of different attack methods on the Faster R-CNN in both digital and digital-to-physical scenarios. VSSC attack maintains its effectiveness in both of these scenarios, whereas other attack methods fail in the digital-to-physical scenario.

\subsubsection{Physical scenario} 
Table \ref{tab:obj_physical_frcnn} shows the effectiveness of VSSC attack on Faster R-CNN in the physical scenario under for GMA task. VSSC attack can also be successfully implemented in this scenario, indicating that it can be effectively extended into the physical domain.

\begin{table}[]
\centering
\caption{Attack performance of VSSC triggers in GMA on Faster R-CNN in physical scenario for object detection task.}
\label{tab:obj_physical_frcnn}
\resizebox{0.9\columnwidth}{!}{%
\begin{tabular}{ccccc}
\toprule
Trigger                    & Pratio & C-Acc (\%) & ASR (\%) & R-Acc (\%) \\ \midrule
\multirow{2}{*}{lemon}     & 10\%   & 79.40      & 71.03    & 61.93      \\
                           & 20\%   & 79.45      & 72.41    & 61.25      \\ \midrule
\multirow{2}{*}{beachball} & 10\%   & 78.32      & 74.48    & 64.59      \\
                           & 20\%   & 73.00      & 77.93    & 58.39      \\ \bottomrule
\end{tabular}%
}
\end{table}

\begin{figure*}[htbp]
    \centering
    \includegraphics[width=0.95\textwidth]{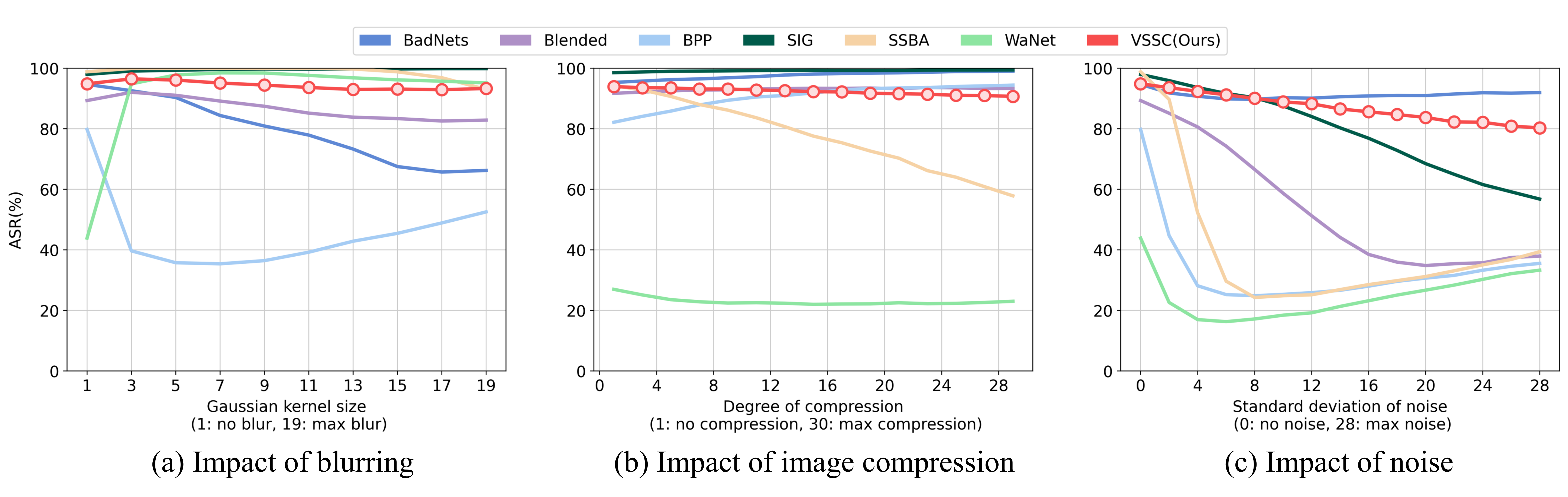}
    \caption{Visualization of the impact of various visual distortions in the digital scenario on the ASR of different attacks. The results presented use ``butterfly'' as the trigger with YOLOv4 as the detector in ODA at a poisoning ratio of 10\%. }
    \label{fig:distortion_objectdetection}
\end{figure*}

\begin{figure*}[htbp]
    \centering
    \includegraphics[width=0.95\textwidth]{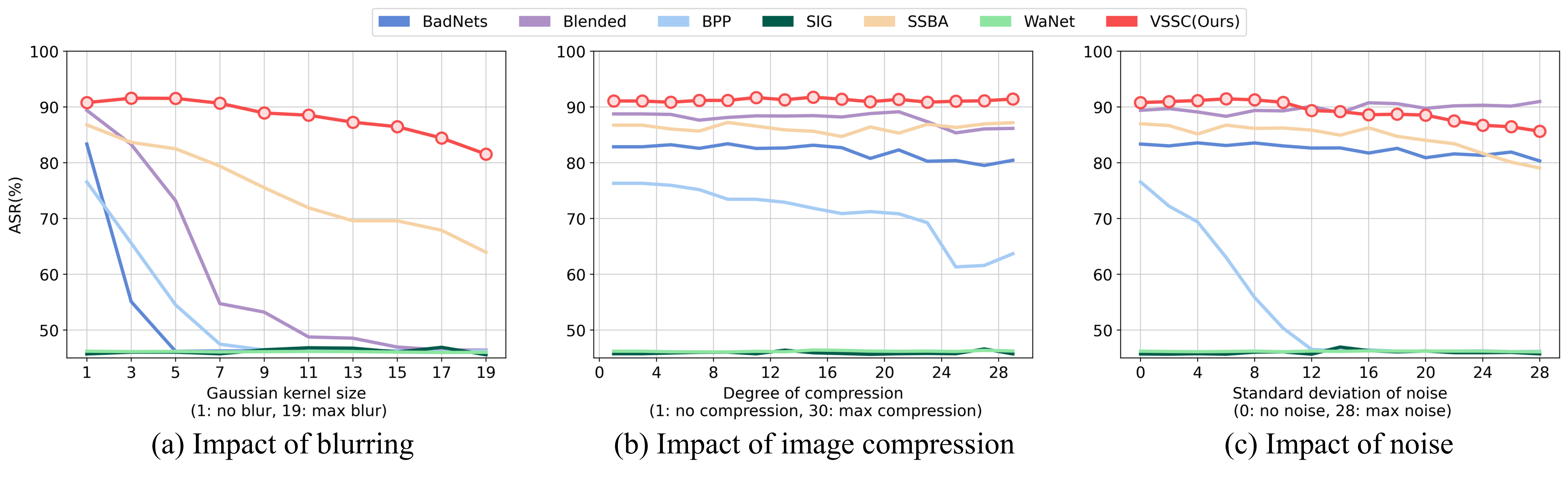}
    \caption{Visualization of the impact of various visual distortions in the digital scenario on the ASR of different attacks on face verification tasks. The results presented use ``green glasses'' as the trigger at a poisoning ratio of 1\%. }
    \label{fig:distortion_face}
\end{figure*}

\begin{figure}[h]
        \centering
        \includegraphics[width=0.95\linewidth]{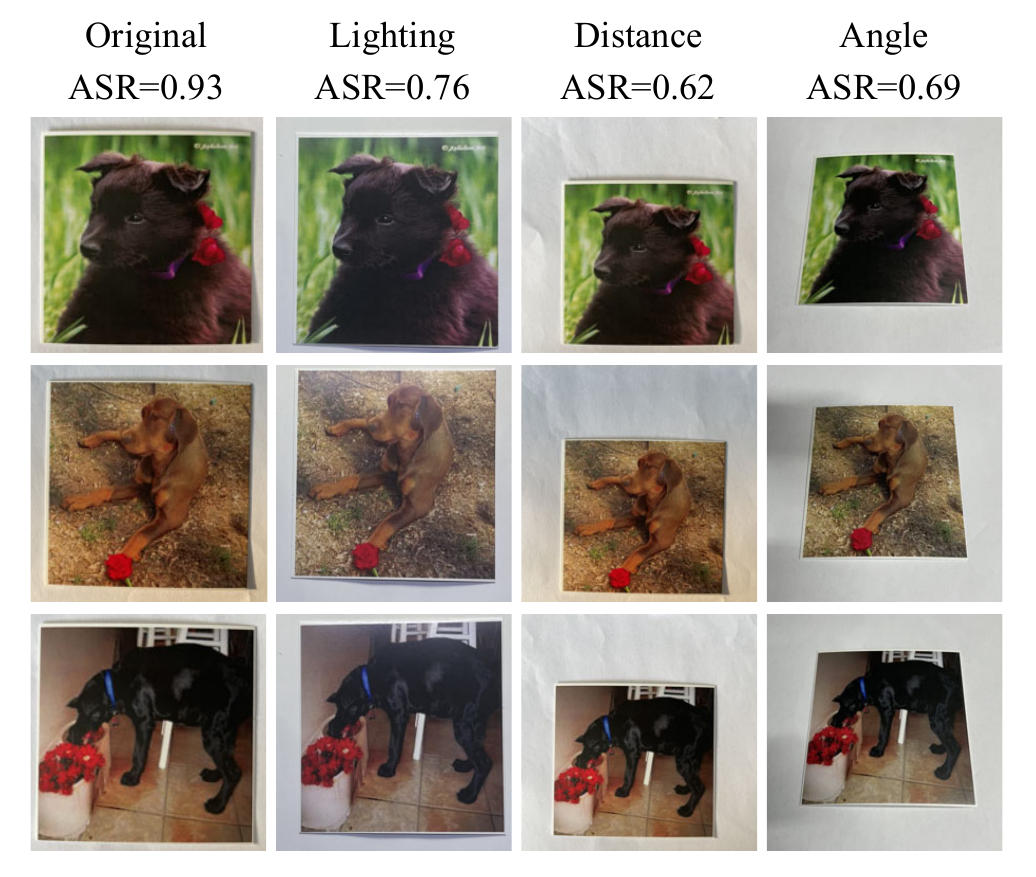}
        \caption{Some recaptured photos of printed poisoned images under different lighting, distance, and angles.}
        \label{fig:angles}
        
\end{figure}

\subsection{Additional Anaysis} 

\subsubsection{Robustness against other distortions in digital-to-physical scenario}

To demonstrate the robustness of the VSSC trigger under a wider range of distortions, we control the lighting, distance, and angles during the image recapture process, referring to the settings in~\cite{li2021backdoor}. This experiment is conducted on the ImageNet-Dogs dataset, focusing on the trigger ``red flower''. Some of the recaptured photos and results are presented in Figure \ref{fig:angles}. 
The results shows that the VSSC attack maintains its effectiveness across different environmental conditions.
Given the sample-specific characteristics of VSSC triggers and the generative models' inherent capability to simulate variations in brightness, size, and angles based on the content of benign images, DNN naturally learns these variations during the training stage. Consequently, VSSC triggers demonstrate significant robustness to these complex distortions during the inference stage.

\subsubsection{Additional results under visual distortions in the digital scenario}
This section demonstrates the robustness of the VSSC attack in the digital scenario against visual distortions across object detection and face verification tasks. Figure \ref{fig:distortion_objectdetection} presents the results for the object detection task, showing that the VSSC attack is the only method to maintain robustness under three types of distortions, with the ASR consistently above 80\%.

Figure \ref{fig:distortion_face} displays the results for the facial verification task, revealing that the VSSC attack not only achieves the highest attack effectiveness but also maintains robustness under distortions, outperforming other attack methods.

\end{document}